\documentclass[11pt,a4paper]{article}
\usepackage[left=3cm,right=3cm,top=3cm, bottom=2cm]{geometry}
\usepackage{amsmath,amsfonts,bm}

\def\eqref#1{equation~\ref{#1}}
\def\1{\bm{1}}

\DeclareMathAlphabet{\mathsfit}{\encodingdefault}{\sfdefault}{m}{sl}
\SetMathAlphabet{\mathsfit}{bold}{\encodingdefault}{\sfdefault}{bx}{n}

\newcommand{\E}{\mathbb{E}}

\newcommand{\R}{\mathbb{R}}

\newcommand{\Var}{\mathrm{Var}}

\usepackage{hyperref}
\usepackage{url}

\usepackage[utf8]{inputenc} % allow utf-8 input
\usepackage{hyperref}       % hyperlinks
\usepackage{booktabs}       % professional-quality tables
\usepackage{amsfonts}       % blackboard math symbols
\usepackage{nicefrac}       % compact symbols for 1/2, etc.
\usepackage{microtype}      % microtypography
\usepackage[linesnumbered]{algorithm2e}
\usepackage{enumitem}
\usepackage{amsthm}
\usepackage{booktabs}
\usepackage[autostyle=try, english=american]{csquotes}
\MakeOuterQuote{"}
\usepackage{amsmath}
\usepackage{cancel}
\usepackage{footnote}
\usepackage[symbol]{footmisc}

\usepackage{tablefootnote}
\usepackage{wrapfig}
\usepackage[export]{adjustbox}
\usepackage{amssymb}
\usepackage{xcolor}
\usepackage{hyperref}
\usepackage{float} % http://ctan.org/pkg/float
\usepackage[lofdepth,lotdepth]{subfig}

\usepackage{microtype}
\usepackage{graphicx}
\usepackage{booktabs} % for professional tables
\usepackage{hyperref}
\usepackage[colorinlistoftodos,prependcaption]{todonotes}
\usepackage{multirow}

\newtheorem{theorem}{Theorem}

\newtheorem{definition}{Definition}
\newtheorem{assumption}{Assumption}
\newtheorem{proposition}{Proposition}

\DeclareMathOperator{\FT}{FT}

\newcommand{\Prob}{\mathbb{P}}
\newcommand{\Be}{{\rm{Be}}}

\newtheoremstyle{exampstyle}
  {1pt} % Space above
  {0pt} % Space below
  {} % Body font
  {} % Indent amount
  {\bfseries} % Theorem head font
  {.5em} % Space after theorem head
  {} % Theorem head spec (can be left empty, meaning `normal')

\theoremstyle{exampstyle}

\usepackage{titlesec}
\titlespacing{\section}{2pt}{2pt}{2pt}
\titlespacing{\subsection}{2pt}{0pt}{0pt}

\title{The Probabilistic Fault Tolerance of Neural Networks in The Continuous Limit}

\author{
	El-Mahdi El-Mhamdi, Rachid Guerraoui, Andrei Kucharavy, Sergei Volodin \\
	Distributed Computing Laboratory\\
	Swiss Federal Institute of Technology in Lausanne (EPFL)\\
	1015-Lausanne, Switzerland \\
	\{elmahdi.elmhamdi, rachid.guerraoui, andrei.kucharavy, sergei.volodin\}@epfl.ch
}

\begin{document}

\maketitle

\begin{abstract}
The loss of a few neurons in a brain rarely results in any visible loss of function.
However, the insight into what "few" means in this context is
unclear.
How many random neuron failures will it take to lead to a visible loss of function?
In this paper, we address the fundamental question of the impact of the crash of a random subset of neurons on the overall computation of a neural network and the error in the output it produces. We study fault tolerance of neural networks subject to small random neuron/weight crash failures in a probabilistic setting. We give provable guarantees on the robustness of the network to these crashes.
Our main contribution is a bound on the error in the output of a network under small random Bernoulli crashes proved by using a Taylor expansion in the continuous limit, where close-by neurons at a layer are similar.
%The assumptions underlying our guarantees are grounded in the properties of neuromorphic hardware, an emerging computing paradigm.
The failure mode we adopt in our model is characteristic of neuromorphic hardware, a promising technology to speed up artificial neural networks, as well as of biological networks.
We show that our theoretical bounds can be used to compare the fault tolerance of different architectures and to design a regularizer improving the fault tolerance of a given architecture. We design an algorithm achieving fault tolerance using a reasonable number of neurons. In addition to the theoretical proof, we also provide experimental validation of our results and suggest a connection to the generalization capacity problem.
\end{abstract}

%\tableofcontents

% list of todos
%\hrule
%\listoftodos
%\hrule

%\todo[inline]{Make sure the shortened abstract is good}
%\todo[inline]{Make the intro more clear and short}
%\todo[inline]{Use some clear symbol for "see supplementary" because we have tons of references there}
%\todo[inline]{Shrink size to 8 pages}
%\todo[inline]{Reduce number of footnotes, now have 15!! -- moving loss bounded, inspiration, certify - shorten and move, minima sharpness, limits of appl shorten and move, total loss, P3->suppl NTK to suppl}

\section{Introduction}
% {\bf Why Fault Tolerance?}
% \andrei{Nope, too specialized of a start, Mehdi's start was pretty good actually.}

Understanding the inner working of artificial neural networks (NNs) is currently one of the most pressing questions~\cite{lecun2015deep} in learning theory. As of now, neural networks are the backbone of the most successful machine learning solutions~\cite{silver2016mastering, alexnet}. They are deployed in safety-critical tasks in which there is little room for mistakes \cite{esteva2017dermatologist,stilgoe2018machine}. Nevertheless, such issues are regularly reported since attention was brought to the NNs vulnerabilities over the past few years~\cite{silver2016mastering, biggio2013evasion, moosavi2017universal, bulyanPaper}. 
%\andrei{Is this last sentece really necessary?}
%This leads to the field of technical Artificial Intelligence (AI) safety \cite{amodei2016concrete}.

{\bf Fault tolerance as a part of theoretical NNs research.}
Understanding complex systems requires understanding how they can tolerate failures of their components. This has been a particularly fruitful method in systems biology, where the mapping of the full network of metabolite molecules is a computationally quixotic venture. Instead of fully mapping the network, biologists improved their understanding of biological networks
by studying the effect of deleting some of their components, one or a few perturbations at a time~\cite{costanzo2016global,glass2006essential}. Biological systems in general are found to be fault tolerant~\cite{navlakha2015distributed}, which is thus an important criterion for biological plausibility of mathematical models.

{\bf Neuromorphic hardware (NH).} Current Machine Learning systems are bottlenecked by the underlying computational power \cite{amodei}. One significant improvement over the now prevailing CPU/GPUs is {\em neuromorphic hardware}. In this paradigm of computation, each neuron is a physical entity \cite{neuromorphicIBM}, and the forward pass is done (theoretically) at the speed of light. However, components of such hardware are small and unreliable, leading to small random perturbations of the weights of the model \cite{torres2017fault}. Thus, robustness to weight faults is an overlooked concrete Artificial Intelligence (AI) safety problem \cite{amodei2016concrete}. Since we ground the assumptions of our model in the properties of NH and of biological networks, our fundamental theoretical results can be directly applied in these computing paradigms.

{\bf Research on NN fault tolerance.} %In the early days of connectionism, fault tolerance was often listed among the motivations behind neural networks, with arguments such as the \emph{graceful degradation}~\cite{haykin2009neural, kerlirzin, bengio} of neural networks, imposing a continuous worsening of performance under neuron crashes, that can be traced back to the foundational work of Rosenblatt on the \emph{perceptron}~\cite{perceptron}.
In the 2000s, the fault tolerance of NNs was a major motivation for studying them \cite{haykin2009neural,kerlirzin,bengio}. In the 1990s, the exploration of microscopic failures was fueled by the hopes of developing neuromorphic hardware (NH) \cite{darpa, chiu1993robustness, piuri}.
% \footnote{It is ironic that connectionism back then, was called \emph{Parallel and Distributed Processing~\cite{connectionism},} the name of a field which is all about fault tolerance but almost not at all about neural networks today.}
%The quest of building energy-efficient hardware that is itself a neural network, freed from the Von Neumann bottleneck, kept a line of research on fault tolerance alive until the late 1990s with VLSI circuits of neural networks as a motivation~\cite{darpa, chiu1993robustness, piuri}. 
Taylor expansion was one of the tools used for the study of fault tolerance \cite{hammadi1997learning,murray1994enhanced}. Another line of research proposes sufficient conditions for robustness \cite{phatak1995complete}. However, most of these studies are either empirical or are limited to simple architectures \cite{torres2017fault}. In addition, those studies address the worst case \cite{biggio2013evasion}, which is known to be more severe than a random perturbation.
Recently, fault tolerance was studied experimentally as well. DeepMind proposes to focus on neuron removal \cite{deepmindremoval} to understand NNs. NVIDIA \cite{NVIDIAerror} studies error propagation caused by micro-failures in hardware \cite{arechiga2018robustness}. In addition, mathematically similar problems are raised in the study of generalization \cite{Neyshabur2017,BehnamNeyshaburSrinadhBhojanapalli2018} and robustness \cite{weng2018proven}.

{\bf The quest for guarantees.} Existing NN approaches do not guarantee fault tolerance: they  only provide heuristics and evaluate them experimentally.
Theoretical papers, in turn, focus on the worst case and not on errors in a probabilistic sense. It is known that there exists a set of small {\em worst-case} perturbations, {\em adversarial examples} \cite{biggio2013evasion}, leading to pessimistic bounds not suitable for the {\em average case} of random failures, which is the most realistic case  for hardware faults. Other branch of theoretical research studies robustness and arrives at error bounds which, unfortunately, scale exponentially with the depth of the network \cite{Neyshabur2017}. We define the goal of this paper to guarantee that the probability of loss exceeding a threshold is lower than a pre-determined small value. This condition is sensible. For example, self-driving cars are deemed to be safe once their probability of a crash is several orders of magnitude less than of human drivers \cite{stilgoe2018machine,kalra2016many,schwarting2018planning}. In addition, current fault tolerant architectures use {\em mean} as the aggregation of copies of networks to achieve redundancy. This is known to require exponentially more redundancy compared to the {\em median} approach and, thus, hardware cost. In order to apply this  powerful technique and reduce costs, certain conditions need to be satisfied which we will evaluate for neural networks.
%\andrei{I do no understand the interest of that last sentence. I did not get that we were doing median vs mean as well.}\sergei{The interest is that we allow to have exponentially less redundance. The median or the mean is a well-known dilemma when estimating over a population}.

{\bf Contributions.}
Our main contribution is a theoretical bound on the error in the output of an NN in the case of random neuron crashes obtained in the continuous limit, where close-by neurons compute similar functions. We show that, while the general problem of fault tolerance is NP-hard, realistic assumptions with regard to neuromorphic hardware, and a probabilistic approach to the problem, allow us to apply a Taylor expansion for the vast majority of the cases, as the weight perturbation is small with high probability. In order for the Taylor expansion to work, we assume that a network is smooth enough, introducing the {\em continuous limit} \cite{sonoda2017double} to prove the properties of NNs: it requires neighboring neurons at each layer to be similar. This makes the moments of the error linear-time computable. To our knowledge, the tightness of the bounds we obtain is a novel result. In turn, the bound allows us to build an algorithm that enhances fault tolerance of neural networks.
%Crucially, for the algorithm to work, we show that by using a median approach, we are able to reduce drastically the probability bound of high errors, providing a guarantee on the network architecture stability,
Our algorithm uses median aggregation which results in only a logarithmic extra cost -- a drastic improvement on the initial NP-hardness of the problem. Finally, we show how to apply the bounds to specific architectures and evaluate them experimentally on real-world networks, notably the widely used VGG \cite{simonyan2014very}.
{\bf Outline.} In Sections~\ref{sec:theory}-\ref{sec:realistic}, we set the formalism, then state our bounds.
In Section~\ref{sec:guarantee}, we present applications of our bounds on characterizing the fault tolerance of different architectures. In Section~\ref{sec:algorithm} we present our algorithm for certifying fault tolerance.
In Section~\ref{sec:experiments}, we present our experimental evaluation. Finally, in Section~\ref{sec:conclusion}, we discuss the consequences of our findings.
Full proofs are available in the supplementary material. Code is provided at the address {\tt \href{https://github.com/LPD-EPFL/ProbabilisticFaultToleranceNNs}{github.com/LPD-EPFL/ProbabilisticFaultToleranceNNs}}. We abbreviate Assumption 1 $\to$ A1, Proposition $1$ $\to$ P1, Theorem 1 $\to$ T1, Definition 1 $\to$ D1.

\section{Definitions of Probabilistic Fault Tolerance}
In this section, we define a fully-connected network and fault tolerance formally.

\label{sec:theory}

%\todo[inline]{Make notation and defs compatible with the ML community}

{\bf Notations.} For any two vectors $x,y\in\mathbb{R}^n$ we use the notation $(x,y)=\sum_{i=1}^n x_iy_i$ for the standard scalar product.
%The notation $|W|$ means element-wise absolute values of the matrix: $(|W|)_{ij}=|W_{ij}|$.
Matrix $\gamma$-norm for $\gamma=(0,+\infty]$ is defined as $\|A\|_{\gamma}=\sup_{x\neq 0}\|Ax\|_{\gamma}/\|x\|_{\gamma}$. We use the infinity norm $\|x\|_{\infty}=\max|x_i|$ and the corresponding operator matrix norm.
We call a vector $0\neq x\in\mathbb{R}^n$ $q$-balanced if $\min |x_i|\geq q \max |x_i|$. We denote $[n]=\{1,2,...,n\}$. We define the Hessian $H_{ij}={\partial^2 y(x)}/{\partial x^i\partial x^j}$ as a matrix of second derivatives. We write layer indices down and element indices up: $W_l^{ij}$. For the input, we write $x_i\equiv x^i$. If the layer is fixed, we omit its index. We use the element-wise Hadamard product $(x\odot y)_i=x_iy_i$.

\begin{definition}{(Neural network)}
    \label{def:nn}
	A neural network with $L$ layers is a function $y_L\colon \mathbb{R}^{n_0}\to\mathbb{R}^{n_L}$ defined by a tuple $(L,W,B,\varphi)$ with a tuple of weight matrices $W=(W_1,...,W_L)$ (or their distributions) of size $W_l\colon n_l\times n_{l-1}$, biases $B=(b_1,...,b_L)$ (or their distributions) of size $b_l\in \mathbb{R}^{n_l}$ by the expression
	$y_l=\varphi(z_l)$ with pre-activations $z_l=W_ly_{l-1}+b_l,\,l\in[L]$, $y_0=x$ and $y_L=z_L$. Note that the last layer is linear. We additionally require $\varphi$ to be 1-Lipschitz \footnote{1-Lipschitz $\varphi$ s.t. $|\varphi(x)-\varphi(y)|\leqslant |x-y|$. If $\varphi$ is $K$-Lipschitz, we rescale the weights to make $K=1$: $W^{ij}_l\to W^{ij}_l/K$. This is the general case. Indeed, if we rescale $\varphi(x)\to K\varphi(x)$, then, $y_{l-1}\to K y_{l-1}'$, and in the sum $z_l'=\sum W^{ij}/K\cdot K y_{l-1}\equiv z_l$}.
	We assume that the network was trained using input-output pairs $x,y^*\sim X\times Y$ using ERM\footnote{Empirical Risk Minimization -- the standard task $1/k\sum_{k=1}^m\omega(y_L(x_k), y^*_k)\to\min$} for a loss $\omega$.
    Loss layer for input $x$ and the true label $y^*(x)$ is defined as $y_{L+1}(x)=\E_{y^*\sim Y|x}\omega(y_L(x),y^*))$ with $\omega\in[-1,1]$\footnote{The loss is bounded for the proof of Algorithm 1's running time to work}
\end{definition}
\begin{definition}{(Weight failure)}
    \label{def:weightfail}
	Network $(L,W,B,\varphi)$ with weight failures $U$ of distribution $U\sim D|(x,W)$ is the network $(L,W+U,B,\varphi)$ for $U\sim D|(x,W)$. We denote a (random) output of this network as $y^{W+U}(x)=\hat{y}_L(x)$ with activations $\hat{y}_l$ and pre-activations $\hat{z}_l$, as in D\ref{def:nn}. %We sometimes write $u$ and $w$ meaning matrices $u$
\end{definition}
%\begin{definition}{(Bernoilli weight failures)}
    %\label{def:bernoullifail}
    %Bernoulli weight crash distribution with probabilities $p_1,...,p_L$ is the distribution with fully independent $U_l^{ij}=-Be(p_l)\cdot W_l^{ij}$
%\end{definition}
%\todo[color=red]{Drop weight failure temporarily and focus on only neuron failure to save time?}
%\begin{definition}{(Bernoilli two-sided failures)}
    %\label{def:twosidedbernoullifail}
    %Bernoulli two-sided weight crash distribution with probabilities $p_1,...,p_L$ is the distribution with fully independent $U_l^{ij}=-\xi_l^{ij}\cdot W_l^{ij}$ where $\xi$ takes values $0,+1,-1$ with probabilities $1-p,p/2,p/2$.
%\end{definition}
\begin{definition}{(Bernoulli neuron failures)}
    \label{def:neuronfail}
    Bernoulli neuron crash distribution is the distribution with i.i.d. $\xi_l^i\sim \Be(p_l)$, $U_l^{ij}=-\xi_l^i\cdot W_l^{ij}$. For each possible crashing neuron $i$ at layer $l$ we define $U_l^i=\sum_{j}|U_l^{ij}|$ and $W_l^i=\sum_j |W_l^{ij}|$, the crashed incoming weights and total incoming weights. We note that we see neuron failure as a sub-type of weight failure.
\end{definition}
This definition means that neurons crash independently, and they start to output $0$ when they do. We use this model
%not only because of its simplicity, but also 
because it mimics essential properties of NH \cite{torres2017fault}. Components fail relatively independently, as we model faults as random \cite{torres2017fault}. %We consider the error probability between neurons of a layer to be the same, and the rationale is that neurons are assumed to be made from the same components.% The probability of failure $p$ during time $t$ can be computed from the rate of failure $\lambda$ simply as $p=\lambda t$.
In terms of \cite{torres2017fault}, we consider {\em stuck-at-0 crashes}, and {\em passive} fault tolerance in terms of {\em reliability}.
% We note that this is not the perfect model of faults: for example, faulty weights would be more realistic.
%\begin{definition}{(Bernoulli input failure)}
%\label{def:inputfail}
	%A network $(L,W,B,\varphi)$ is said to receive a Bernoulli failing input with $p$ if it has Bernoulli weight failure %with probabilities $(p,0,...,0)$.
%\end{definition}

\begin{definition}{(Output error for a weight distribution)}
    \label{def:outputfail}
	The error in case of weight failure with distribution $D|(x,W)$ is $\Delta_l(x)=y^{W+U}_l(x)-y^{W}_l(x)$ for layers $l\in[L+1]$
\end{definition}
We extend the definition of $\varepsilon$-fault tolerance from \cite{whenneuronsfail17} to the probabilistic case:

\begin{definition}{(Probabilistic fault tolerance)}
\label{def:faulttolerance}
A network $(L,W,B,\varphi)$ is said to be $(\varepsilon,\delta)$-fault tolerant over an input distribution $(x,y^*)\sim X\times Y$ and a crash distribution $U\sim D|(x,W)$ if
$
\mathbb P_{(x,y^*)\sim X\times Y,\,U\sim D|(x,W)}\{\Delta_{L+1}(x)\geq \varepsilon\}\leq\delta
$.
For such network, we write $(W,B)\in \FT(L,\varphi,p,\varepsilon,\delta)$.
\end{definition}
{\bf Interpretation.} To evaluate the fault tolerance of a network, we compute the first moments of $\Delta_{L+1}$. Next, we use tail bounds to guarantee $(\varepsilon,\delta)$-FT. This definition means that with high probability $1-\delta$ additional loss due to faults does not exceed $\varepsilon$. Expectation over the crashes $U\sim D|x$ can be interpreted in two ways. First, for a large number of neural networks, each having permanent crashes, $\E\Delta$ is the expectation over all instances of a network implemented in the hardware multiple times. For a single network with intermittent crashes, $\E\Delta$ is the output of this one network over repetitions. The recent review study \cite{torres2017fault} identifies three types of faults: permanent, transient, and intermittent. Our definition \ref{def:weightfail} thus covers all these cases.

%\begin{definition}[Sensitivity, data Jacobian or data Lipschitz coefficient]
%The sensitivity of the network at input $x$ is the quantity
%$\frac{\partial y_L}{\partial x}(x)$
%\end{definition}

Now that we have a definition of fault tolerance, we show in the next section that the task of certifying or even computing it is hard.

\section{The Hardness of Fault Tolerance}
\label{sec:hardness}
In this section, we show why fault tolerance is a hard problem. Not only it is NP-hard in the most general setting but, also, even for small perturbations, the error of the output of can be unacceptable.

\subsection{NP-Hardness}
\label{sec:nphardness}
A precise assessment of an NN's fault tolerance should ideally diagnose a network by looking at the outcome of every possible failure, i.e. at the \emph{Forward Propagated Error}~\cite{whenneuronsfail17} resulting from removing every possible subset of neurons. This would lead to an exact assessment, but would be impractical in the face of an exponential explosion of possibilities as by Proposition \ref{th:nphard} (proof in the supplementary material).

\begin{proposition}
\label{th:nphard}
The task of evaluating $\E\Delta^k$ for any $k=1,2,...$ with constant additive or multiplicative error for a neural network with $\varphi\in C^\infty$, Bernoulli neuron crashes and a constant number of layers is NP-hard.
\end{proposition}

We provide a theoretical alternative for the practical case of neuromorphic hardware. We overcome NP-hardness in Section \ref{sec:realistic} by providing an approximation dependent on the network, and not a constant factor one: for weights $W$ we give $\overline{\Delta}$ and $\underline{\Delta}$ dependent on $W$ such that $\underline{\Delta}(W)\leq\E\Delta\leq\overline{\Delta}(W)$. In addition, we only consider some subclass of all networks.

\subsection{Pessimistic Spectral Bounds}
\label{sec:spectral}
By Definition \ref{def:outputfail}, the fault tolerance assessment requires to consider a weight perturbation $W+U$ given current weights $W$ and the loss change $y_{L+1}(W+U)-y_{L+1}(W)$ caused by it. Mathematically, this means calculating a local Lipschitz coefficient $K$ \cite{zou2018lipschitz} connecting $|y_{L+1}(W+U)-y_{L+1}(W)|\leq K|U|$. In the literature, there are known {\em spectral} bounds on the Lipschitz coefficient for the case of input perturbations. These bounds use the spectral norm of the matrix $\|\cdot\|_2$ and give a global result, valid for any input. This estimate is loose due to its exponential growth in the number of layers, as $\|W\|_2$ is rarely $<1$. See Proposition \ref{th:spectral} for the statement:
\begin{proposition}[$K$ using spectral properties]
\label{th:spectral}
$
\|y_L(x_2)-y_L(x_1)\|_2\leqslant  \|x_2-x_1\|_2\cdot \prod_{l=1}^L\|W_l\|_2
$
\end{proposition}

The proof can be found in \cite{Neyshabur2017} or in the supplementary material. It is also known that high perturbations under small input changes are attainable. Adversarial examples \cite{biggio2013evasion} are small changes to the input resulting in a high change in the output. This bound is equal to the one of \cite{whenneuronsfail17}, which is tight in case if the network has the {\em fewest} neurons. In contrast, in Section \ref{sec:realistic}, we derive our bound in the limit $n\to\infty$.

%Making them $<1$ would result in vanishing gradients. One solution would be to use Parseval networks \cite{parseval} \todo[color=red]{Would Parseval networks just solve the problem then?}.

%\begin{proposition}[Lipschitz coefficient via norms]
%\label{th:bound_v1}
%The difference in output can be bounded as:
%$$
%\|y_L(x_2)-y_L(x_1)\|\leqslant \|W_L\|\cdot\ldots\cdot\|W_1\|\|x_2-x_1\|
%$$
%\end{proposition}

%.(W, x)=\frac{\|Wx\|}{\|x\|}
%\begin{proposition}[Input-specific spectral Lipschitz coefficient]
%For a neural network $(L,W,B,\varphi)$ with $\sigma_i=\sigma(W_i, \hat{y}_{i-1}-y_{i-1})$,
%$$
%\|y_L(x_2)-y_L(x_1)\|\leq \sigma_L\cdot\sigma_{L-1}\cdot...\cdot \sigma_1\|x_2-x_1\|
%$$
%\end{proposition}
%This proposition shows that in order to compute a more realistic estimate of the Lipschitz coefficient for a distribution of crashes, one needs to mathematically derive $\E \sigma_L\cdot\sigma_{L-1}\cdot...\cdot \sigma_1$ which can be problematic due to non-linear dependence of $y_{i}$ on the weight perturbation $W$.
%Another trivial idea of obtaining the Lipshitz coefficient is to use the triangle inequality and absolute values, but this also results in a loose estimate.
%\begin{proposition}[Lipschitz coefficient using the triangle inequality]
%The error on input $x$ can be upper-bounded as:
%$$\E |\Delta_L|\leqslant p|W_L|\cdot\ldots\cdot |W_1||x|$$
%\end{proposition}

We have now shown that even evaluating fault tolerance of a given network can be a hard problem. In order to make the analysis practical, we use additional assumptions based on the properties of neuromorphic hardware.

\section{Realistic Simplifying Assumptions for Neuromorphic Hardware}
\label{sec:realistic}
% https://drive.google.com/file/d/12hDkrN3l1sLPRW4jdenviVAMIeCBWcSf/view?usp=sharing
%\begin{figure}
    %\centering
    %\includegraphics[height=2cm]{figures/p_delta_taylor.pdf}
%    \includegraphics[height=2cm]{figures/tail_bound.pdf}
    %\caption{Left to right: (a) bounding the probability space $\Omega$ (b) A tail bound $\mathbb P\{X\geq \mathbb E X+\varepsilon\}\leq\delta$}
    %\label{fig:probability_space}
%\end{figure}

In this section, we introduce realistic simplifying assumptions grounded in neuromorphic hardware characteristics. We first show that if faults are not too frequent, the weight perturbation would be small. Inspired by this, we then apply a Taylor expansion to the study of the most probable case.
%See Figure \ref{fig:probability_space} for a graphical summary of the section.
\footnote{
%The idea to split the loss calculation into favourable and unfavourable cases can be seen in
The inspiration for splitting the loss calculation into favorable and unfavorable cases comes from
\cite{Nagarajan2019}}

%\subsection{Conditions of Small Weight Perturbation}
%In this section, we give sufficient conditions for which case the probability of large weight perturbation under a crash distribution is small.
%We note that this is not directly required to prove our main result. Rather, this is simpler guarantee which involves both techniques we will use later: sufficient condition for "sameness" of network's weights and the use of tail bounds.

%This assumption is made to simplify the calculations. Later we will show that our other assumptions lead to trivial error superposition.

\begin{assumption}
\label{assumption:psmall}
The probability of failure $p=\max\{p_l\big| l\in[L]\}$ is small: $p\lesssim 10^{-4}..10^{-3}$
\end{assumption}
%\andrei{Is it actually $\sim$ or would a $>$ work as well? The latter would be more general.}

This assumption is based on the properties of neuromorphic hardware \cite{schuman2017survey}. Next, we then use the internal structure of neural networks.

\begin{assumption}
\label{assumption:nbig}
The number of neurons at each layer $n_l$ is sufficiently big, $n_l\gtrsim 10^2$
\end{assumption}
This assumption comes from the properties of state-of-the-art networks \cite{amodei}.

%\andrei{From your presentation it seemed to me that it was more than a toy example but instead a proof by absurd that we could not do better than $\E\Delta=p$ and $\Var\Delta=p$ for $n$ going to infinity. If it actually is just a toy example, you might consider axing it to free up place.}
{\bf The best and the worst fault tolerance.} Consider a 1-layer NN with $n=n_0$ and $n_L=n_1=1$ at input $x_i=1$: $y(x)=\sum x_i/n$. We must divide $1/n$ to preserve $y(x)$ as $n$ grows. This is the most robust network, as all neurons are interchangeable. Here $\E\Delta=-p$ and $\Var\Delta=p/n$, variance decays with $n$. In contrast, the worst case $y(x)=x_1$ has all but one neuron unused. Therefore $\E\Delta=p$ and $\Var\Delta=p$, variance does not decay with $n$.

The next proposition shows that under a mild additional regularity assumption on the network, Assumptions \ref{assumption:psmall} and \ref{assumption:nbig} are sufficient to show that the perturbation of the norm of the weights is small.
\begin{proposition}
	\label{prop:usmall}
	Under A\ref{assumption:psmall},\ref{assumption:nbig} and if $\{W_l^i\}_{i=1}^{n_l}$ are $q$-balanced, for $\alpha>p$, the norm of the weight perturbation $U_l^i$ at layer $l$ is probabilistically bounded as:
	$\delta_0=\Prob\{\|U_l^{\cdot}\|_1\geq \alpha \|W_l^{\cdot}\|\}\leq \exp\left(-n_l\cdot q\cdot d_{KL}(\alpha||p_l)\right)$ with KL-divergence between numbers $a,b\in(0,1)$, $d_{KL}(a,b)=a\log{a}/{b}+(1-a)\log{(1-a)}/{(1-b)}$ and $W_l^i$ from D\ref{def:neuronfail}
\end{proposition}
Inspired by this result, next, we compute the error $\Delta$ given a small weight perturbation $U$ using a Taylor expansion. \footnote{In order to {\em certify} fault tolerance, we need a precise bounds on the remainder of the Taylor approximation. For example, for ReLU functions, Taylor approximation fails. The supplementary material contains another counter-example to the Taylor expansion of an NN. Instead, we give sufficient conditions for which the Taylor approximation indeed holds.}
%\andrei{Shouldn't that be a comment or a footnote once the bounds are derived? Once again, that allows to gain more space.}
%Next, we calculate the change in the output of the network given a small weight perturbation $U$. We will use the Taylor expansion as we know that the networks' derivatives decay with $n$ under our assumptions:
%Usually when it comes to a Taylor expansion, it is used with %saying that "the argument is small, and therefore higher-order terms can be neglected" (with {\em high} order usually meaning $2$ or $3$). However, we are trying to give a guarantee on robustness and therefore such an argument will not hold. For example, if we consider an activation function which is non-differentiable everywhere (such as ReLU), Taylor polynomial will fail to give the right answer for sufficiently large perturbations.

\newcommand{\Hmax}{H_{\max}}

\begin{table}[tb]
    \centering
    \begin{tabular}{llcl}\hline
    Quantity & Discrete &  & Continuous\\\hline
    Input & $x\colon [n_0]\to \mathbb R$ & \multirow{3}{*}{$\longmapsto$} & $x\colon [0,1]\to\R$\\

    Weights & $W_l\colon [n_l]\times[n_{l-1}]\to\R$ & &  $W_l\colon [0,1]^2\to \R$\\
    
    Pre-activations & $z_l^i=\sum_j W_l^{ij}y_{l-1}^i+b_l^i$ &  & $z_l(t)=\int_0^1 W_l(t,t')y_{l-1}(t')dt'+b_l(t)$\\\hline
    \end{tabular}
    \caption{Correspondence between discrete and continuous quantities. When an regular (discrete) NN is a function mapping vectors to vectors, a continuous NN is an operator mapping functions to functions}
    \label{tab:continuous_discrete_correspondence}
\end{table}

\begin{figure}[tb]
    \centering
    \includegraphics[width=0.9\textwidth]{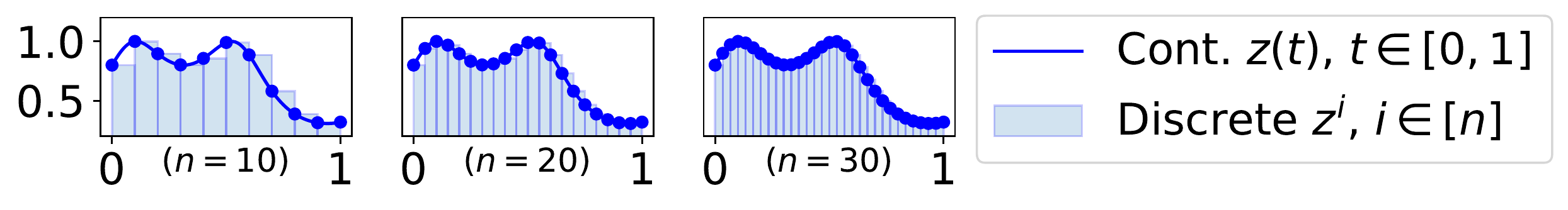}
    \caption{Discrete (standard) neural network approximating a continuous network}
    \label{fig:continuous_net}
\end{figure}

\begin{assumption}
\label{assumption:continuous_net}
As the width $n$ increases, networks $NN_n$ have a continuous limit \cite{sonoda2017double} $NN_n\to NN_c$, where $NN_c$ is a continuous neural network \cite{le2007continuous}, and $n=\min\{n_l\}$. That network $NN_c$ has globally bounded operator derivatives $D_k$ for orders $k=1,2$. We define $D_{12}=\max\{D_1,D_2\}$.\footnote{A necessary condition for $D_k$ to be bounded is to have a reasonable bound on the derivatives of the ground truth function $y^*(x)$. We assume that this function is sufficiently smooth.}
\end{assumption}
See Figure \ref{fig:continuous_net} for a visualization of A\ref{assumption:continuous_net} and Table \ref{tab:continuous_discrete_correspondence} for the description of A\ref{assumption:continuous_net}. The assumption means that with the increase of $n$, the network uses the same internal structure which just becomes more fine-grained. The continuous limit holds in the case of explicit duplication, convolutional networks and corresponding explicit regularization. The supplementary material contains a more complete explanation.

The derivative bound for order $2$ is in contrast to the worse-case spectral bound which would be exponential in depth as in Proposition \ref{th:spectral}. This is consistent with experimental studies \cite{ghorbani2019investigation} and can be connected to generalization properties via minima sharpness \cite{keskar2016large}.
%We will now use variables $\xi_l^i$ from the Definition \ref{def:neuronfail}. We consider what happens if we increase the layer width. %ince the neural network approximates the same function and has balanced neurons (Assumpt. \ref{assumption:w_qbalanced}), we assume that the magnitude of individual values at each layer stays the same. This implies that the derivative $\frac{\partial L}{\partial \xi_l^i}\sim\frac{1}{n_l}$ (see supplementary for more details).
\begin{proposition}
\label{proposition:decay}
Under A\ref{assumption:continuous_net}, derivatives are equal the operator derivatives of the continuous limit:
$$\frac{\partial^k y_L}{\partial y_l^{i^1}...\partial y_l^{i^k}}=\frac{1}{n_l^k}\frac{\delta^k y_L}{\delta y_l(i^1)...\delta y_l(i^k)}+o(1),\,n_l\to\infty$$
\end{proposition}
For example \footnote{The proposition 
is illustrated in proof-of-concept experiments with explicit regularization in the supplementary material. There are networks for which the conclusion of P\ref{proposition:decay} would not hold, for example, a network with $w^{ij}=1$. However, such a network does not approximate the same function as $n$ increases since $y(x)\to\infty$, violating A\ref{assumption:continuous_net}}, consider $y(x)=1/n_1\sum_{i_1=1}^{n_1}\varphi\left(\sum_{i_0=1}^{n_0} x^{i_0}/n_0\right)$ at $x_i\equiv 1$. Factors $1/n_0$ and $1/n_1$ appear because the network must represent the same $y^*$ as $n_0,n_1\to\infty$. Then, $\partial y/\partial x_i=\varphi'(1)/n_1$ and $\partial^2 y/\partial x_i\partial x_j=\varphi''(1)/n_1^2$.
%\andrei{I do not think that a proposition can be tested experimentally in theoretical paper in which the main contribution is claimed to be proved bounds. Illustrated would be a better word; I also believe that this sentence can be moved to a footnote.}
% {\bf Bounding $M_2$.} How to bound $H(x+\xi\delta x)$? Use CurvProp or something like (An Investigation into Neural Net Optimization via Hessian Eigenvalue Density). Use $w\sim 1$ by $q-balanced$. Or use generalization bound and not Taylor... Can do experimental study by computing gradient of $w^TH(w)w$ and doing GD. SOLUTION to $M_2$: first, assume $\E_x H$ has bounded $O(1)$ eigenvalues (confirmed by experiments other projects), then use integral Taylor formula to get $\E_x H$ and bound it via $\alpha^2$. How to make it easily computable if $\lambda_{\max}\ll w^THw$? Use transition $w\sim 1\cdot q$ and then connect $1^TH1$ to spectral norm somehow!... Assuming $\E|\delta x_i|^2\leq \alpha^2|w_i|^2$ (sum, then it's reasonable) and $\E|\delta x_i|\leq \alpha |w_i|$, Taylor remainder gives $\alpha^2w^THw$ for $H=H(x+\xi\delta x)$. That is Hessian-vector product! And it's computed in $O(n)$!
\begin{theorem}
\label{prop:taylor}
For crashes at layer $l$ and output error $\Delta_L$ at layer $L$ under A\ref{assumption:psmall}-\ref{assumption:continuous_net} with $q=1/n_l$ and $r=p+q$, the mean and variance of the error can be approximated as
$$
\E\Delta_L  =p_l\sum\limits_{i=1}^{n_l}\frac{\partial y_L}{\partial \xi^i}\Bigr|_{\xi=0}+\Theta_{\pm}(1)D_2 r^2,\,
\Var\Delta_L=p_l\sum\limits_{i=1}^{n_l}\left(\frac{\partial y_L}{\partial \xi^i}\Bigr|_{\xi=0}\right)^2+\Theta_{\pm}(1)D_{12}r^3\\
$$
By $\Theta_{\pm}(1)$ we denote any function taking values in $[-1,1]$.\footnote{The derivative $\partial y_L/\partial \xi^i(\xi)\equiv-\partial y_L(y_l-\xi\odot y_l)/\partial y_l^i\cdot y_l^i$ is interpreted as if $\xi^i$ was a real variable.}% We denote the first term in $\Var\Delta=p_lT_1+\theta_{\pm}(D_{12}r^3)$.
%\vspace{-10pt}
\end{theorem}
The full proof of the theorem is in the supplementary material.
%{\bf Remark.} Neuron Bernoulli crashes can be computed faster, in $\mathcal O(N)$ instead of $\mathcal O(N^2)$, see supplementary material
%{\bf Connection to the weight saliency.} The expression is somewhat similar to {\em saliency} $f''_{ww}w^2$ from \cite{lecun1990optimal} but uses first-order derivatives. Another similarity is from \cite{doi:10.1162/089976600300014782}.
%For Bernoulli crashes the condition $\|U\|\leq \alpha\|W\|$ does not hold for the whole probability space. However, even for Bernoulli case, it is true that that condition holds with {\em high probability}.
%This bound is not directly applicable to Bernoulli, only to limited Bernoulli with small probability of large perturbation. $\Var u\sim O(\alpha^2)$ but $\Var u\sim 
The remainder terms are small as both $p$ and $q={1}/{n_l}$ are small quantities under A\ref{assumption:psmall}-\ref{assumption:nbig}. In addition, P\ref{proposition:decay} implies $\partial y_L/\partial \xi^i\sim 1/n_l$ and thus, when $n_l\to\infty$, $\E\Delta=\mathcal O(1)$ remains constant, and $\Var\Delta_L=\mathcal O(1/n_l)$. This is the standard rate in case if we estimate the mean of a random variable by averaging over $n_l$ independent samples, and our previous example in the beginning of the Section shows that it is the best possible rate. Our result shows sufficient conditions under which neural networks allow for such a simplification.%\footnote{Namely, our framework shows that we can view neurons as independent "trials", each of them contributing equally to the "estimate".}
\footnote{However, the dependency $\Var\Delta\sim 1/n_l$ is only valid if $n< p^{-2}\sim 10^8$ to guarantee the first-order term to dominate, $p/n>r^3$. In case if this is not true, we can still render the network more robust by aggregating multiple copies with a mean, instead of adding more neurons. Our current guarantees thus work in case if $p^2\leq n^{-1}\leq p$. In the supplementary material, we show that a more tight remainder, depending only on $p/n$, hence decreasing with $n$, is possible. However, it complicates the equation as it requires $D_3$.}
In the next sections we use the obtained theoretical evaluation to develop a regularizer increasing fault tolerance, and say which architectures are more fault-tolerant. %\andrei{are we actually comparing the architectures or are we proposing the regularizer? The latter seems to be the more important contribution, highlighted both in the intro and the abstract}.
\section{Probabilistic Guarantees on Fault Tolerance Using Tail Bounds}
\label{sec:guarantee} In this section, we apply the results from the previous sections to obtain a probabilistic guarantee on fault tolerance. We identify which kinds of architectures are more fault-tolerant.% We conclude with a result showing how NNs are inherently fault tolerant \andrei{once again are those all the NNs or just specific ones? Trained with the proposed regularizer?}.% all of them, but the result if minor.

% {\bf Architecture and fault tolerance.} 
Under the assumptions of previous sections, the variance of the error decays as $\Var\Delta\sim \sum C_lp_l/n_l$ as the error superposition is linear (see supplementary material for a proof), with $C_l$ not dependent on $n_l$. Given a fixed budget of neurons, the most fault-tolerant NN has its layers balanced: one layer with too few neurons becomes a single point of failure. Specifically, an optimal architecture with a fixed sum $N=\sum n_l$ has $n_l\sim \sqrt{p_lC_l}$

Given the previous results, certifying $(\varepsilon,\delta)$-fault tolerance is trivial via a Chebyshev tail bound (proof in the supplementary material):

\begin{proposition}
\label{th:guarantee}
A neural network under assumptions \ref{assumption:psmall}-\ref{assumption:continuous_net} is $(\varepsilon,\delta)$-fault tolerant for $t=\varepsilon-\E\Delta_L>0$ with $\delta=t^{-2}\Var\Delta_L$ for $\E\Delta$ and $\Var\Delta$ calculated by Theorem \ref{prop:taylor}. %\ref{prop:usmall}.
\end{proposition}
%The next proposition shows that starting from sufficiently large $n$, the probability of bad loss would go down exponentially. However, we still use the Chebyshev version in our experiments since $L'w$ can take non-balanced values in general case.
%\begin{proposition}
%For a neural network with $(\alpha,\delta)$-guarantee of weight perturbation and for $U\sim D$ with $q_1$-balanced $|L'_iw_i|$ with $W,M_2,\alpha,\delta$ from the previous section
%$$
%\Prob\{\Delta\geq (1+\beta)L_i'\E U+\alpha^2\|W\|_1^2M_2\}\leq \exp(-2\beta^2q_1n)+\delta
%$$
%\end{proposition}
%\subsection{Expected fault tolerance over the training data}
%In this section we are interested in making $\Prob_{x,\xi}(\Delta\geq \Delta_0)\leq \delta$. The trivial idea to use a union bound does not work since number of inputs can be huge, or a system can even be presented with a new input at each time. We therefore utilize the Properties of fault tolerance on total variance and on expected fault tolerance (Proposition \ref{th:gd_zero}. The first one means that a network compensating for $\E\Delta$ is not required to be implemented in case if we only care about mean input. Moreover, this also means that previous analysis can be applied using formulas with $\E_x$ added. For example,
%$$
%\Var_{\xi,x}\Delta=\E_x((f'_i)^2,\Var u)+\Theta_{\pm}(\alpha^2\|W\|^2\E_xM_2)+\E_x\Theta_{\pm}(\E\Delta)^2
%$$
% {\bf Computational complexity of the bounds.} 
Evaluation of $\E\Delta$ or $\Var\Delta$ using Theorem \ref{prop:taylor} would take the same amount of time as one forward pass. However, the exact assessment would need $\mathcal O(2^n)$ forward passes by Proposition \ref{th:nphard}. 
In order to make the networks more fault tolerant, we now want to solve the problem
%\andrei{it is really unclear what citation is doing here. Was it formulated in that paper? If yes, maybe put it in the end of the phrase with something along the lines of "as previously formulated in (45)"}
of loss minimization under fault tolerance rather than ERM (as previously formulated in \cite{torres2017fault}):
$\inf_{(W,B)\in \FT} \mathcal L(w,B)$ where $\FT=\FT(L,\varphi,p,\varepsilon,\delta)$ from Definition \ref{def:faulttolerance}. Regularizing\footnote{We note that the gradient of $\Var\Delta$ is linear time-computable since it is a Hessian-vector product.} with Equation \ref{eq:regul} can be seen as an approximate solution to the problem above. Indeed, $\Var\Delta\approx p_l\sum_i\left(\frac{\partial L}{\partial y_l^i}\cdot y_l^i\right)^2$ (from T\ref{prop:taylor}) is connected to the target probability (P\ref{th:guarantee}). Moreover, the network is required to be continuous by A\ref{assumption:continuous_net}, which is achieved by making nearby neurons' weights close using a smoothing regularizing function $\mbox{smooth}(W)\approx\int |W'_{t}(t,t')|dtdt'$. The $\mu$ term for $q$-balancedness comes from P\ref{prop:usmall} as it is a necessary condition for A\ref{assumption:continuous_net}. See the supplementary material for complete details. Here $\hat{\mathcal L}$ is the regularized loss, $\mathcal L$ the original one, and $\lambda,\,\mu,\,\nu,\,\psi$ are the parameters:
\begin{equation}
\label{eq:regul}
\hat{\mathcal L}(W)=\mathcal L(W)+\lambda\sum\limits_{i=1}^{n_l}\left(\frac{\partial \mathcal L}{\partial y^i}\cdot y^i\right)^2+\mu\left(\frac{\max_i W^{i}_l}{\min_i W^{i}_l}\right)^2+\psi\cdot\mbox{smooth}(W_l)+\nu \|W\|_{\infty}
\end{equation}
% {\bf The redundancy and the Median Trick} 
We define the terms corresponding to $\lambda, \mu,\psi$ as $R_1\approx\Var\Delta/p_l$, $R_2=q^2$, $R_3=\mbox{smooth}(W_l)$. If we have achieved $\delta<1/3$ by P\ref{th:guarantee}, we can apply the well-known {\em median trick} technique~\cite{niemiro2009fixed}, drastically increasing fault tolerance. We only use $R$ repetitions of the network with component-wise median aggregation to obtain $(\varepsilon,\delta\cdot\exp(-R))$-fault tolerance guarantee. See supplementary material for the calculations.
In addition, we show that after training, when $\E_x\nabla_W y_{L+1}(x)=0$, then $\E_x\E_\xi\Delta_{L+1}=0+\mathcal{O}(r^2)$ (proof in the supplementary material).
This result sheds some light on why neural networks are inherently fault-tolerant in a sense that the mean $\Delta_{L+1}$ is $0$.
%Under our assumptions, error superposition is linear (proof in the supplementary), so there is no need to consider more than one layer in the analysis.
%\todo[color=red]{Does superposition hold with new Taylor bounds?}
%\begin{proposition}
%\label{th:lin}
%A network $(L,W,B,\varphi)$ with crashes with probability $p_l$, $l\in\overline{0,L}$ has:
%$$
%\arraycolsep=0.6pt
%\begin{array}{rlrllrll}
%\E\Delta_L^{p_0,...,p_L}  &=&\E&\Delta_L^{p_0}  &+\ldots+&\E&\Delta_L^{p_L}  &+\mathcal{O}(\alpha^2)\\
%\Var\Delta_L^{p_0,...,p_L}&=&\Var&\Delta_L^{p_0}&+\ldots+&\Var&\Delta_L^{p_L}&+\mathcal{O}(\alpha^2)
%\end{array}
%$$
%\end{proposition}
%{\bf The loss or the output.} All previous theorems also apply for $\Delta_{L+1}$ since a loss function can be seen as adding another differentiable layer.
% {\bf Convolutional architectures.}
Convolutional networks of architecture {\tt Conv-Activation-Pool} can be seen as a sub-type of fully connected ones, as they just have locally-connected matrices $W_l$, and therefore our techniques still apply. Using large kernel sizes (see supplementary material for discussion), smooth pooling and activations lead to a better approximation. % Since a convolutional network with configuration Conv+MaxPool+ReLU can be seen as a special case of a fully-connected network, the bounds from this paper still apply. We note that using average pooling instead of max pooling and differentiable activation function will make the network analyzable using our techniques.

We developed techniques to assess fault tolerance and to improve it. Now we combine all the results into a single algorithm to certify fault tolerance.
\section{An algorithm for Certifying Fault Tolerance}
\label{sec:algorithm}
\begin{algorithm}[tb]
 \KwData{Dataset $D$, input-output point $(x,y^*)$, failure probabilities $p_l$, depth $L$, activation function $\varphi\in C^{\infty}$, target $\varepsilon$ and $\delta'$, the error tolerance parameters from the Definition \ref{def:faulttolerance}, maximal complexity guess $C\approx \int |y_l'(t)|dt\approx R_3^{guess}$
 
 }
 \KwResult{An architecture with $(\varepsilon,\delta')$-fault-tolerance on $x$}
 Select initial width $N=(n_1,...,n_{L-1})$\;
 \While{true}{
  Train a network to obtain $W,B$\;
  Compute $q$ from Proposition \ref{prop:usmall}\;
  {\bf If} $q<10^{-2}$, increase regularization parameter $\mu$ from Eq. \ref{eq:regul}, {\bf continue};\,{\em // go to line 3}\;
  Compute $\delta_0$ from Proposition \ref{prop:usmall} using $q$\;
  {{\bf If} $\delta_0>1/3$, increase $n$ by a constant amount, {\bf continue}}\;
  Compute $R_3$ from Eq. \ref{eq:regul}\;
  {{\bf If} $R_3> C$, increase regularization parameter $\psi$ from Eq. \ref{eq:regul}, {\bf continue}}\;
  Compute $\E\Delta$ and $\Var\Delta$ from Theorem \ref{prop:taylor}\;
  {\bf If} $\E\Delta>\varepsilon$, {\bf output} infeasible;\,{\em // cannot do better than the mean}\footnotemark\;
  Compute $\delta$ from Proposition \ref{th:guarantee}\;
  {{\bf If} $\delta>1/3$}, increase $n$ by a constant amount and increase $\lambda$ in Eq. \ref{eq:regul}, {\bf continue}\;
  Compute $R=\mathcal O(\log\frac{1}{\delta'})$\;
  {\bf Output} number of repetitions $R$, layer widths $N$, parameters $W,B$\;
 }
 \caption{Achieving fault tolerance after training. The numbers $q_{\max}=10^{-2}$ and $\delta_{\max}=1/3$ are chosen for simplicity of proofs. The asymptotic behavior does not change with different numbers, as long as $\delta_{\max}<1/2$ and the constraints on $q$ mentioned in the supplementary material are met}
 \label{algo:ft}
\end{algorithm}
\footnotetext{More neurons do not solve the problem, as $\E\Delta$ stays constant with the growth of $n$ by Theorem \ref{prop:taylor}. Intuitively, this is due to the fact that if a mean of a random variable is too high, more repetitions do not make the estimate lower.}

We are now in the position to provide an algorithm (Algorithm \ref{algo:ft}) allowing to reach the desired $(\varepsilon,\delta)$-fault tolerance via training with our regularizer and then physically duplicating the network a logarithmic amount of times in hardware, assuming independent faults. We note that our algorithm works for a single input $x$ but is easily extensible if the expressions in Propositions are replaced with expectations over inputs (see supplementary material).

% {\bf Required number of neurons.}
In order to estimate the required number of neurons, we use  bounds from T\ref{prop:taylor} and P\ref{th:guarantee} which require $n\sim p/\varepsilon^2$. However, using the median approach allows for a fast exponential decrease in failure probability. Once the threshold of failing with probability $1/3$ is reached by P\ref{th:guarantee}, it becomes easy to reach {\em any} required guarantee. The time complexity (compared to the one of training) of the algorithm is $\mathcal O(D_{12}+C_lp_l/\varepsilon^2)$ and space complexity is equal to that of one training call. See supplementary material for the proofs of resource requirements and correctness.

%\todo[color=red]{Add howto algorithm with precise calculations}
%\begin{enumerate}
    %\item Establish probability of crash $p$, tolerable probability of failure $q$ and tolerable loss $L=L$
    %\item Ensure that not too many neurons crash at once, select $\alpha$
    %\item Compute via Chernoff/Chebyshev probability of bad loss, ensure that it is less than $0.5$
    %\item Repeat $k=-\log q$ times
%\end{enumerate}

\section{Experimental Evaluation}
%In this section we evaluate our algorithm in a proof-of-concept setting.
\label{sec:experiments}

%\todo[inline]{Do and report actually impotant experiments}

%Code can be found at
%\begin{center}
%	\href{https://github.com/LPD-EPFL/FatalBrainDamage}{github.com/LPD-EPFL/FatalBrainDamage}
%\end{center}
In this section, we test the theory developed in previous sections in proof-of-concept experiments. We first show that we can correctly estimate the first moments of the fault tolerance using T\ref{prop:taylor} for small (10-50 neurons) and larger networks (VGG). We test the predictions of our theory such as decay of $\Var\Delta$, the effect of our regularizer and the guarantee from Algorithm \ref{algo:ft}. See the supplementary material for the technical details where we validate the assumption of derivative decay (A\ref{assumption:continuous_net}) explicitly. Our code is provided at the address {\tt \href{https://github.com/LPD-EPFL/ProbabilisticFaultToleranceNNs}{github.com/LPD-EPFL/ProbabilisticFaultToleranceNNs}}.%we show that our tail bounds result in realistic parameters for networks, allowing for a cost-effective neuromorphic implementation.
%{\bf Motivation.} Our bounds are stated for a general neural network, and therefore they are applicable for larger architectures. One might wonder in which conditions they actually work for larger networks. We note however that this would be out of the scope of one single paper because in the current work we focus on developing the bounds and testing them on proof-of-concept setups. We note that applying our bounds to larger networks and bigger datasets and establishing the region where they are applicable would be an interesting future research direction.
%We tested error superposition and proposition \ref{th:error_mean_eq} and verified them.
%\subsection{Predicting first moments of $\Delta$}
%\label{sec:error_prediction}

{\bf Increasing training dropout.} We train sigmoid networks with $N\sim 100$ on MNIST (see {\tt ComparisonIncreasingDropoutMNIST.ipynb}). We use probabilities of failure at inference and training stages $p_i=0.05$ at the first layer and $10$ values of $p_t\in[0,1.2p_i]$. The experiment is repeated $10$ times. When estimating the error experimentally, we choose $6$ repetitions of the training dataset to ensure that the variance of the estimate is low. The results are in the Table \ref{tab:comparison_metrics}. The experiments show that crashing MAE (Mean Absolute Error for the network with crashes at inference) is most dramatically affected by dropout. Specifically, training with $p_t\sim p_i$ makes network more robust at inference, which was well-established before. Moreover, the bound from T\ref{prop:taylor} can correctly order which network is trained with bigger dropout parameter with only $4\%$ rank loss, which is the fraction of incorrectly ordered pairs. All other quantities, including norms of the weights, are not able to order networks correctly. See supplementary material for a complete list of metrics in the experiment.
%\todo[color=red]{Test tail bound and moment bound on small and big networks, do rank loss test}
%\todo{Describe how b1 and b2 fail}
%\todo[color=red]{Divide variance by $n$ to calculate sample mean variance instead of variance.}
%\todo{Test quantized networks $w=\pm 1$ because they are $q=1$-balanced!}
%\todo{Test on CIFAR-10 and w/o skip connections and batch norm net}
%\todo{Unroll an RNN (might be simple than CNN)}
%\todo[color=red]{Compare to existing FT increase techniques \cite{torres2017fault}}
\begin{figure}
\centering
%\qquad
%\subfloat[Accuracy and error as function of the regularization parameter][Accuracy and error as function of the regularization parameter $\lambda$]{
%\includegraphics[width=0.3\textwidth]{figures/comparison_acc_do_reg_mnist.pdf}
%\label{fig:subfig3}}
%\qquad
%\subfloat[MNIST width and fault tolerance NNs][Layer width and fault tolerance for MNIST-trained NNs]{
\includegraphics[width=0.9\textwidth]{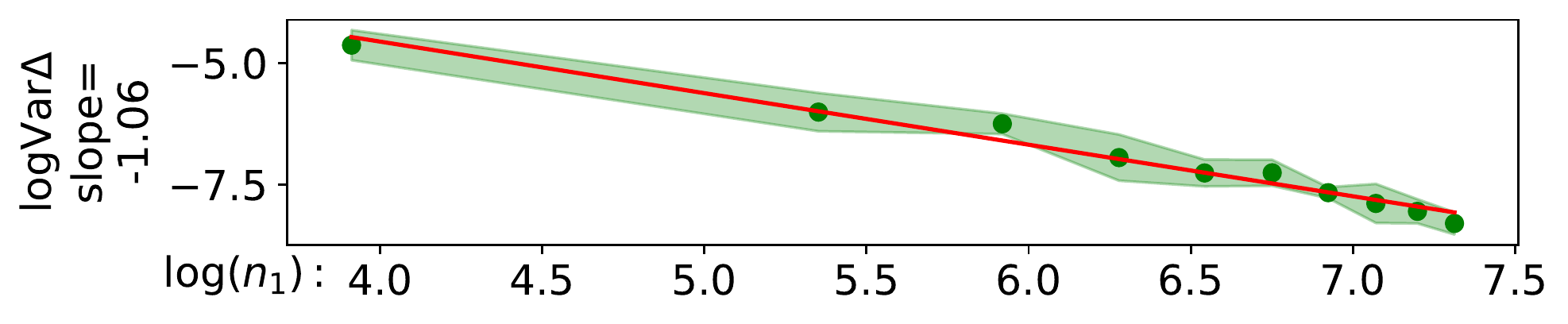}
% I have no idea what protect does but it works with it...
% https://tex.stackexchange.com/questions/228973/argument-of-captionydblarg-has-an-extra
% Here's the explanation: https://tex.stackexchange.com/questions/4736/what-is-the-difference-between-fragile-and-robust-commands
%\protect\subref{fig:subfig4}: 
\caption{The effect of the layer width $n_l$, $l=1$, horizontal axis on the variance of the fault tolerance error $\Var\Delta$, vertical axis}
%\label{fig:comp_do}
\label{fig:comparison_do}
\end{figure}

\begin{table}[tb]
    \centering
\subfloat[Experimental metrics]{
\begin{tabular}{llr}
\toprule
Quantity & Train rank loss & Test rank loss\\
\midrule
Crashing, MAE &      {\bf 5.6\%}  & {\bf 5.6\%}\\
Crashing, Accuracy &      19.8\%  & 17.7\%\\
Correct, MAE &      23.3\% & 22.0\% \\
Correct, Accuracy &      31.7\%  & 39.9\%\\
\bottomrule
\end{tabular}
}
\qquad
\subfloat[Theoretical bounds]{\begin{tabular}{lr}
\toprule
Quantity & Rank Loss \\
\midrule
T\ref{prop:taylor} $\Var\Delta$ &     {\bf 3.6\%} \\
P\ref{th:spectral} $\E\Delta$ &    24.8\% \\
P\ref{th:spectral} $\Var\Delta$ &     31.7\% \\
T\ref{prop:taylor} $\E\Delta$ &    40.8\% \\
\bottomrule
\end{tabular}}
    \caption[Test]{Comparison of networks trained with increased dropout. Rank loss between $p_{train}$ and various metrics. Experiment shows crashing and correct networks, MAE or accuracy and test/train datasets. Theoretical bounds include P\ref{th:spectral}\footnotemark and T\ref{prop:taylor}.}
    % \andrei{Ok, a couple of remarks. First, why the $P5Var\Delta$ has so many more experiments? Second, I would create a rank loss and rank loss test/train columns, as well as the one for theoretical prediction just to make it more clear, given you have width space, if you are going with table. I still believe however that a boxplot/violinplot/point plot with a trait for theoretical prediction, train/test pairing by proximity and correct/crash coding by color (red/green) could be more interesting}
    \label{tab:comparison_metrics}
\end{table}
\footnotetext{Variance for P\ref{th:spectral} is derived in the supplementary material}

{\bf Regularization for fault tolerance}. Previously, our bound is demonstrated to be able to  correctly predict which network is more resilient. We therefore use it as a regularization technique suggested by Eq. \ref{eq:regul}, see {\tt Regularization.ipynb}. We establish that the resilience of the network regularized with Dropout is similar to that of a network regularized with the bound%, see Figure~\ref{fig:subfig3}.
%Variance bound is used for regularization instead of the exact expression which would replace dropout provided by Proposition~\ref{th:dropout_interpretation}. Even if $\Var\Delta$ is not exact expression, the resulting fault tolerance is competitive still. Both dropout and our bound use similar probabilistic model which could explain this result.

{\bf Testing the bound on larger networks.} We test the bound on VGG16 and on a smaller convnet, see {\tt ConvNetTest-MNIST.ipynb} and {\tt ConvNetTest-VGG16.ipynb} and verify that they correctly predict the magnitude of the error

%\begin{figure}
    %\centering
    %\includegraphics[width=0.3\textwidth]{figures/ae_singlepixel_mnist.png}
    %\caption{Single pixel attack success and Dropout}
    %\label{fig:ae_sp}
%\end{figure}

%Figure \ref{fig:width_ft} shows how layer width affects fault tolerance. The dependence fits the predicted law of $\Var\sim 1/n$.

%\subsection{Architecture and fault tolerance}
{\bf Architecture and fault tolerance.} Comparing different architectures on a single image with $p=0.01$ (VGG16, VGG19, MobileNet) shows (and {\tt ConvNetTest-ft.ipynb}) that the bigger the mean width of the layer (approximated by the number of parameters), the better is the fault tolerance, as predicted in Section \ref{sec:guarantee}. In addition, training networks on the MNIST dataset (see {\tt FaultTolerance-Continuity-FC-MNIST.ipynb}) shows a decrease in variance with $n_l$ as predicted by Theorem \ref{prop:taylor}, see Figure \ref{fig:comparison_do}: the variance decays as $1/n_l$. We regularize with $\psi=(10^{-4}, 10^{-2})$ for derivatives and smoothing respectively (see supplementary material for explanation of coefficients) and $\lambda=0.001$.

%\todo[inline]{What will happen w/o regularization? For differenct coefficients?}

{\bf Testing the algorithm.} We test the Algorithm 1 on the MNIST dataset for $\varepsilon=9\cdot 10^{-3}$, $\delta=10^{-5}$ and obtain $R=20$, $n_1=500$, $\lambda=10^{-6}$, $\mu=10^{-10}$, $\psi=(10^{-4}, 10^{-2})$. We evaluate the tail bound experimentally. Our experiment demonstrates the guarantee given by Proposition \ref{th:guarantee} and can be seen as an experimental confirmation of the algorithm's correctness. See {\tt TheAlgorithm.ipynb}.%\andrei{I believe that statement actually would deserve a figure to illustrate it and show networks regularized with our method indeed work better.}

We hence conclude that our proof-of-concept experiments show an overall validity of our assumptions and of our approach.

%\subsection{Adversarial examples and fault tolerance}
%FGSM method results in many components changed, and regularization with Dropout does not help, see {\tt AE\_fgsm\_mnist.ipynb}. However, our method helps for single pixel attack, see {\tt AE\_onepixel\_mnist.ipynb}

\section{Conclusion}

\label{sec:conclusion}
Fault tolerance is an important overlooked concrete AI safety issue \cite{amodei2016concrete}. 
% \andrei{explain what you at is better than before, don't diss the existing research given that's the people who will likely review you}
%that current approaches to this problem lack formal correctness.
This paper describes a probabilistic fault tolerance framework for NNs that allows to get around the NP-hardness of the problem. Since the crash probability in neuromorphic hardware is low, we can simplify the problem to allow for a polynomial computation time. We use the tail bounds to motivate the assumption that the weight perturbation is small. This allows us to use a Taylor expansion to compute the error. To bound the remainder, we require sufficient smoothness of the network, for which we use the continuous limit: nearby neurons compute similar things. After we transform the expansion into a tail bound to give a bound on the loss of the network. This gives a probabilistic guarantee of fault tolerance. Using the framework, we are able to guarantee sufficient fault tolerance of a neural network given parameters of the crash distribution. We then analyze the obtained expressions to compare fault tolerance between architectures and optimize for fault tolerance of one architecture. We test our findings experimentally on small networks (MNIST) as well as on larger ones (VGG-16, MobileNet). Using our framework, one is able to deploy safer networks into neuromorphic hardware.%In terms of neuromorphic hardware, our approach allows to predict how a network behaves under crashes. In addition, we can make the network more resilient during training. Overall, equipped with our theory, one can deploy fault tolerant networks into neuromorphic hardware under non-restrictive assumptions.

Mathematically, the problem that we consider is connected to the problem of generalization \cite{Neyshabur2017,Nagarajan2019} since the latter also considers the expected loss change under a small random perturbation $\E_{W+U}\mathcal L(W+U)-\mathcal L(W)$, except that these papers consider Gaussian noise and we consider Bernoulli noise. Evidence \cite{phatak1999relationship}, however, shows that sometimes networks that generalize well are not necessarily fault-tolerant. Since the tools we develop for the study of fault tolerance could as well be applied in the context of generalization, they could be used to clarify this matter.

\subsubsection*{Acknowledgements}
We thank Michael Kapralov, Martin Jaggi, Manuel Le Gallo, Arthur Jacot, Clement Hongler for helpful discussions. %
We thank the anonymous ICML and NeurIPS reviewers for constructive criticism of earlier versions of this work.

\end{document}

\maketitle
%\linenumbers

%\listoftodos
%\tableofcontents

\section{Introduction}

First we prove all the propositions (labeled Proposition 1, 2, ...) from the main paper. Their names and sections match the main paper. We also give additional results, they are labelled Additional Proposition 1, 2, etc. If they are formal statements of the results referred to in the main paper, they are in the same section as the reference. We abbreviate Assumption 1 $\to$ A1, Proposition $1$ $\to$ P1, Theorem 1 $\to$ T1, Definition 1 $\to$ D1. 

\begin{flushright}
\begin{sidenote}
\paragraph{} Less precise statements with possible future research directions on fundamental questions required to make the guarantee even more strong are flushed right.
\end{sidenote}
\end{flushright}
%\todo[inline]{Remove additional propositions, make them mains}

\section{Definition of Probabilistic Fault Tolerance}
\label{sec:theory}
{\bf Notations.} For any two vectors $x,y\in\mathbb{R}^n$ we use the notation $(x,y)=\sum_{i=1}^n x_iy_i$ for the standard scalar product.
%The notation $|W|$ means element-wise absolute values of the matrix: $(|W|)_{ij}=|W_{ij}|$.
Matrix $\gamma$-norm for $\gamma=(0,+\infty]$ is defined as $\|A\|_{\gamma}=\sup_{x\neq 0}\|Ax\|_{\gamma}/\|x\|_{\gamma}$. We use the infinity norm $\|x\|_{\infty}=\max|x_i|$ and the corresponding operator matrix norm.
We call a vector $0\neq x\in\mathbb{R}^n$ $q$-balanced if $\min |x_i|\geq q \max |x_i|$. We denote $[n]=\{1,2,...,n\}$. We define the Hessian $H_{ij}={\partial^2 y(x)}/{\partial x^i\partial x^j}$ as a matrix of second derivatives. We write layer indices down and element indices up: $W_l^{ij}$. For the input, we write $x_i\equiv x^i$. If the layer is fixed, we omit its index. We use the element-wise Hadamard product $(x\odot y)_i=x_iy_i$.

\begin{definition}{(Neural network)}
    \label{def:nn}
	A neural network with $L$ layers is a function $y_L\colon \mathbb{R}^{n_0}\to\mathbb{R}^{n_L}$ defined by a tuple $(L,W,B,\varphi)$ with a tuple of weight matrices $W=(W_1,...,W_L)$ (or their distributions) of size $W_l\colon n_l\times n_{l-1}$, biases $B=(b_1,...,b_L)$ (or their distributions) of size $b_l\in \mathbb{R}^{n_l}$ by the expression
	$y_l=\varphi(z_l)$ with pre-activations $z_l=W_ly_{l-1}+b_l,\,l\in[L]$, $y_0=x$ and $y_L=z_L$. Note that the last layer is linear. We additionally require $\varphi$ to be 1-Lipschitz \footnote{1-Lipschitz $\varphi$ s.t. $|\varphi(x)-\varphi(y)|\leqslant |x-y|$. If $\varphi$ is $K$-Lipschitz, we rescale the weights to make $K=1$: $W^{ij}_l\to W^{ij}_l/K$. This is the general case. Indeed, if we rescale $\varphi(x)\to K\varphi(x)$, then, $y_{l-1}\to K y_{l-1}'$, and in the sum $z_l'=\sum W^{ij}/K\cdot K y_{l-1}\equiv z_l$}.
	We assume that the network was trained using input-output pairs $x,y^*\sim X\times Y$ using ERM\footnote{Empirical Risk Minimization -- the standard task $1/k\sum_{k=1}^m\omega(y_L(x_k), y^*_k)\to\min$} for a loss $\omega$.
    Loss layer for input $x$ and the true label $y^*(x)$ is defined as $y_{L+1}(x)=\E_{y^*\sim Y|x}\omega(y_L(x),y^*))$ with $\omega\in[-1,1]$\footnote{The loss is bounded for the proof of Algorithm 1's running time to work}
\end{definition}

\begin{definition}{(Weight failure)}
    \label{def:weightfail}
	Network $(L,W,B,\varphi)$ with weight failures $U$ of distribution $U\sim D|(x,W)$ is the network $(L,W+U,B,\varphi)$ for $U\sim D|(x,W)$. We denote a (random) output of this network as $y^{W+U}(x)=\hat{y}_L(x)$ with activations $\hat{y}_l$ and pre-activations $\hat{z}_l$, as in D\ref{def:nn}. %We sometimes write $u$ and $w$ meaning matrices $u$
\end{definition}

\begin{definition}{(Bernoulli neuron failures)}
    \label{def:neuronfail}
    Bernoulli neuron crash distribution is the distribution with i.i.d. $\xi_l^i\sim \Be(p_l)$, $U_l^{ij}=-\xi_l^i\cdot W_l^{ij}$. For each possible crashing neuron $i$ at layer $l$ we define $U_l^i=\sum_{j}|U_l^{ij}|$ and $W_l^i=\sum_j |W_l^{ij}|$, the crashed incoming weights and total incoming weights. We note that we see neuron failure as a sub-type of weight failure.
\end{definition}

\begin{definition}{(Output error for a weight distribution)}
    \label{def:outputfail}
	The error in case of weight failure with distribution $D|(x,W)$ is $\Delta_l(x)=y^{W+U}_l(x)-y^{W}_l(x)$ for layers $l\in[L+1]$
\end{definition}
We extend the definition of $\varepsilon$-fault tolerance from \cite{whenneuronsfail17} to the probabilistic case:

\begin{definition}{(Probabilistic fault tolerance)}
\label{def:faulttolerance}
A network $(L,W,B,\varphi)$ is said to be $(\varepsilon,\delta)$-fault tolerant over an input distribution $(x,y^*)\sim X\times Y$ and a crash distribution $U\sim D|(x,W)$ if
$
\mathbb P_{(x,y^*)\sim X\times Y,\,U\sim D|(x,W)}\{\Delta_{L+1}(x)\geq \varepsilon\}\leq\delta
$.
For such network, we write $(W,B)\in \FT(L,\varphi,p,\varepsilon,\delta)$.
\end{definition}

\section{The Hardness of The Fault Tolerance}
\label{sec:hardness}

\subsection{NP-Hardness}
\label{sec:nphardness}

\begin{mainprop}
	\label{th:nphard}
	The task of evaluating $\E\Delta^k$ for any $k=1,2,...$ with constant additive or multiplicative error for a neural network with $\varphi\in C^\infty$, Bernoulli neuron crashes and a constant number of layers is NP-hard.
\end{mainprop}
\begin{proof}
To prove that a problem is NP-hard, it suffices to take another NP-hard problem, and reduce any instance of that problem to our problem, meaning that solving our problem would solve the original one.

We take the NP-hard Subset Sum problem. It states: given a finite set of integers $x_i\in\mathbb{Z}$, $i\in[M]$, determine if there exist a non-empty subset $S\subseteq[M]$ such that $\sum\limits_{x\in S}x_i=0$.

We take a subset sum instance $x_i\in\mathbb{Z}$ and feed it as an input to a neural network with two first layer neurons with $\varphi$ being a piecewise-linear function from $0$ to $1$ at points $-\varepsilon$ and $\varepsilon$ for some fixed $\varepsilon\in(0,1)$. Note that in this proof we only compute $\varphi$ at integer points, and therefore it is possible re-define $\varphi\in C^\infty$ such that it has same values in natural points.

Note that inputs to it are integers, therefore the outputs are $1$ if and only if (iff) the sum is greater than zero. First neuron has coefficients all $1$s and second all $-1$s with no bias. The next neuron has coefficients $1$ and $1$ for both inputs, a bias $-1.5$ and threshold activation function, outputting $1$ only if both inputs are $1$. Again, since inputs to this neuron are integers, we can re-define $\varphi$ to be $C^{\infty}$. The final neuron is a (linear) identity function to satisfy Definition \ref{def:nn}. It takes the previous neuron as its only input. Now, we see that $y(x)=1$ if and only if the sum of inputs to a network is $0$.

We now feed the entire set $S$ to the network as its input $x$. In case if $y(x)=1$ (which is easy to check), we have arrived at a solution, as the whole set has zero sum. In the following we consider the case when $y(x)=0$.

Suppose that there exist an algorithm calculating answer to the question $\E\Delta^k>z$ for any finite-precision $z$. Now, the expectation has terms inside $\E\Delta^k=\sum\limits_{s\in S}p^{|s|}(1-p)^{1-|s|}y^k(x\odot s)$. Suppose that one of the terms is non-zero. Then $y(x\odot s)\neq 0$, which means that the threshold neuron outputs a value $>0$ which means that sum of its inputs is greater than $1.5$. This can only happen if they are both $1$, which means that sum for a particular subset is both $\geq 0$ and $\leq 0$. By setting $z=0$ we solve the subset sum problem using one call to an algorithm determining if $\E\Delta^k>0$. Indeed, in case if the original problem has no solution, the algorithm will output $\E\Delta^k\leq 0$ as there are no non-zero terms inside the sum. In case if there is a solution, there will be a non-zero term in the sum and thus $\E\Delta^k>0$ (all terms are non-negative).

Now, suppose there exist an algorithm which always outputs an additive approximation to $\E\Delta^k$ giving numbers $\mu$ and $\varepsilon$ such that $\E\Delta^k\in (\mu-\varepsilon,\mu+\varepsilon)$ for some constant $\varepsilon$. Take a network from previous example with additional layer scaling output by $C=\frac{2\varepsilon}{p^M}$. Take a subset sum instance $x_i$. Output YES iff $0\notin (\mu-\varepsilon,\mu+\varepsilon)$. This is correct because if $0\notin(\mu-\varepsilon,\mu+\varepsilon)$, then it must be that $\E\Delta^k>0$ and vice versa. Multiplicative approximation has an analogous proof where we scale the outputs even more.
\end{proof}

We note that computing the distribution of $\Delta$ for one neuron, binary input and weights and a threshold activation function is known as {\em noise sensitivity} in theoretical Computer Science. There exists an exact assessment for $W_1^{1i}\equiv w^i=1$\footnote{As we only have one neuron, the index is one-dimensional}, however for the case $w^i\neq 1$ the exact distribution is unknown \cite{benjamini1999noise}.

\subsection{Pessimistic Spectral Bounds}
\label{sec:spectral}

\begin{proposition}[Norm bound or bound \boundfirst{}]
\label{th:bound_v1}
For any norm $\|\cdot\|$ the error $\Delta_L$ at the last layer on input $x$ and failures in input can be upper-bounded as (for $\xi\odot x$ being the crashed input)
$$
\|\Delta_L\|\leqslant \|W_L\|\cdot\ldots\cdot\|W_1\|\|\xi\odot x - x\|
$$
\end{proposition}

\begin{proof}
We assume a faulty input (crashes at layer $0$). By Definition~\ref{def:outputfail} the error at the last layer is $\Delta_L=\hat{y}_L-y_L$. By definition~\ref{def:nn}, $y_L=W_Ly_{L-1}+b_L$ and by Definition~\ref{def:neuronfail}, $\hat{y}_L=W_L\hat{y}_{L-1}+b_L$. Thus, $\|\Delta_L\|=\|W_L\hat{y}_{L-1}+b_L-W_Ly_{L-1}-b_L\|=\|W_L(\hat{y}_{L-1}-y_{L-1})\|\leq \|W_L\|\|\hat{y}_{L-1}-y_{L-1}\|$.

Next, since $\hat{y}_{L-1}=\varphi(W_{L-1}\hat{y}_{L-2}+b_{L-1})$ and $y_{L-1}=\varphi(W_{L-1}y_{L-2}+b_{L-1})$, we use the 1-Lipschitzness of $\varphi$. In particular, we use that for two vectors $x$ and $y$ and $\varphi(x),\varphi(y)$ applied element-wise, we have $\|\varphi(x)-\varphi(y)\|\leq\|x-y\|$ because the absolute difference in each component is less on the left-hand side. We thus get $\|\hat{y}_{L-1}-y_{L-1}\|\leq \|W_{L-1}\hat{y}_{L-2}+b_{L-1}-W_{L-1}\hat{y}_{L-2}-b_{L-1}\|=\|W_{L-1}(\hat{y}_{L-2}-y_{L-2})\|\leq \|W_{L-1}\|\|\hat{y}_{L-2}-y_{L-2}\|$. Plugging it in to the previous equation, we have $\|\Delta_L\|\leq\|W_L\|\cdot\|W_{L-1}\|\|\hat{y}_{L-2}-y_{L-2}\|$

Now, since the inner layer act in a similar manner, we inductively apply the same argument for underlying layers and get
$$
\|\Delta_L\|\leq\|W_L\|\cdot\ldots\cdot\|W_1\|\|\hat{x}-x\|
$$

Moreover we have failing input, thus $\hat{x}=\xi\odot x$ which completes the proof.
\end{proof}

\begin{mainprop}[$K$ using spectral properties \cite{Neyshabur2017}]
	\label{th:spectral}
	$
	\|y_L(x_2)-y_L(x_1)\|_2\leqslant  \|x_2-x_1\|\cdot \prod_{l=1}^L\|W_l\|_2
	$
\end{mainprop}
\begin{proof}
Application of AP\ref{th:bound_v1} for $\|\cdot\|=\|\cdot\|_2$ and $x_2=\xi\odot x$
\end{proof}

We see, there are bounds considering different norms of matrices $\|A\|_{\gamma}$ (AP\ref{th:bound_v1}) or using the triangle inequality and absolute values (AP\ref{th:bound_v2}). They still lead to pessimistic estimates. All these bounds are known to be loose \cite{Nagarajan2019} due to $\|W\|=\sup_{x\neq 0}\|Wx\|/\|x\|$ being larger than the input-specific $\|Wx\|/\|x\|$. We will circumvent this issue by considering the {\em average} case instead of the worst case.

\begin{corollary}[Infinity norm, connecting \cite{ontherobustness17} to norm bounds]
For an input $x$ with $C=\|x\|_\infty$ for failures at the input with $\mathcal O(1)$ dependent only on layer size, but not on the weights,
$$\|\Delta_L\|_\infty\leqslant pC\|W_L\|_\infty\cdot\ldots\cdot\|W_1\|_\infty\cdot \mathcal O(1)+\mathcal O(p^2)$$
Here $\|x\|_{\infty}=\max|x_i|$
\end{corollary}

\begin{proof}
First we examine the expression (4) from the other paper~\cite{ontherobustness17} and show that it is equivalent to the result we are proving now:
$$
Erf=\E\|\Delta\|_\infty\leq \sum\limits_{l=1}^L C_l f_l K^{L-l}w_m^{(L)}\prod\limits_{l'=l+1}^L(N_{l'}-f_{l'})w_m^{(l')}
$$
here we have $C_l$ the maximal value at layer $l$, $K$ the Lipschitz constant, $w_m$ is the maximal over output neurons (rows of $W$) and mean absolute value over input neurons (columns of $W$) weight, $f_l$ is the number of crashed neurons.

Now we set $f_1=pN_1$ and $f_i=0$ for $i>1$ and moreover we assume $K=1$ as in the main paper.

Therefore the bound is rewritten as:
$$
\E\|\Delta\|_\infty\leq C_1N_1pw_m^{(L)}\prod\limits_{l'=2}^L N_{l'}w_m^{(l')}
$$

Now we notice that the quantity $N_lw_m^{l}=\|W^l\|_\infty$ and therefore
$$
\E\|\Delta\|_\infty\leq pC_1\frac{N_1}{N_L}\|W_2\|_\infty\cdot...\|W_L\|_\infty
$$

Now we assume that the network has one more layer so that the bound from~\cite{ontherobustness17} works for a faulty input in the original network:
$$
\E\|\Delta\|_\infty \leq pC\frac{N_0}{N_L}\|W_1\|_\infty\cdot\ldots\cdot\|W_L\|_\infty
$$

Here $C=\max\{|x_i|\}=\|x\|_\infty$. Next we prove that result independently using Additional Proposition~\ref{th:bound_v1} for $\|\cdot\|_\infty$
$$
\|\Delta_L\|_\infty\leq\|W_L\|_\infty\cdot\ldots\cdot\|W_1\|_\infty\|\xi\odot x-x\|_\infty
$$

Now we calculate $\E\|\xi\odot x - x\|_\infty$. We write the definition of the expectation with $f(p, n, k)=p^k(1-p)^{n-k}=p^k+\mathcal O(p^{k+1})$ being the probability that a binary string of length $n$ has a particular configuration with $k$ ones, if its entries are i.i.d. Bernoulli $Be(p)$. Here $S_l$ is the set of all possible network crash configurations at layer $l$. Each configuration $s_l\in S_l$ describes which neurons are crashed and which are working. We have $|S_l|=2^{n_l}$.
$$
\E\|\xi\odot x -x\|_\infty=\sum\limits_{s\in S}f(p,n,|s|)\max\{s\odot x - x\}
$$

Now since for $|s|=0$ the $\max\{s\odot x - x\}=\max\{x-x\}=0$, we consider cases $|s|=1$ and $|s|>1$. For $|s|>1$ the quantity $f(p,n,|s|)=\mathcal O(p^2)$ and therefore, in the first order:
$$
\E\|\xi\odot x-x\|_\infty=p\sum\limits_{i}|x_i|+\mathcal O(p^2)\leq pN_0\|x\|_\infty+\mathcal O(p^2)
$$

Next we plug that back into the expression for $\E\|\Delta\|_\infty$:
$$
\E\|\Delta\|_\infty\leq \|W_L\|_\infty\cdot\ldots\cdot\|W_1\|_\infty pN_0\|x\|_\infty+\mathcal O(p^2)
$$

Now we note that this expression and the expression from~\cite{ontherobustness17} differ only in a numerical constant in front of the bound: $\frac{N_0}{N_L}$ instead of $N_0$, but the bounds behave in the same way with respect to the weights.

%This proof shows how bounds proved in the main paper relate to bounds in~\cite{ontherobustness17}.
\end{proof}

\begin{proposition}[Absolute value bound or bound b2]
\label{th:bound_v2}
The error on input $x$ can be upper-bounded as:
$$\E |\Delta_L|\leqslant p|W_L|\cdot\ldots\cdot |W_1||x|$$

For $|W|$ being the matrix of absolute values $(|W|)_{ij}=|W_{ij}|$. $|x|$ means component-wise absolute values of the vector.
\end{proposition}

\begin{proof}
This expression involves absolute value of the matrices multiplied together as matrices and then multiplied by an absolute value of the column vector. The absolute value of a column vector is a vector of element-wise absolute values.

We assume a faulty input. By Definition~\ref{def:outputfail}, the error at the last layer is $\Delta_L=\hat{y}_L-y_L$. By definition~\ref{def:nn}, $y_L=W_Ly_{L-1}+b_L$ and by Definition~\ref{def:neuronfail}, $\hat{y}_L=W_L\hat{y}_{L-1}+b_L$. Thus for the $i$'th component of the error, $$|\Delta_L^i|=|W^i_L\hat{y}_{L-1}+b^i_L-W^i_Ly_{L-1}-b^i_L|=|W^i_L(\hat{y}_{L-1}-y_{L-1})|=|\sum\limits_jW^{ij}(\hat{y}^j_{L-1}-y^j_{L-1})|$$

By the triangle inequality,
$$|\Delta_L^i|\leq \sum\limits_j|W^{ij}_L|\cdot |\hat{y}^j_{L-1}-y^j_{L-1}|=(|W_L||\hat{y}_{L-1}-y_{L-1}|)^i$$

Next we go one level deeper according to Definition~\ref{def:nn}:
$$
|\hat{y}_{L-1}^j-y_{L-1}^j|=|\varphi(W^j_{L-1}\hat{y}_{L-2}+b^j_{L-1})-\varphi(W^j_{L-1}y_{L-2}+b^j_{L-1})|
$$

And then apply the 1-Lipschitzness property of $\varphi$:
$$|\hat{y}_{L-1}^j-y_{L-1}^j|\leq|W^j_{L-1}\hat{y}_{L-2}+b^j_{L-1}-W^j_{L-1}y_{L-2}-b^j_{L-1}|=|W^j_{L-1}(\hat{y}_{L-2}-y_{L-2})|$$

This brings us to the previous case and thus we analogously have
$$
|\Delta_L|\leq |W_L|\cdot |W_{L-1}|\cdot |\hat{y}_{L-2}-y_{L-2}|
$$

Inductively repeating these steps, we obtain:
$$
|\Delta_L|\leq |W_L|\cdot\ldots\cdot|W_1||\hat{x}-x|
$$

Now we take the expectation and move it inside the matrix product by linearity of expectation:
$$
\E|\Delta_L|\leq\E|W_L|\cdot\ldots\cdot|W_1||\hat{x}-x|=|W_L|\cdot\ldots\cdot|W_1|\E|\hat{x}-x|
$$

The last expression involves $\E|\hat{x}-x|$. This is component-wise expectation of a vector and we examine a single component. Since $\xi^i$ is a Bernoulli random variable,
$$
\E|x_i\xi^i-x_i|=p\cdot|x_i|+(1-p)\cdot 0
$$
Plugging it into the last expression for $\E|\Delta_L|$ proves the proposition.
\end{proof}

\section{Realistic Simplifying Assumptions for Neuromorphic Hardware}

\subsection{When is The Weight Perturbation Small?}
In this section we give sufficient conditions for which case the probability of large weight perturbation under a crash distribution is small. First, we define the properties of neuromorphic hardware.

\begin{assumption}
\label{assumption:psmall}
The probability of failure $p=\max\{p_l\big| l\in[L]\}$ is small: $p\lesssim 10^{-4}..10^{-3}$
\end{assumption}

\begin{assumption}
\label{assumption:nbig}
The number of neurons at each layer $n_l$ is sufficiently big, $n_l\gtrsim 10^2$
\end{assumption}

\begin{addassumption}
	\label{assumption:w_qbalanced}
	Vectors $W_l^i=\sum_j|W_l^{ij}|$ are $q$-balanced for $q>0$ for each layer $l$.\footnote{A similar assumption on an "even distribution" of weights was made in \cite{548899}.}
\end{addassumption}

\paragraph{Toy examples.} Naturally, we expect the error to decay with an increase in number of neurons $n$, because of redundancy. We show that this might not be always the case. First, consider a 1-layer NN with $n=n_0$ and $n_L=n_1=1$ at input $x_i=1$: $y(x)=\sum x_i/n$. This is most robust network, as all neurons are interchangeable as they are computing the same function each. Essentially, this is an estimate of the mean of $\xi_i$ given $n$ samples. Here $\E\Delta=-\sum \E\xi_ix_i=-p$ and $\Var\Delta=1/n\sum \Var\xi_i=p(1-p)/n\sim p/n$, variance decays with $n$. In contrast, the worst case $y(x)=x_1$ has all but one neuron unused. Therefore $\E\Delta=x_i\E\xi_i p$ and $\Var\Delta=x_i^2\Var\xi_i=p(1-p)\sim p$, variance does not decay with $n$.

This proposition gives sufficient conditions under which the weight perturbation is small and it is less and less as $n$ increases and $p$ decreases:
\begin{mainprop}
	\label{prop:usmall}
	Under A\ref{assumption:psmall},\ref{assumption:nbig} and AA\ref{assumption:w_qbalanced}, for $\alpha>p$, the norm of the weight perturbation $U_l^i$ at layer $l$ is probabilistically bounded as:
	$\delta_0=\Prob\{\|U_l^{\cdot}\|_1\geq \alpha \|W_l^{\cdot}\|\}\leq \exp\left(-n_l\cdot q\cdot d_{KL}(\alpha||p_l)\right)$ with KL-divergence between numbers $a,b\in(0,1)$, $d_{KL}(a,b)=a\log{a}/{b}+(1-a)\log{(1-a)}/{(1-b)}$
\end{mainprop}
\begin{proof}
See \cite{arratia1989tutorial} for the proof details as this is a standard technique based on Chernoff bounds for Binomial distribution with $p\to 0$. These are a quantified version of the Law of Large Numbers (which states that an average of identical and independent random variables tends to its mean). Specifically, Chernoff bounds show that the tail of a distribution is exponentially small: $\mathbb P\{X\geq \E X+\varepsilon \E X\}\leq \exp(-c\varepsilon^2)$.

Specifically, if we consider a case $X=Bin(n,p)$ with $p\to 0$, for which $q=1$, we have \cite{arratia1989tutorial,wiki_binomial} for $\alpha=k/n>p$:
$$
\mathbb P[X\geq k]\leq \exp\left(-nD_{KL}\left(\frac{k}{n}\big|\big|\alpha\right)\right)
$$
In case if we rewrite $k=\alpha n$, this gives us the result.

Specifically, if we consider $\|U_l^\cdot\|_1=\sum\limits_i |U_l^i|\approx k |W_l|$. Therefore, this probability is $\mathbb P[X\geq k]=P[\|U_l^{\cdot}\|_1\geq \alpha \|W_l^{\cdot}\|_1]$. This shows that the probability that a fraction of Bernoulli successes is more and more concentrated around its mean, $np$. Therefore, it is less and less probable that the this fraction is $\geq \alpha > p$.

Factor $q$ appears because in the analysis of the sum
$$\frac{\left(\sum\limits_{i=1}^n|W^i|\right)^2}{\sum\limits_{i=1}^n|W^i|^2}\geq \frac{n^2W_{\min}^2}{nW_{\max}^2}=nq$$
\end{proof}

\subsection{Taylor Expansion for a Small Weight Perturbation}
In this section, we develop a more precise expression for $\E\Delta$ and $\Var\Delta$. Previously, we have seen that the perfect fault-tolerant network has $\Var\Delta=\mathcal O(p/n)$. In this section, we give sufficient conditions when complex neural networks behave as the toy example from the previous section as well.

We would like to obtain a Taylor expansion of the expectation of random variable $\E\Delta=T_1+T_2$ in terms of $p$ and $q=1/n$ with $r=p+q$ where $T_1=\mathcal O(p)$ and $T_2=\mathcal O(r^2)$. For the variance, we want to have $\Var\Delta=T_3+T_4$ with $T_3=\mathcal O(p/n)$ and $T_4=\mathcal O(r^3)$. Our goal here is to make this expression decay as $n\to\infty$ as in the toy example. We will show in Theorem \ref{prop:taylor}, the first-order terms indeed behave as we expect. However, the expansion also contains a remainder (terms $T_2$ and $T_4$). In order for the remainder to be small, we need additional assumptions. It is easy to come up with an example for which the first term $T_1$ is zero, but the error is still non-zero, illustrating that the remainder is important. Consider one neuron working and the rest having zero weights. Consider a summation of outputs of the neurons, each with a quadratic activation. Then $y=(x_i-\alpha)^2$, and $\nabla_x y=0$ at $x_i=\alpha$. In addition, $\E\Delta=\Theta(p)$, however, the first term in Taylor expansion is $T_1=0$. The remainder here is $T_2=O(p)$ and not $O(r^2)$, no matter how many neurons we have. The problem with this example is discontinuity: one neuron with non-zero weights is not at all like its neighbors. We thus show that discontinuity can lead to a lack of fault tolerance. Next, we generalize this sketch and show that some form of continuity is sufficient for the network to be fault-tolerant.

First, we reiterate on the toy motivating example from the previous section. Consider a 1-layer neural network $$y(x)=\sum\limits_{i=1}^nw_ix_i$$
We assume that all neurons and inputs are used. Specifically, for the $q$-factor $q(x)=\max|x_i|/\min|x_i|$, we have $q(x)\approx q(w)\sim 1$. We are interested in how the network behaves as $n\to\infty$ (the infinite width limit). For the input, we want the magnitude of individual entries to stay constant in this limit. Thus, $|x_i|=\mathcal O(1)$. Now we look at the function $y(x)$ that the network computes. Since the number of terms grows, each of them must decay as $1/n$: $w_i\sim 1/n$. The simplest example has $x=(1,...,1)$ and $w=(1/n,...,1/n)$, which results in $y(x)\equiv 1$ for all $n$. Now we consider fault tolerance of such a network. We take $\Delta=\sum\limits_{i=1}^nw_i((1-\xi^i)-1)x_i$ for $\xi^i\sim Be(p)$ being the indicator that the $i$'th input neuron has failed. Therefore, $\E\Delta=-p/n\sum x_iw_i=-p$, and $\Var\Delta=\sum x_i^2w_i^2p(1-p)\sim p/n$. We see that the expectation does not change when $n$ grows, but variance decays as $1/n$. These are the values that we will try to obtain from real networks. Intuitively, we expect that the fault tolerance increases when width increases, because there are more neurons. This is the case for the simple example above. However, it is not the case if all but one neuron are unused. Then, the probability of failure is always $p$, no matter how many neurons there are. Thus, the variance does not decrease with $n$. We thus are interested in utilizing the neurons we have to their maximal capacity to increase the fault tolerance.

%\todo[inline]{M: pointer here}

\begin{figure}[tbh]
    \centering
    \includegraphics[width=0.9\textwidth]{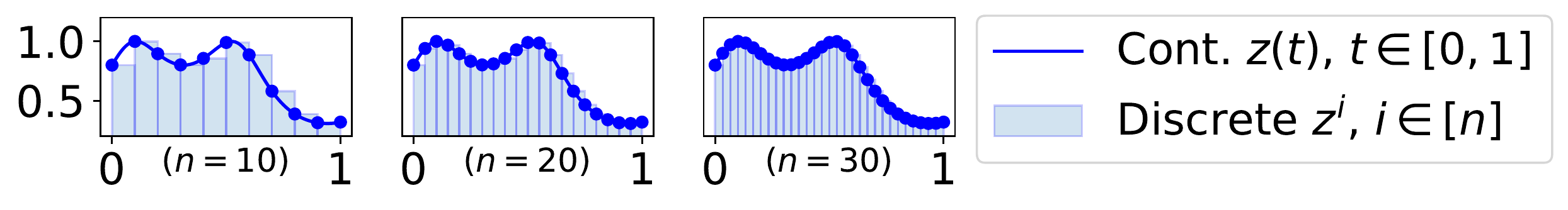}
    \caption{Discrete (standard) neural network approximating a continuous network}
    \label{fig:continuous_net}
\end{figure}

Overall, we have the following plan. We give an expansion of $\Var\Delta$ in terms of $p$ and $q=1/n$, and give sufficient conditions for which the remainder terms are small. We use the first term explicitly when regularizing the network. In order for the expansion to work, we formalize the difference between "all neurons doing the same task" and "all but one are unused". Specifically, we define a class of "good" networks for which fault tolerance is sufficient, via the {\em continuous limit} \cite{sonoda2017double}.

\newcommand{\ff}{{\mathcal F}}

\paragraph{Functions, functionals and operators.} We call maps from numbers to numbers as functions, maps from functions to numbers as functionals and maps from functions to functions as operators.

We consider a subset of the space of real-valued functions with domain $T$: $\ff(T)=BPC^{\infty}(T)$ for $T=[0,1]$. This is a space of bounded piecewise-continuous functions $f\in\mathcal F$, $f\colon T\to \mathbb R$ such that there is a finite set of points $\{x_i\}\subset T$ such that $f|_{(x_i,x_{i+1})}\in C^{\infty}(T)$ -- infinitely-differentiable. We note that $\mathcal F\subset \mathcal L^1$ which means that $\int_{t\in T} |f(t)|dt<\infty$. In addition, $\int |f^{k}(t)|<\infty$. Moreover, for two $f,g\in \ff$, $(f\cdot g)(t)=f(t)\cdot g(t)\in \ff$.

\begin{adddefinition}{(Continuous neural network)}
\label{def:cont_nn}
A continuous neural network \cite{le2007continuous} of function class $\mathcal H$ with $L$ layers is an operator $y_L\colon \mathcal H\to\mathcal H$ defined by a tuple $(L,W,B,\varphi)$ with a tuple of weights functions $W=(W_1,...,W_L)$ with $W_l\in\mathcal H(T^2)$ (or their distributions), bias functions $B=(b_1,...,b_L)$ (or their distributions) with $b_l\in \ff$ by the expression
	$y_l(t)=\varphi(z_l(t)),\,z_l(t)=\int\limits_{t'\in T}W_l(t,t')y_{l-1}(t')dt'+b_l(t),\,l\in\overline{1,L}$, $z_0=y_0=x$ and $y_L(t)=z_L(t)$.
\end{adddefinition}

We note that a regular neural networks has $T_l=[n_l]$. Continuous networks were re-introduced in \cite{le2007continuous}. The classes of discrete and continuous networks can be generalized as Deep Function Machines \cite{guss2016deep}, as a special case when $L_1$-norms of inputs and activations of discrete networks approximating continuous ones increase linearly with $n$. In \cite{sonoda2017double}, depth of a network is also considered to be continuous. Continuous networks are also called the integral representation \cite{sonoda2017double}. Previous work considers other limits rather than $L_1$-norm growing linearly, but here we only consider this case.

Next, we connect continuous and regular (discrete) networks. To do that, we define how each quantity transforms when $n$ grows.

\begin{adddefinition}{(Continuous-discrete correspondence with $\|x\|_1=\mathcal O(n_0)$)}
\label{def:cont_discr_nn}
Consider a continuous $NN_c=(L,W_c,B_c,\varphi)$ and a discrete one $NN_d=(L,W_d,B_d,\varphi)$. We consider a distribution of continuous inputs $\ff\ni x\sim P_c$ and a distribution of discrete inputs $\mathbb{R}^n\ni x\sim P_d$. These distributions are linked: each element from $P_c$ corresponds to exactly one from $P_d$. We define the approximation error $A$ as

$$
A=\sup\limits_{(x_c,x_d)\sim P_c\times P_d}\left[\sup\limits_{0\leq l\leq L}\left[\sup\limits_{1\leq i\leq n_l}\left|(y^d_l)^i-(y^c_l)\left(\frac{i-1}{n_l-1}\right)\right|\right]\right]
$$

We define $n=\min\{n_0,...,n_L\}$ the minimal width of the discrete network. If a series of discrete networks $NN_n$ has $A_n\to 0$, $n\to\infty$ for some $NN_c$, we say that $NN_n\to NN_c$.
\end{adddefinition}

The summary of correspondences between discrete and continuous quantities is shown in Table \ref{tab:continuous_discrete_correspondence}

\newcommand{\R}{{\mathbb R}}

\begin{table}[thb]
    \centering
    \begin{tabular}{cccc}
    \toprule
    Quantity & Discrete (D\ref{def:nn}) & Continuous (AD\ref{def:cont_nn}) & Relation (AD\ref{def:cont_discr_nn}, AP\ref{prop:cont_to_discr})\\\midrule
    Input & $x\colon [n_0]\to \mathbb R$ & $x\colon [0,1]\to\R$ & $x_i\approx x\left(\frac{i-1}{n_0-1}\right)$\\
    
    Norm  & $\|x_d\|_1=\sum_i|x_i|$ & $\|x_c\|_1=\int|x(t)|dt$ & $\frac{1}{n_0}\|x_d\|_1\approx \|x_c\|_1$\\

    Bias & $b_l\colon [n_l]\to\R$ & $b_l\colon [0,1]\to \R$ & $b_l^i\approx b_l\left(\frac{i-1}{n_l-1}\right)$\\
    
    Weights & $W_l\colon [n_l]\times[n_{l-1}]\to\R$ & $W_l\colon [0,1]^2\to \R$ & $W_l^{ij}\approx{\bf \frac{1}{n_{l-1}}} W_l\left(\frac{i-1}{n_l-1}, \frac{j-1}{n_{l-1}-1}\right)$\\
    
    Pre-activations & $z_l^i=\sum_j W_l^{ij}y_{l-1}^i+b_l^i$ & $z_l(t)=\int_0^1 W_l(t,t')y_{l-1}(t')dt'+b_l(t)$ & $z_l^i\approx z_l\left(\frac{i-1}{n_l-1}\right)$\\
    
    Number of changes & $C_d=\sum_{ij}\left|W_l^{i+1,j}-W_l^{ij}\right|$ & $C_c=\int\int dtdt'|W'_t(t,t')|$ & $C_d\approx C_c$\\\bottomrule
    \end{tabular}
    \caption{Correspondence between discrete and continuous quantities}
    \label{tab:continuous_discrete_correspondence}
\end{table}

\begin{proposition}
\label{prop:completeness}
Continuous networks with $\mathcal H=\ff$ are universal approximators
\end{proposition}
\begin{proof}
Based on the proof of \cite{le2007continuous}. By the property of discrete networks, they are universal approximators \cite{le2007continuous}. Take a discrete network $y$ with a sufficiently low error, and define a continuous network $NN_c$ by using a piecewise-constant $W$ and $b$ from $y$. Then, $NN_c\equiv NN_d$. The weights and biases are bounded and piecewise-continuous with $n_l$ discontinuities at each layer. In addition, all functions are bounded since the weights are finite.
\end{proof}

In the following we will always use $\mathcal H=\ff$, as it is expressive enough (AP\ref{prop:completeness}), and it is useful for us. We give a sufficient condition for which $A_n\to 0$ as $n\to\infty$.

\begin{proposition}
\label{prop:cont_to_discr}
If a discrete network $NN_n$ is defined from a continuous network $NN_c$ in the following way, then $NN_n\to NN_c$ with error $A=\mathcal O(1/n)$. See (Figure \ref{fig:continuous_net})
$$x^i=x\left(\frac{i-1}{n_0-1}\right),\,W^{ij}_l=\frac{1}{n_{l-1}}W_l\left(\frac{i-1}{n_l-1},\frac{j-1}{n_{l-1}-1}\right),\,b_l^i=b_l\left(\frac{i-1}{n_l-1}\right)$$
\end{proposition}

In the following we write $\hat{i}_l=\frac{i-1}{n_l-1}$ for an index $i=1..n_l$, as the range for each of the indices is known.

\begin{proof}
For layer $0$, the error is $0$ by definition of $x^i$. Also, $x$ and its derivatives are bounded. Suppose we have shown that the error for layers $1..l-1$ is $\leq\varepsilon$, and that $y_{l'}$ is bounded with globally bounded derivatives. Consider $\left|(y^d_l)^i-(y^c_l)\left(\frac{i-1}{n_l-1}\right)\right|=|b_l^i-b_l^i+\sum_jW_{ij}y_{l-1}^j-\int W(\hat{i},t')y_{l-1}(t')dt'|\leq|1/n_{l-1}\sum_jW(\hat{i},\hat{j})(y_{l-1}^j+\varepsilon)-\int W(\hat{i},t')y_{l-1}(t')dt'|\leq \varepsilon\int |W|dtdt'+|\int f(t)dt-1/n\sum f(\hat{j})|$ for $f(t)=W(\hat{i},t)y_{l-1}(t)$. The second term is a Riemann sum remainder and can be upper-bounded as  $\frac{|T|^2}{2n}\sup |f'(t)|$. We note that $|[0,1]|=1$ and that $|f'(t)|=|W'_{t}(\hat{i},t)y_{l-1}(t)+W(\hat{i},t)y'_{l-1}(t)|\leq \sup |W'|\sup |y|+\sup |W|\sup |y'|<\infty$ and does not depend on $n$ or the input, as the bound is global. Therefore, the error at layer $l$ is $\leq (\varepsilon+1/n)C\leq D/n$ for the initial choice $\varepsilon=1/n$. Thus, the full error is decaying as $1/n$: $A_n=\mathcal O(1/n)$.
\end{proof}

Using that result, we conclude that for any sufficiently smooth operator $y$, there exist a sequence $NN_n$ approximating $y$. First, by \cite{le2007continuous}, continuous networks are universal approximators. Taking one and creating discrete networks from it (as in AP\ref{prop:cont_to_discr}) gives the desired result as $\sup\limits_x\|y_c-y_d\|_{\infty}\leq A=\mathcal O(1/n)\leq\varepsilon $ and $\|y_c-y\|\leq \varepsilon$ by the approximation theorem for continuous nets.

\begin{assumption}
\label{assumption:continuous_net}
We assume that $NN_n\to NN_d$. We assume that there are global reasonable bounds on the derivatives of the function we want to approximate

$$\sup\limits_{x\sim P}\left|\frac{\delta^k y_L(t_L)}{\delta y_l(i_l^1)...\delta y_l(i_l^k)}\frac{1}{(i_l^1,...,i_l^k)!}\right|\leq D_k,\,D_{a:b}=\max\{D_a,D_{a+1},...,D_b\}$$

Here for $i_s\in[n]$ and $q_j=|\{i_s=j\}|$ we define $(i_0,...,i_k)!=\frac{n!}{q_1!...q_n!}$ a multinomial coefficient. In the paper, we only need $k\in\{1,2\}$.
\end{assumption}

The limit means that close-by neurons inside a discrete network compute similar functions. This helps fault tolerance because neurons become redundant. There can be many of such sequences $NN_n$, but for the sake of clarity, one might consider that gradient descent always converges to the "same way" to represent a function, regardless of how many neurons the network has. This can be made more realistic if we instead consider the distribution of continuous networks to which the gradient descent converges to. In this limit, the distribution of activations at each layer (including inputs and outputs) stays the same as $n$ grows, and only the number of nodes changes. Note that this limit makes neurons ordered: their order is not random anymore. The limit works when $n_l$ are sufficiently large. After that threshold, the network stops learning new features and only refines the ones that it has created already.

The derivative bound part of A\ref{assumption:continuous_net} can be enforced if there is a very good fit with the ground truth function $\tilde{y}(x)$. Indeed, if $\tilde{y}(x)\equiv y_L(x)$, then the derivative of the network depends on the function being approximated. For discrete-continuous quantities we write $X_d\approx X_c$ meaning that $|X_d-X_c|\leq\varepsilon$ for a sufficiently large $n$.

We note that our new assumptions extend the previous ones:
\begin{proposition}
\label{prop:cont_to_qfactor}
A\ref{assumption:continuous_net} results in AA\ref{assumption:w_qbalanced} with $$q=\frac{\left(\int\int|W(t,t')|dtdt'\right)^2}{\int\left(\int W(t,t')|dt\right)^2dt'}+o(1),\,n_l\to\infty$$
\end{proposition}
\begin{proof}
A simple calculation:
$$\sum |W^i|=n\cdot\frac{1}{n}\sum \sum_{ij}|W^{ij}|\to n\int |W(t,t')|dtdt',\,\sum |W^i|^2\to n\int\left(\int |W(t,t')dt1|\right)^2dt$$
Dividing these two gives the result.
\end{proof}

\paragraph{Can we make the input norm bounded?} Sometimes, input data vectors are normalized: $\|x\|=1$ \cite{jacot2018neural}. In our analysis, it is better to do the opposite: $\|x\|_1=\mathcal O(n_0)$. First, we want to preserve the magnitude of outputs to fit the same function as $n$ increases. In addition, the magnitude of pre-activations must stay in the same range to prevent vanishing or exploding gradients. We extend that condition to the input layer as well for convenience. Another approach would be to, for example, keep the norm $x$ constant as $n$ grows. In that case, to guarantee the same output magnitude, input weights must be larger. This is just a choice of scaling for the convenience of notation. We note that inputs can be still normalized {\em component-wise}.

\paragraph{Less weight decay leads to no fault tolerance.} Consider the simplest case with Gaussian weights: $y=w^Tx$, $w_i\sim N(0,\sigma^2)$. Then $y\sim N(0, \sigma^2\|x\|_2^2)$. In order to avoid vanishing gradients, we set $\sigma^2\|x\|_2^2=const$. If $\|x\|_2^2=\mathcal O(n)$, then we must set $\sigma^2\sim 1/n$ for the variance of the output to stay constant. This is a well-known initialization technique preserving the {\em variance} of pre-activations and thus eliminating vanishing or exploding gradients. In contrast to this, by A\ref{assumption:continuous_net} we preserve the {\em mean} activation at each neuron instead, like for stochastic neural networks \cite{tang2013learning}. In the Gaussian case, the error $\Delta=-\sum \xi^i x_iw_i$. $\E\Delta=0$ and $\Var\Delta=\sum x_i^2\E (\xi^i)^2\E w_i^2=\sigma^2p\|x\|_2^2=p\cdot const$. This does not decay with $n$. A more formal statement can be found in AP\ref{prop:nodecay}.

\paragraph{The NTK limit.} In the NTK limit \cite{jacot2018neural}, there is a decay of the variance $\Var\Delta$ since both $\|x\|_2=1$ and $\sigma^2\sim 1/n$, but that happens because variance decreases in any case, not related to fault tolerance, since $\sigma^2\|x\|_2^2\sim 1/n$. The "lazy" regime \cite{chizat2019lazy} of the NTK limit implies that every hidden neuron is close to its initialization, which is random. Thus, there is no continuity in the network: close-by neurons do not compute similar functions. Thus, the NTK limit is incompatible with our continuous limit.

Below we formalize the claim that a network with constant pre-activations variance is not fault-tolerant. This means that an untrained network with standard initialization is not fault tolerant.

\begin{proposition}
	\label{prop:nodecay}
The fault tolerance error variance for a network $y=w^Tx$ with $\E w_i=0$ and $\Var [y(x)]=const$ does not decay with $n$
\end{proposition}
\begin{proof}
Consider $y(x)=\sum w_ix_i$ and $\Delta=-\sum w_ix_i\xi^i$. Since $\E w_i=0$, $\E y(x)=\E\Delta=0$. In addition, $\Var y(x)=\sum \E w_i^2\E (x^i)^2=const$. Then, $\Var\Delta=\sum \E w_i^2\E (x^i)^2\E(\xi^i)^2=p\Var y(x)=p\cdot const$, not decaying with $n$
\end{proof}

Next, we want to harness the benefits of having a continuous limit by A\ref{assumption:continuous_net}. We want to bound the Taylor remainder, and for that, we bound higher-order derivatives. We bound them via the operator derivative of the continuous limit.

\paragraph{Operator derivatives.} The derivatives for functionals and operators are more cumbersome to handle because of the many arguments that they have (the function the operator is taken at, the function the resulting derivative operator acts on, and the arguments for these functions), so we explicitly define all of them below.  We consider operators of the following form, where $x\in\ff$ and $Y[x]\in\ff$ are functions:
$$Y[x](t)=\sigma\left(\int W(t,t')x(t')dt'\right)$$

This is one layer from AD\ref{def:cont_nn}. We define the {\em operator derivative} of $Y[x]$ point-wise with the RHS being a {\em functional derivative}:
$$\left(\frac{\delta Y[x]}{\delta x}\right)(t)=\frac{\delta (Y[x](t))}{\delta x}$$

For functionals $F[x]$ we use the standard functional derivative:
$$\frac{\delta F[x]}{\delta x}[\varphi]=\lim\limits_{\varepsilon\to 0}\frac{F[x+\varepsilon\varphi]-F[x]}{\varepsilon}$$

For our case $F[x]=F_t[x]=Y[x](t)$, we have
$$
\frac{\delta F[x]}{\delta x}[\varphi]=\lim\limits_{\varepsilon\to 0}\varepsilon^{-1}\cdot\left(\sigma\left(\int W(t,t')(x(t')+\varepsilon\varphi(t'))dt'\right)-\sigma\left(\int W(t,t')x(t')dt'\right)\right)=\sigma'(F[x])\int W(t,t')\varphi(t')dt'
$$

Next, we consider the functional derivative {\em at a point}. This quantity is similar to just one {\em component} of the gradient of an ordinary function, with the complete gradient being similar to the functional derivative defined above. We define the derivative at a point $\delta F[x]/\delta x(s)$ via the Euler-Lagrange equation, since we only consider functionals of the form $F_t[x]=Y[x](t)$ for some fixed $t$ which are then given by an integral expression
$$
F_t[x]=Y[x](t)=\sigma\left(\int \underbrace{W(t,t')x(t')}_{L(t',x(t'),x'(t'))}dt'\right)
$$

We define the inner part $G[x]=\int W(t,t')x(t')dt'$, thus, $F[x]=\sigma(G[x])$. In this case, since the integral only depends on the function $x$ explicitly, but not on its derivatives, the functional derivative at point $s$ is
$$
\frac{\delta F[x]}{\delta x(s)}=\sigma'(G[x])\left(\frac{\partial L}{\partial x}-\frac{d}{ds}\frac{\partial L}{\partial x'}\right)=\sigma'(G[x])\frac{\partial L}{\partial x}=\sigma'(G[x])W(t,s)
$$

The definition of a functional derivative at a point $\delta F[x]/\delta x(s)$ ("component of the gradient") can be reconciled with the definition of the functional derivative $\delta F[x]/\delta x$ ("full gradient") if we consider the Dirac delta-function:
$$
\frac{\delta F[x]}{\delta x}[\varphi=\delta(t'-s)]=\sigma'(G[x])\int W(t,t')\delta(t'-s)=\sigma'(G[x])W(t,s)\equiv \frac{\delta F[x]}{\delta x(s)}
$$

We define the {\em operator derivative at a point} in a point-wise manner via the functional derivative at a point:
$$
\left(\frac{\delta Y[x]}{\delta x(s)}\right)(t)=\frac{\delta Y[x](t)}{\delta x(s)}
$$
Having that definition, we compute for our case $(\delta Y[x]/\delta x(s))(t)=\sigma'(G_t[x])W(t,s)$.

Now we see that the rules for differentiating operators {\em in our case} are the same as the well-known rules for the derivatives of standard vector-functions. Indeed, if we consider $y_i(x)=\sigma(\sum_j W_{ij}x_j)$ with the inner part $g_i(x)=\sum_j W_{ij}x_j$ giving $y_i(x)=\sigma(g_i(x))$, then $\partial y_i/\partial x_k=\sigma'(g(x))W_{ik}$. This looks exactly like the expression for $\partial Y[x](t)/\partial x(s)$. By induction, this correspondence holds for higher derivatives as well. However, this does not imply that these quantities are equal. In fact, we will show that they differ by a factor of $1/n_l^k$ where $k$ is the order of the derivative.

We characterize the derivatives of a discrete NN ${\partial^k y_L}/{\partial y_l^{i_l^1}...\partial y_l^{i_l^k}}$ in terms of operator derivatives of a continuous NN ${\delta^k y_L}/{\delta y_l(i_l^1)...\delta y_l(i_l^k)}$. We only assume the continuous limit:

\begin{mainprop}
	\label{proposition:decay}
For a sequence $NN_n\to NN_c$ with $\varphi\in C^{\infty}$, the derivatives decay as:
$$\frac{\partial^k y_L}{\partial y_l^{i_l^1}...\partial y_l^{i_l^k}}=\frac{1}{n_0^k}\frac{\delta^k y_L}{\delta y_l(i_l^1)...\delta y_l(i_l^k)}+o(1),\,n_l\to\infty$$
\end{mainprop}

Intuitively, this means that the more neurons we have, the less is each of them important. First, consider a simple example $y=\sigma(\sum w_i x_i)$. Here the weight function is $w_i=1/n$, $w(\hat{i})=1$, $x_i=1$,  and $A=0$. Then $\partial y/\partial x_i=\sigma'(\cdot)w_i\sim 1/n$ and $\partial^2 y/\partial x_i\partial x_j=\sigma''(\cdot)w_iw_j\sim 1/n^2$. We note that the expression inside the sigmoid has a limit and it's close to the integral by continuity of $\sigma'$ and $\sigma''$.

\begin{proof}Now we prove P\ref{proposition:decay}.
Consider the first and second derivatives. Note that the operator derivatives only depend on the number of layers and the dataset. It does not depend on any $n$ anymore.

\begin{align*}
\frac{\partial y_L^{i_L}}{\partial x_{i_0}}&=\sum\limits_{i_{L-1},...,i_1}W_L^{i_Li_{L-1}}...W_1^{i_1i_0}\sigma'(z_L^{i_L})...\sigma'(z_1^{i_1}) && \text{By definition of a neural net}\\
&\approx\frac{1}{n_0}\int dt_1 W_1(t_1,t_0)\sigma'(z_1(t_1))...\int dt_{L-1} W_{L-1}(t_{L-1},t_{L-2})\cdot W_L(t_L, t_{L-1})\cdot \sigma'(z_{L-1}(t_{L-1})) && \text{continuous limit}\\
&=\frac{1}{n_0}\frac{\partial y_L(i_L)}{\partial x(i_0)} && \text{operator $\sim$ ordinary}
\end{align*}
Crucially, the factor $1/n_0$ appears because we {\bf do not sum} over $i_0$, as it is fixed, but we have a weight vector $W_1\sim 1/n_0$ nevertheless. For all other indices, we have a weight matrix $W_l\sim 1/n_{l-1}$ {\bf as well as} a summation over $i_{l-1}$.

\begin{align*}
\frac{\partial^2 y_L^{i_L}}{\partial x_{i_0}\partial x_{i_0'}}&=\sum\limits_{i_{L-1}...i_1} W_L^{i_Li_{L-1}}...W_1^{i_1i_0}\sum\limits_{s=1}^L\sigma''(z_s^{i_s})\prod\limits_{s'\neq s}\sigma'(z_{s'}^{i_{s'}})\sum\limits_{i_s'...i_1'}W_s^{i_s'i_{s-1}'}...W_1^{i_1'i_0'} && \text{From previous}\\
&\approx\frac{1}{n_0^2}\frac{\delta^2 y_L(i_L)}{\delta x(i_0)\delta x(i_0')} && \text{continuous+operator $\sim$ ordinary}
\end{align*}
Here, the factor $1/n_0^2$ appears because we never sum over $i_0$, but the weight matrix $W_1\sim 1/n_0$ appears twice.
\end{proof}

\paragraph{When does the limit hold?} Now, we have {\em assumed} that a network has a continuous limit in A\ref{assumption:continuous_net}. However, this might not be the case: the NTK limit \cite{jacot2018neural} is an example of that, as weights there stay close to their random initialization \cite{chizat2019lazy}, thus, they are extremely dissimilar.

\begin{figure}[thb]
    \centering
    \includegraphics[width=0.19\textwidth]{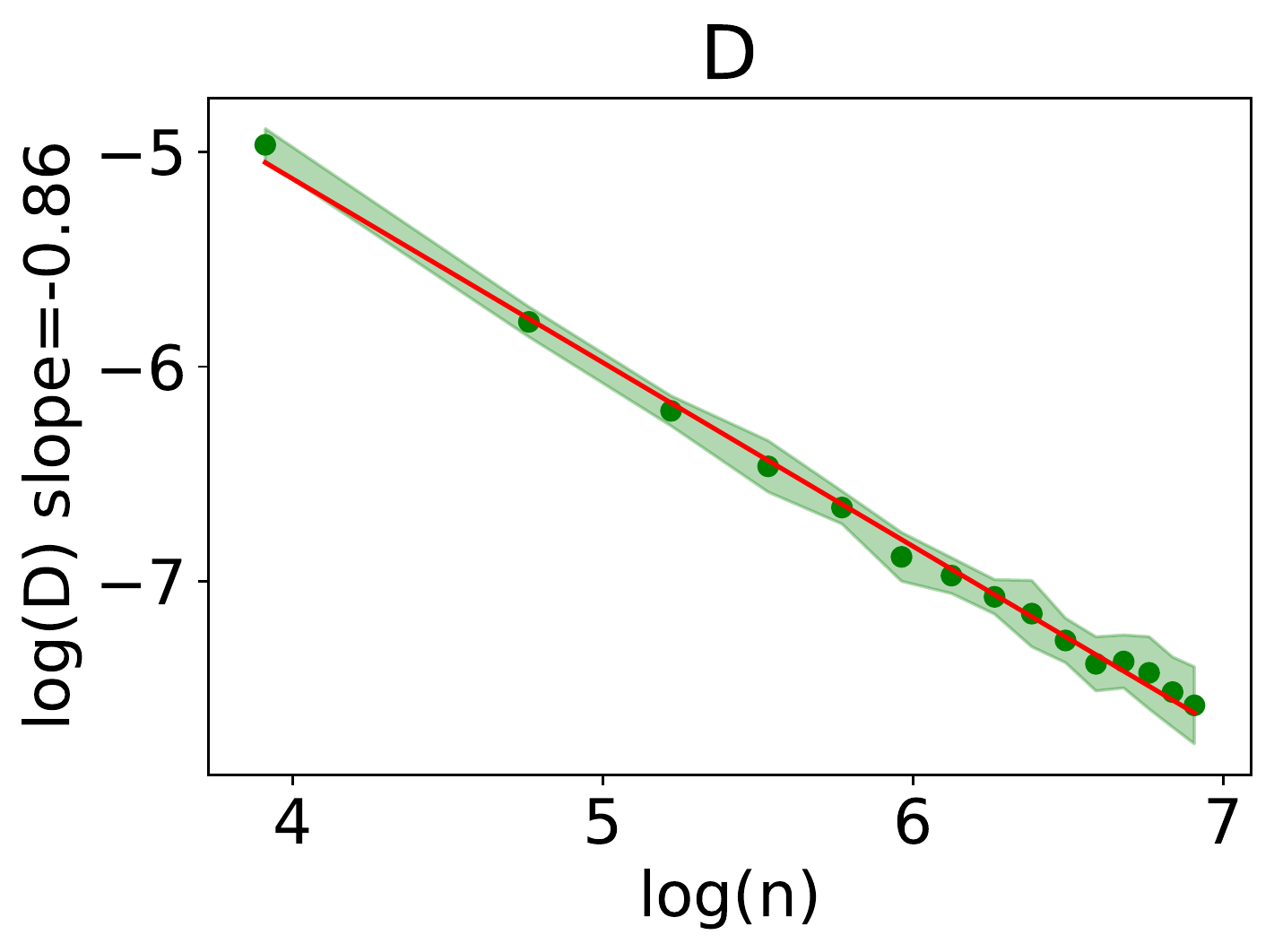}
    \includegraphics[width=0.19\textwidth]{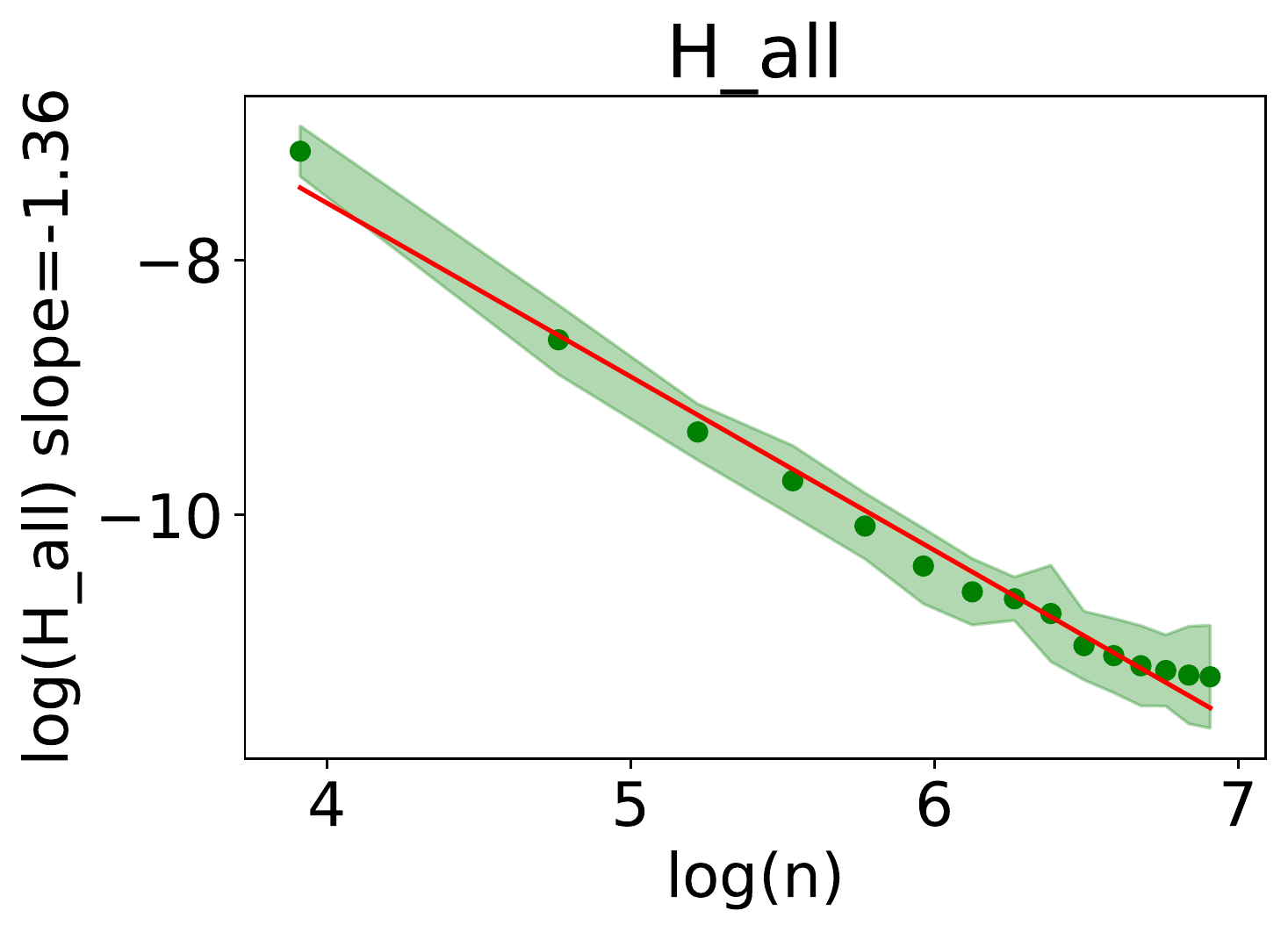}
    \includegraphics[width=0.19\textwidth]{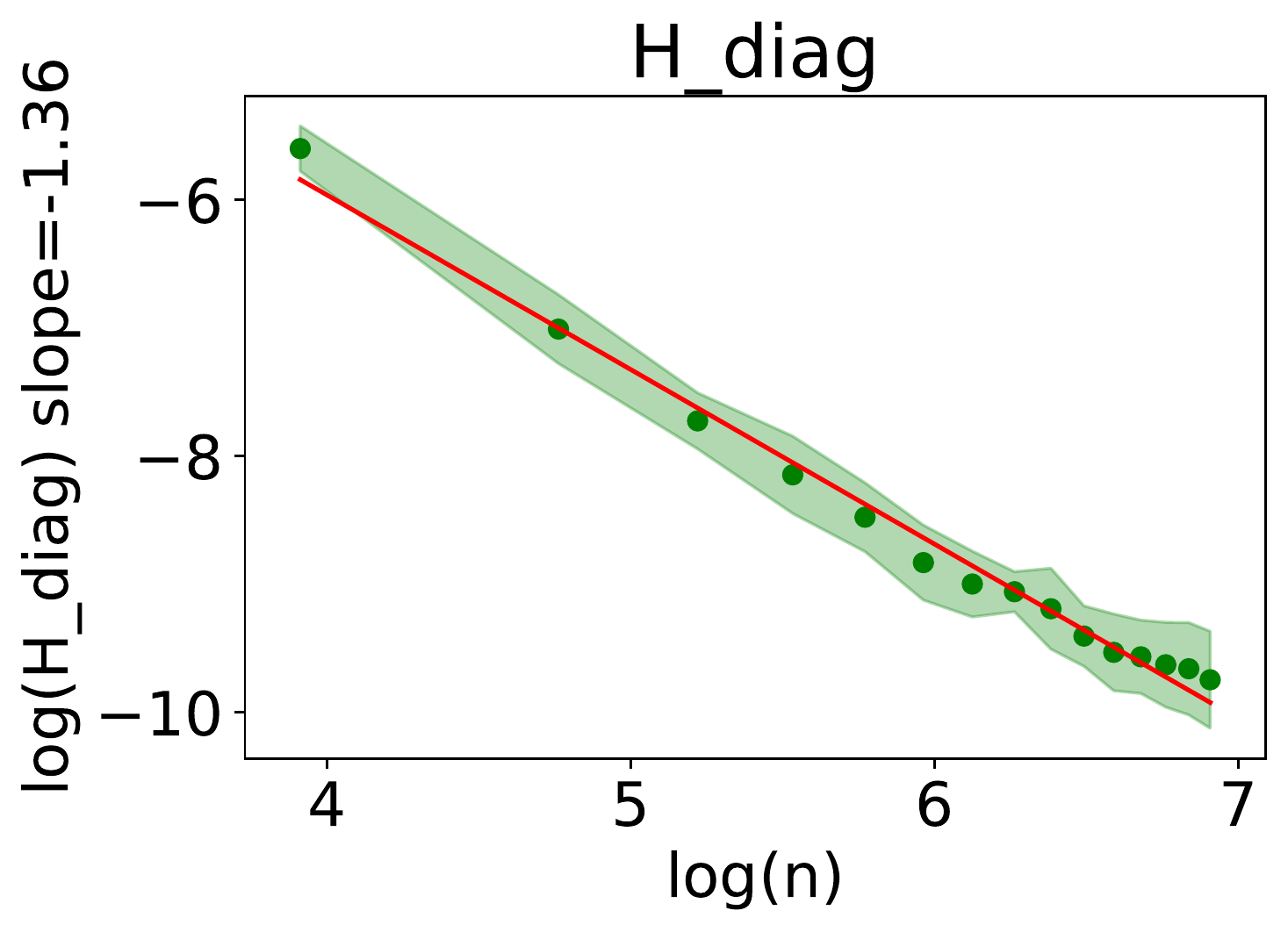}
    \includegraphics[width=0.19\textwidth]{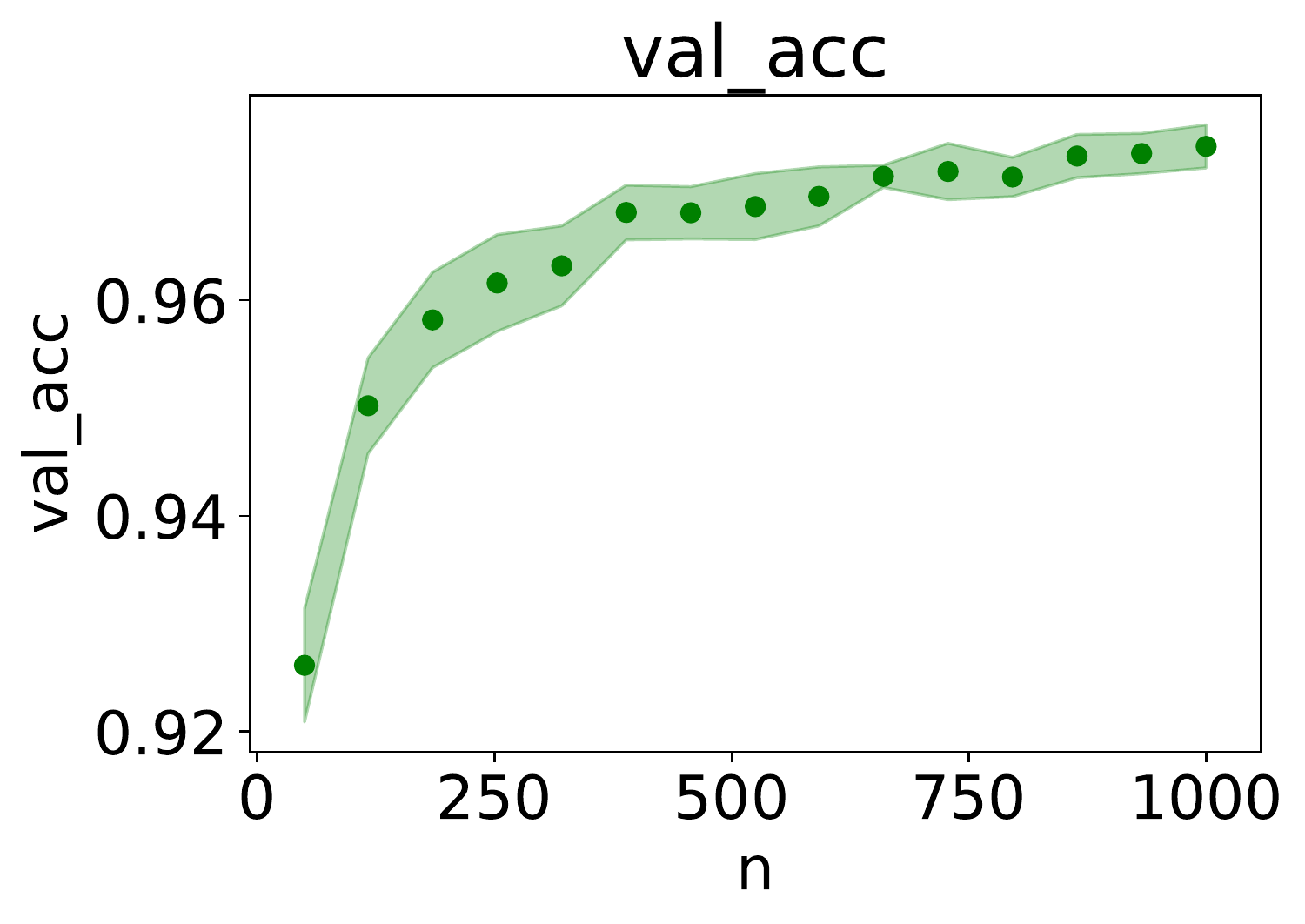}
    \includegraphics[width=0.19\textwidth]{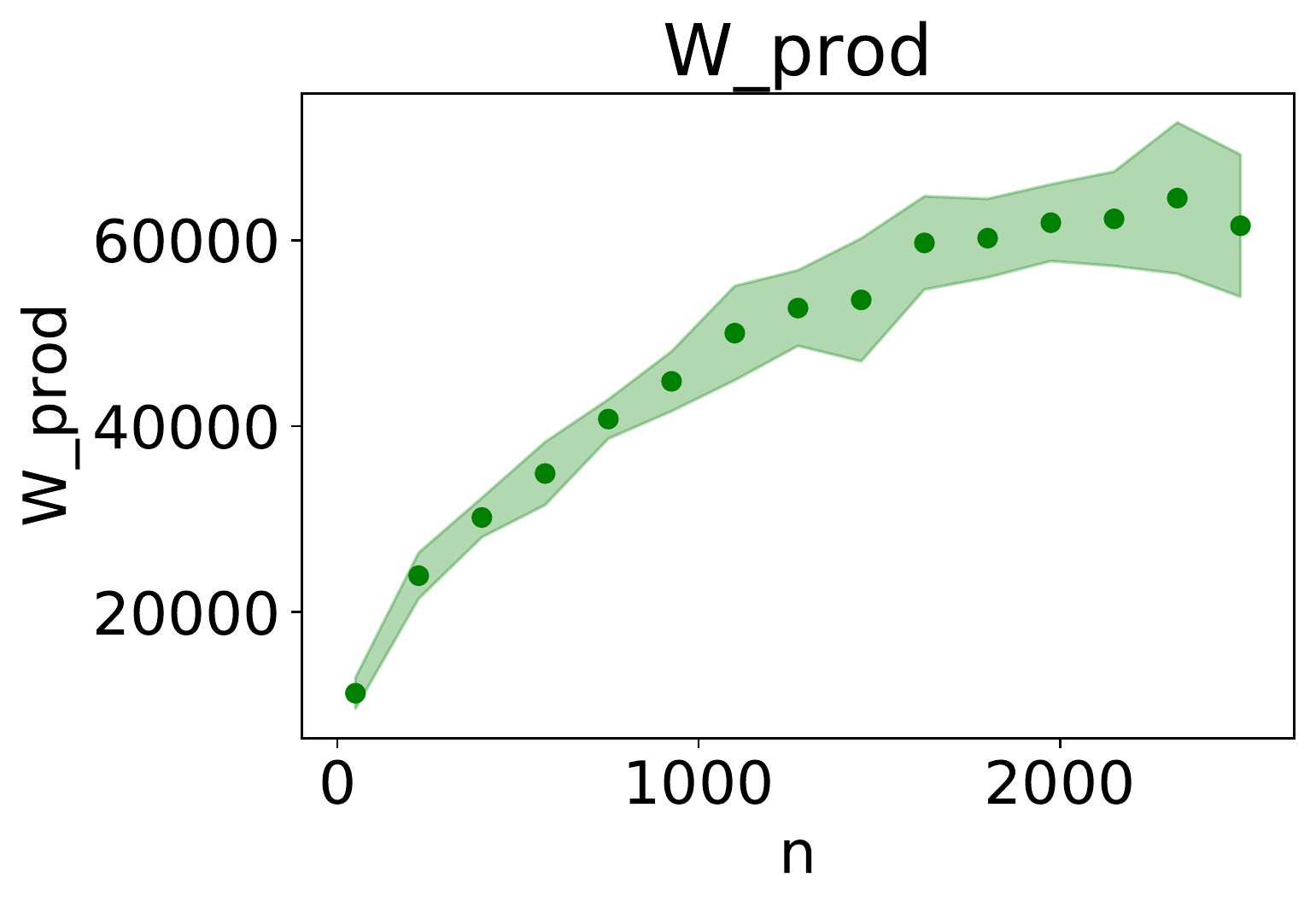}\\
    \includegraphics[width=0.99\textwidth]{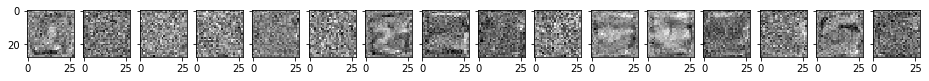}\\
    
    \includegraphics[width=0.19\textwidth]{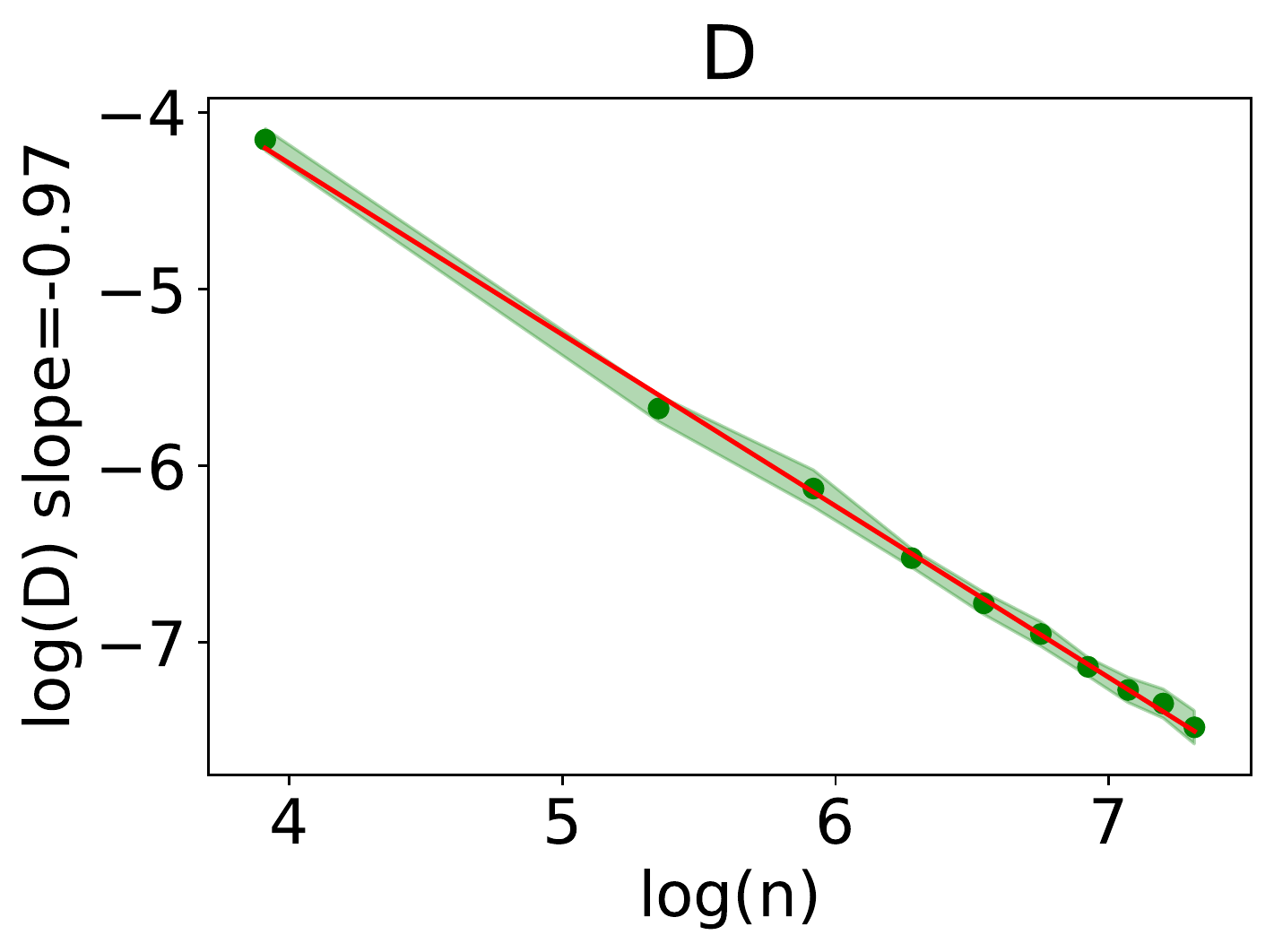}
    \includegraphics[width=0.19\textwidth]{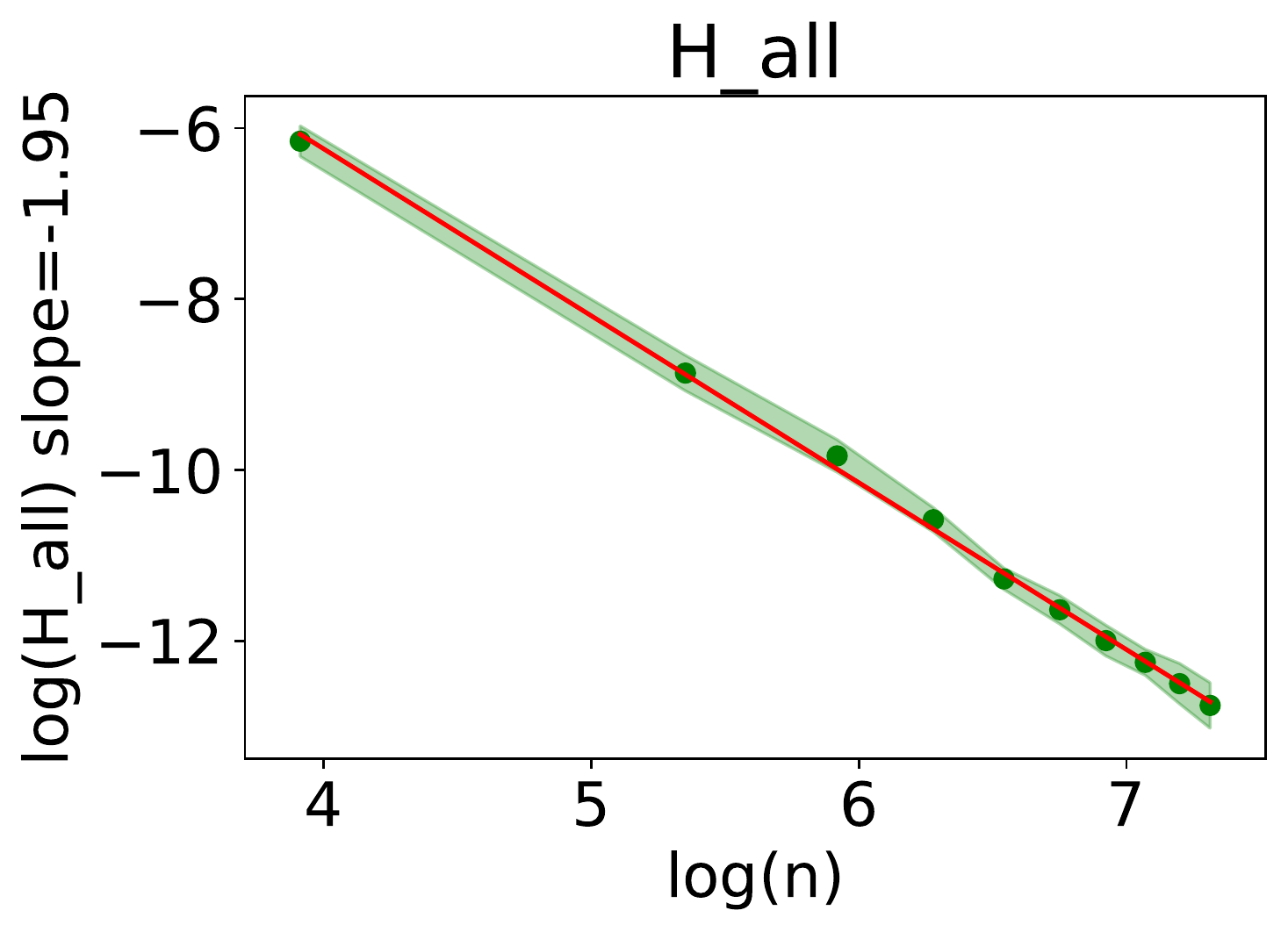}
    \includegraphics[width=0.19\textwidth]{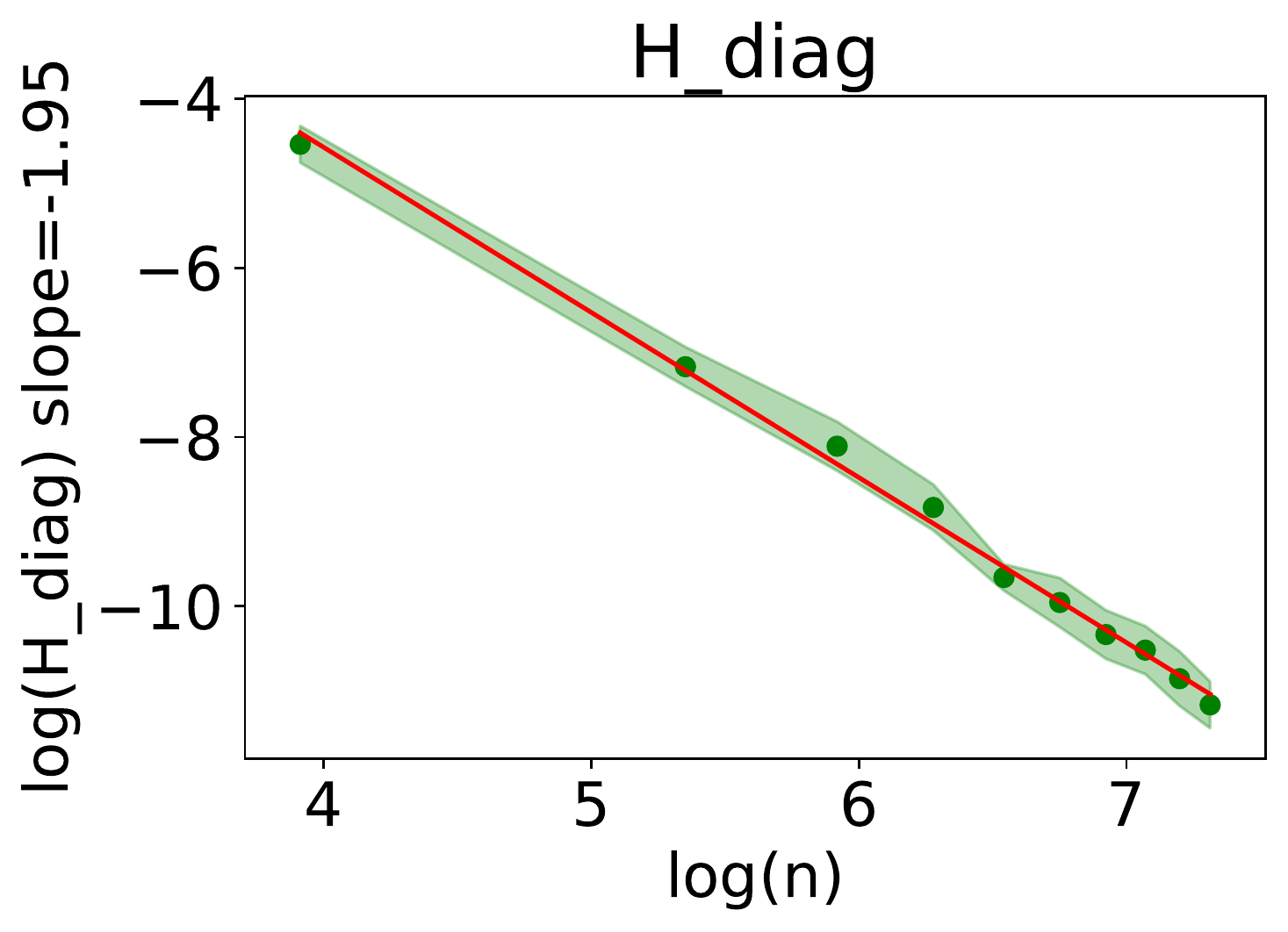}
    \includegraphics[width=0.19\textwidth]{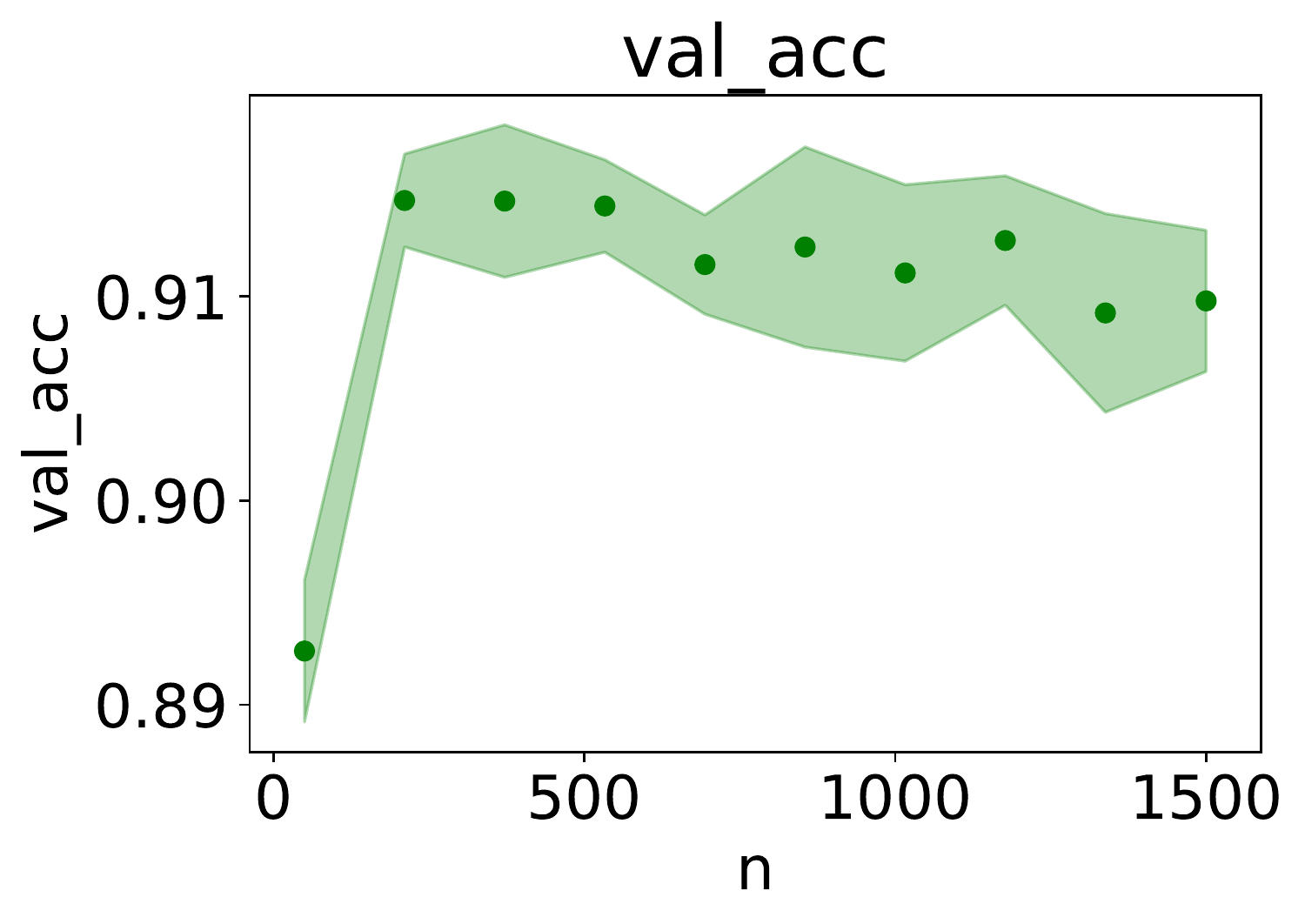}
    \includegraphics[width=0.19\textwidth]{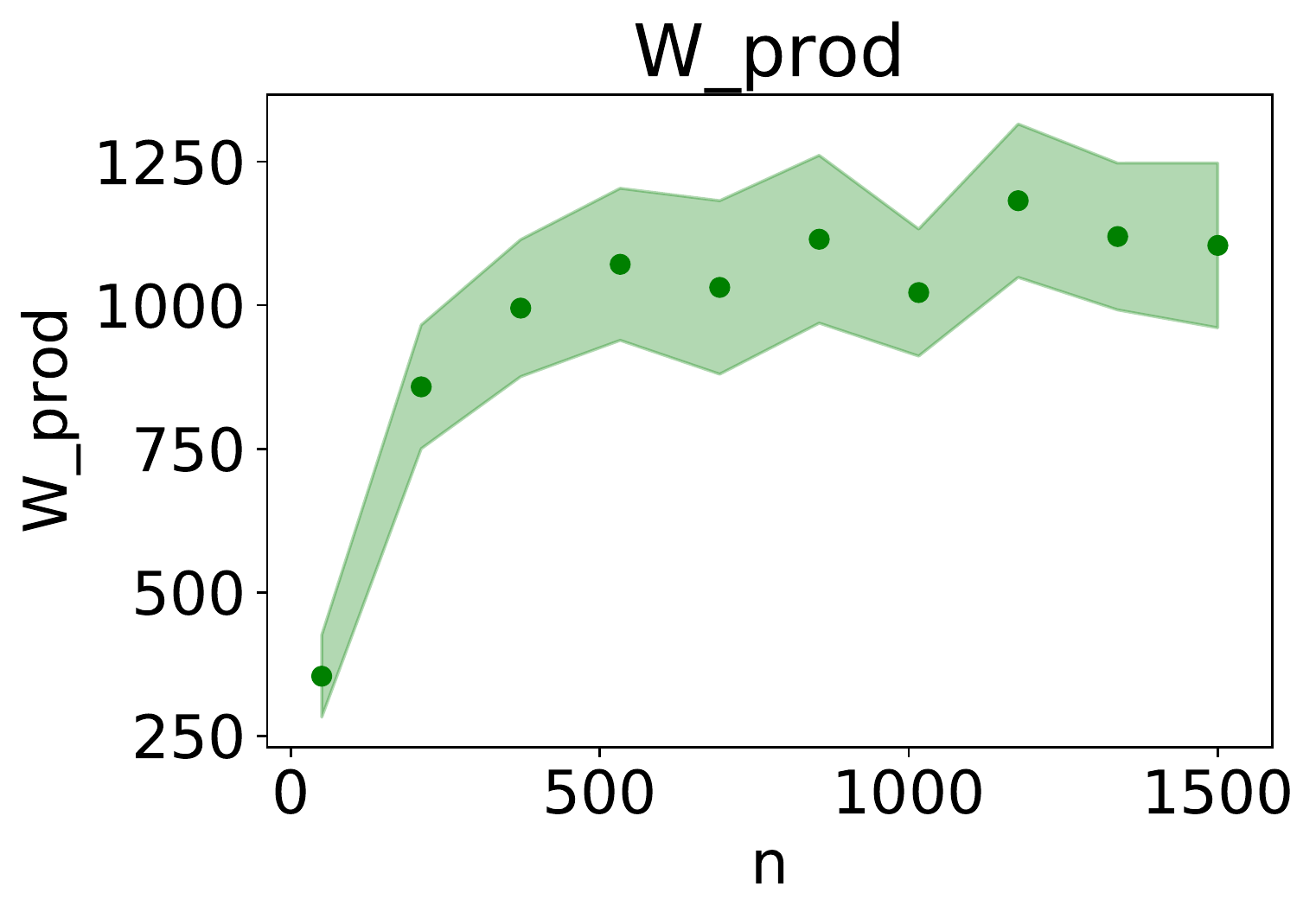}\\
    \includegraphics[width=0.99\textwidth]{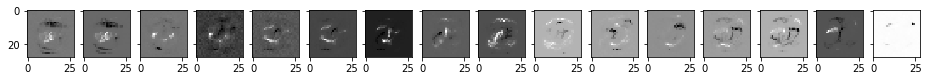}\\

    \caption{Non-regularized MNIST (first two rows), regularized MNIST (second two rows). Odd rows show the charts of the decay of first derivatives (D) and second derivatives (average over all $i,j$ Hessian's components $|H_{ij}|$, and only the diagonal elements $|H_{ii}|$). Next, validation accuracy is shown as well is the product of the infinity norms of the weights $\|W_L\|_{\infty}\cdot...\cdot\|W_1\|_{\infty}$. Even rows show first-layer weights as images}
    \label{fig:decay_mnist}
\end{figure}

\begin{figure}[thb]
    \centering
    \includegraphics[width=0.19\textwidth]{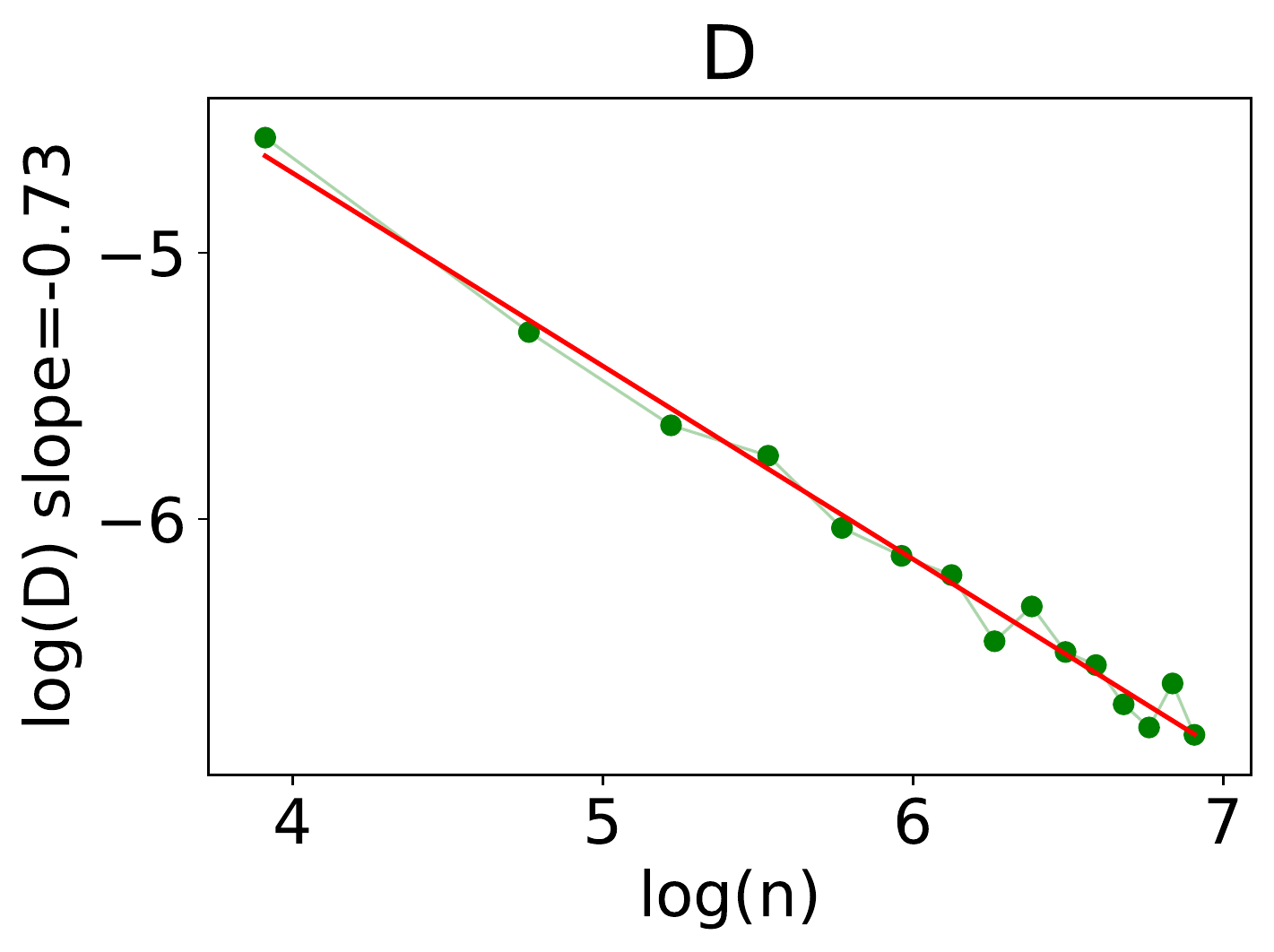}
    \includegraphics[width=0.19\textwidth]{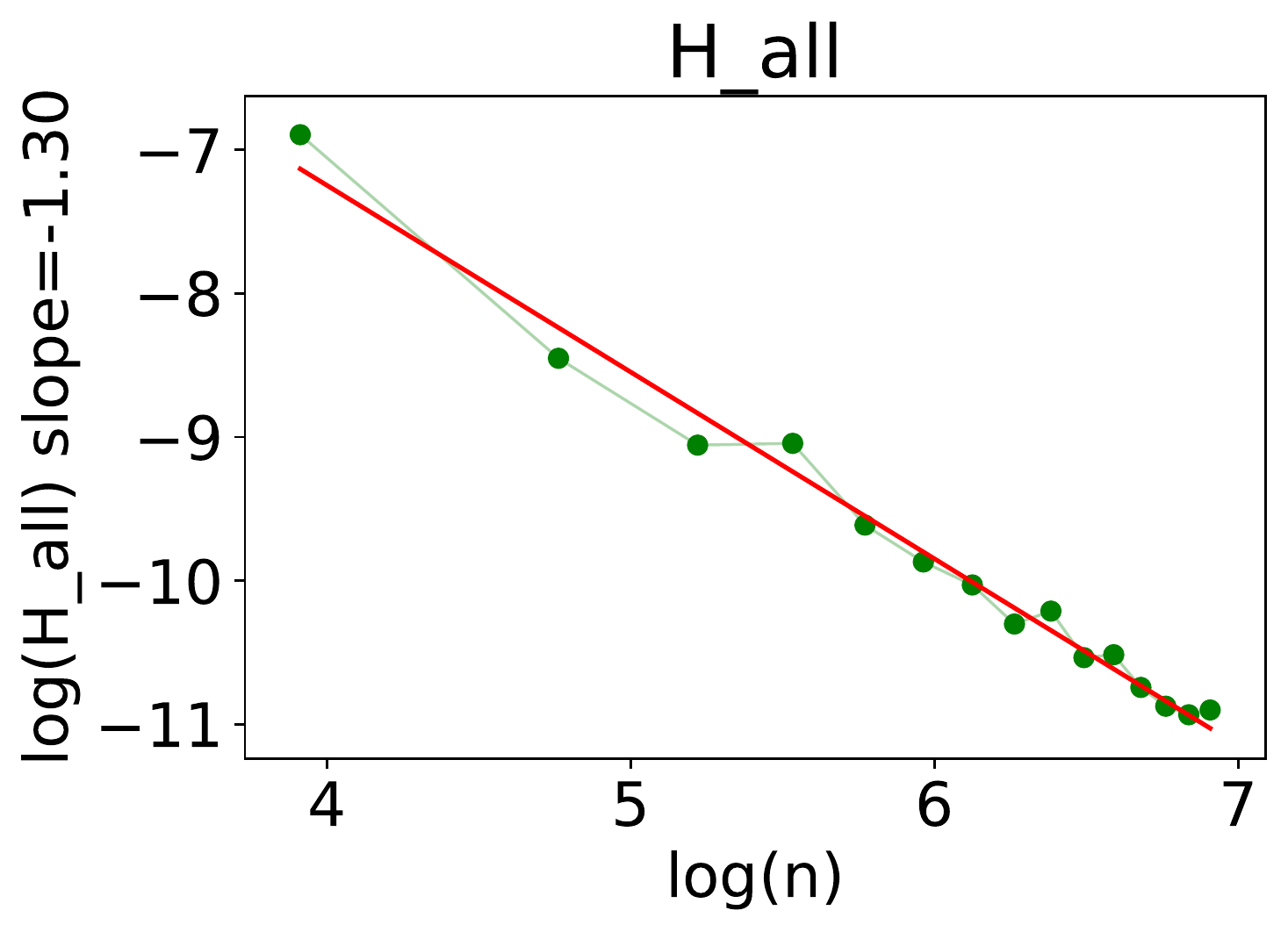}
    \includegraphics[width=0.19\textwidth]{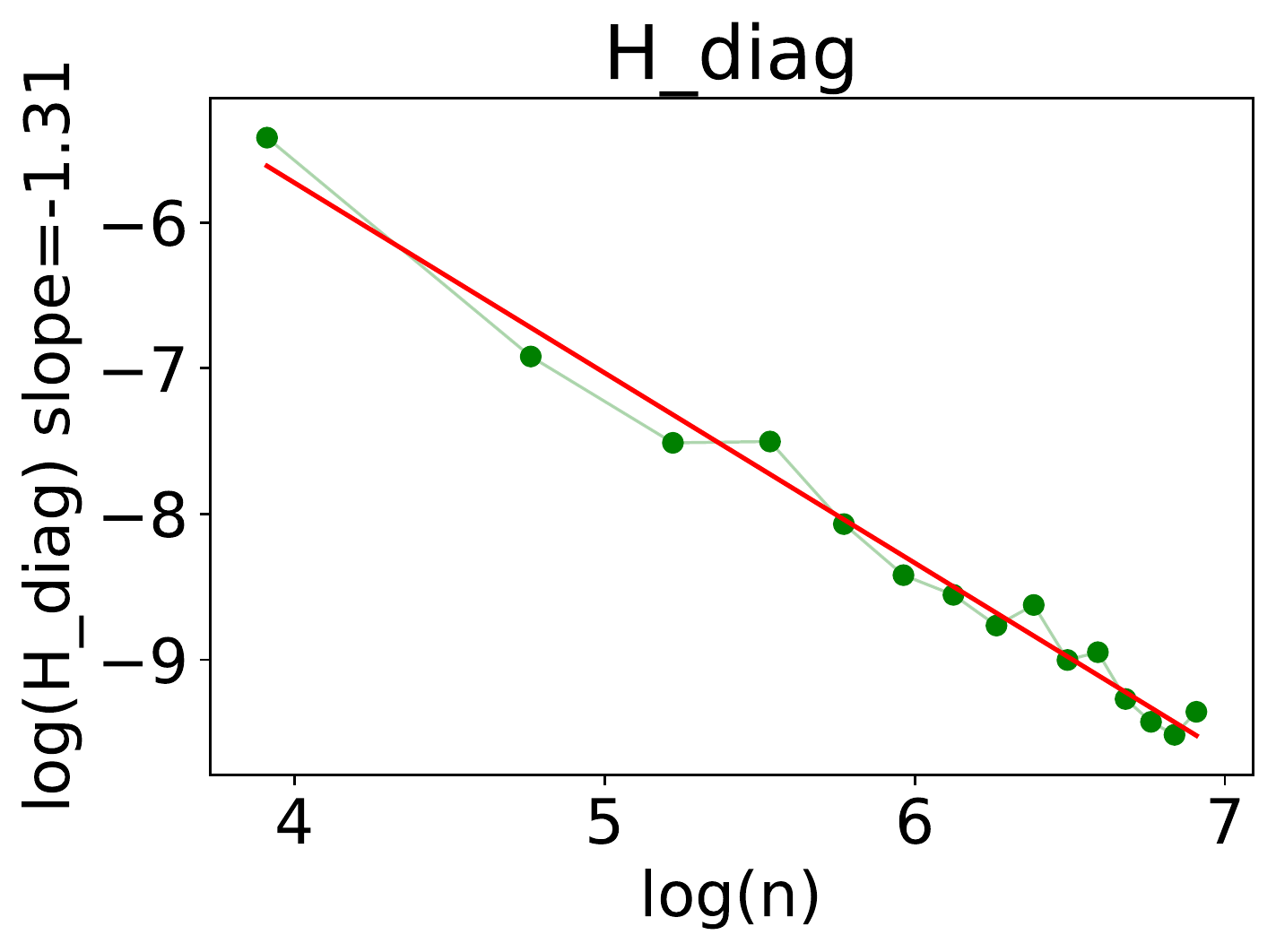}
    \includegraphics[width=0.19\textwidth]{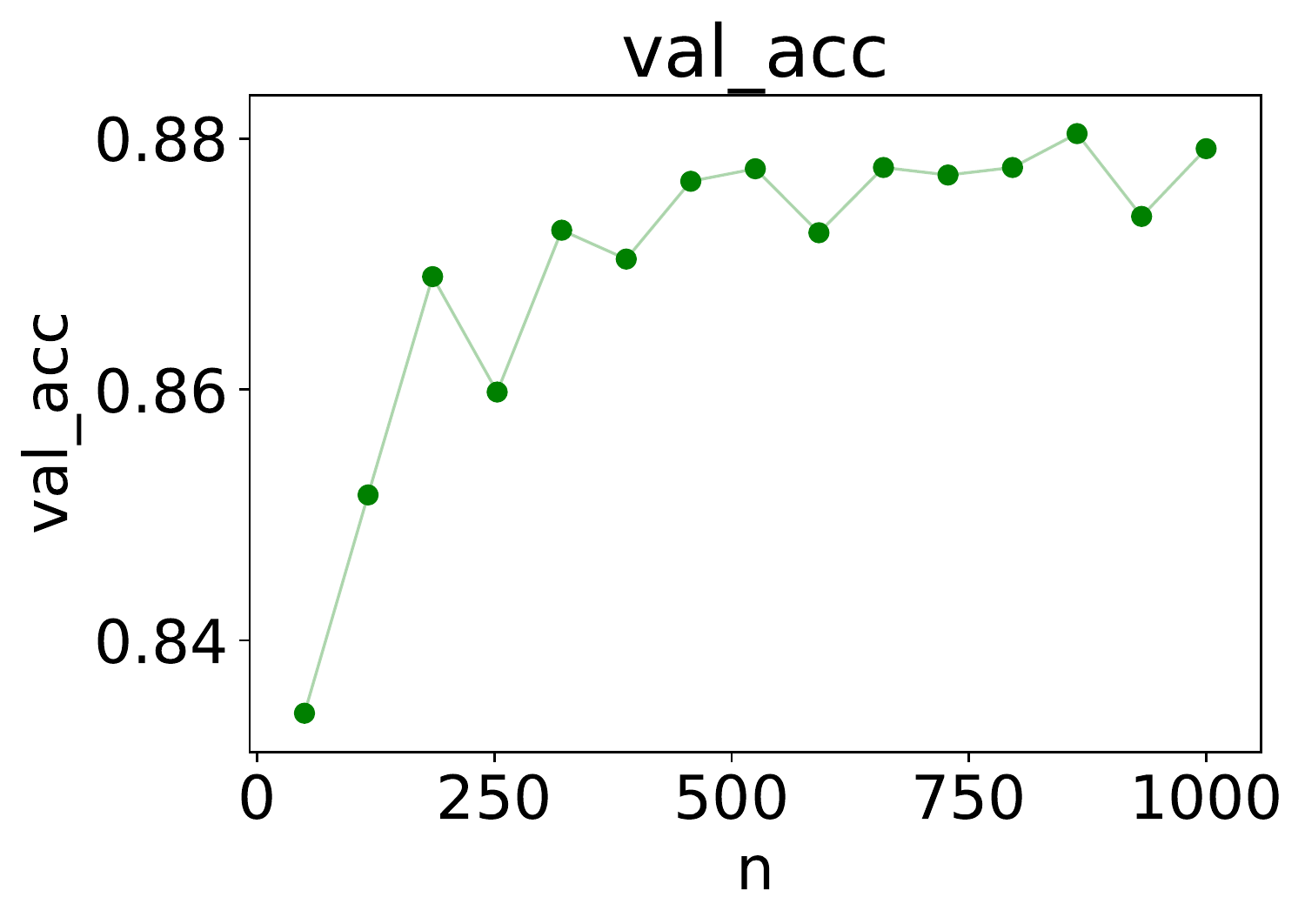}
    \includegraphics[width=0.19\textwidth]{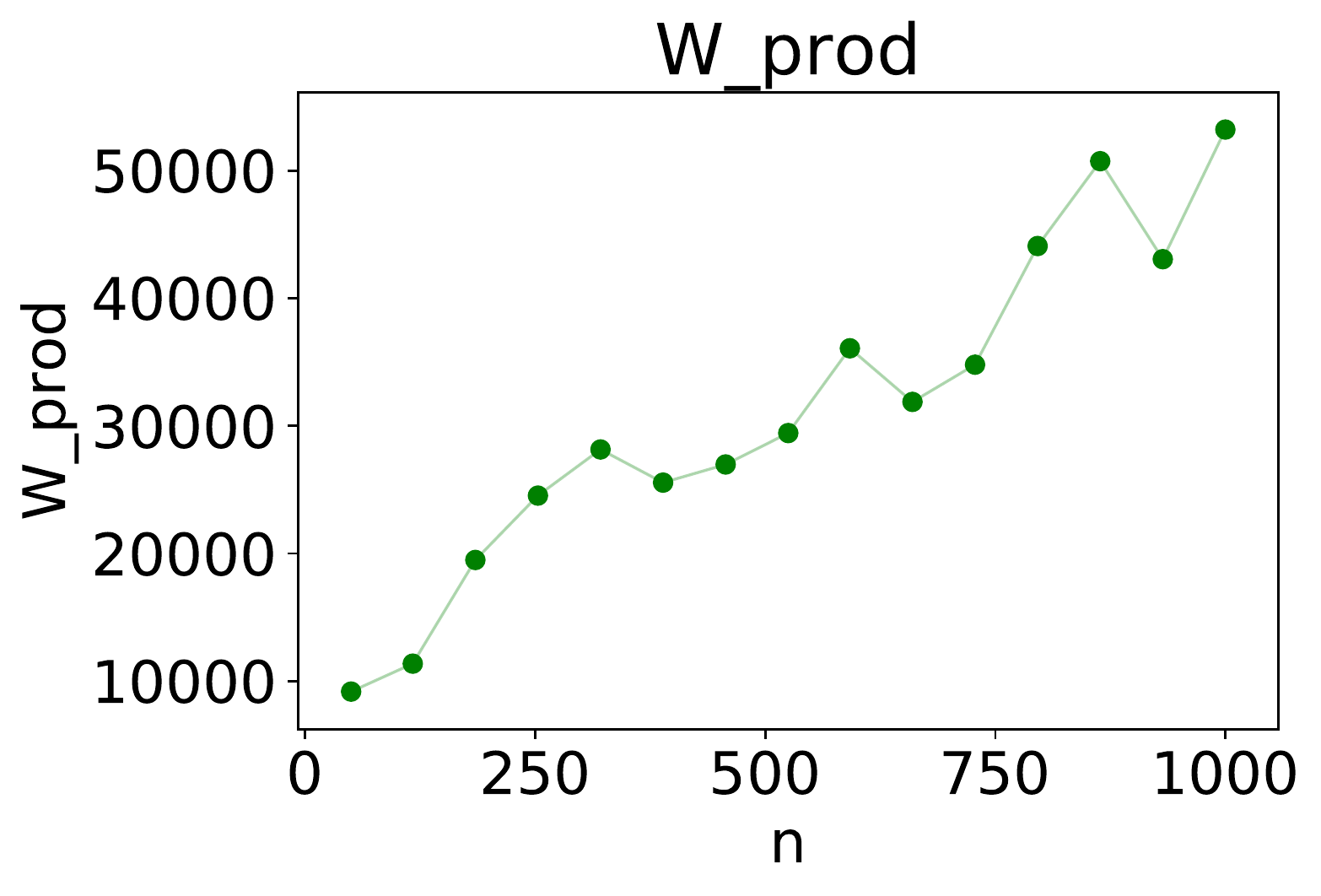}\\
    \includegraphics[width=0.99\textwidth]{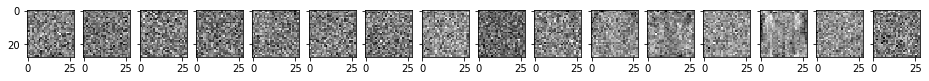}\\
    
    \includegraphics[width=0.19\textwidth]{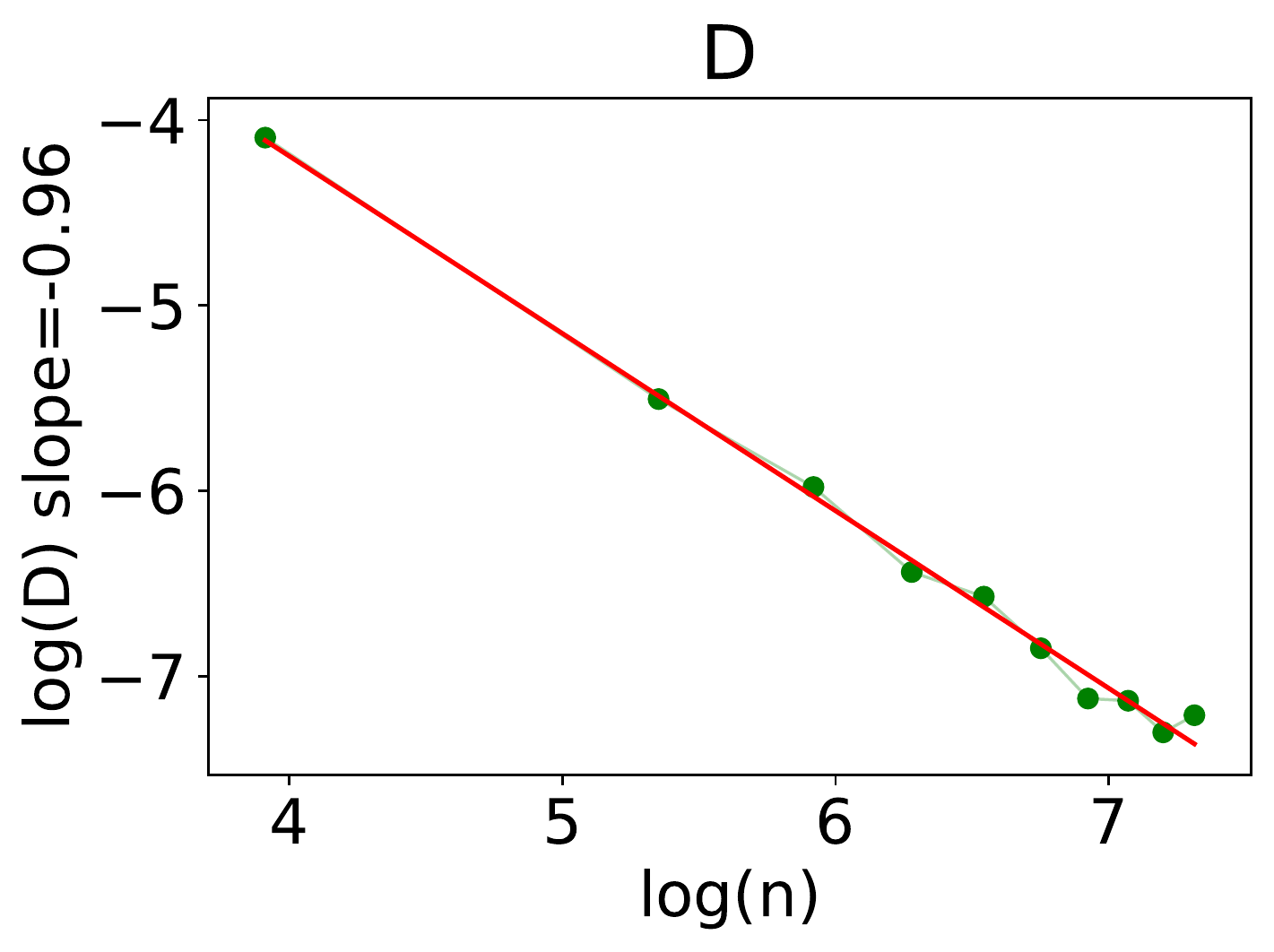}
    \includegraphics[width=0.19\textwidth]{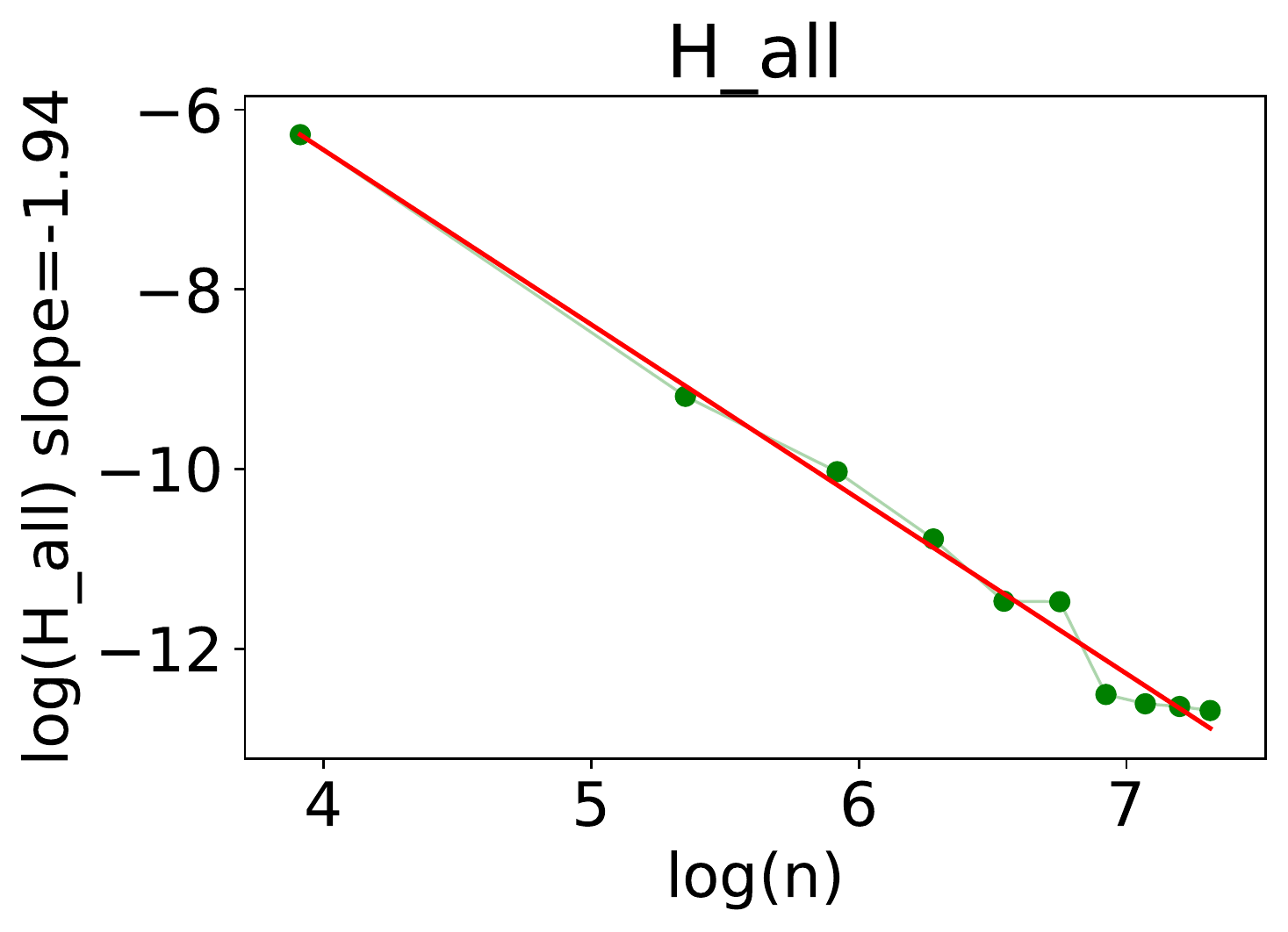}
    \includegraphics[width=0.19\textwidth]{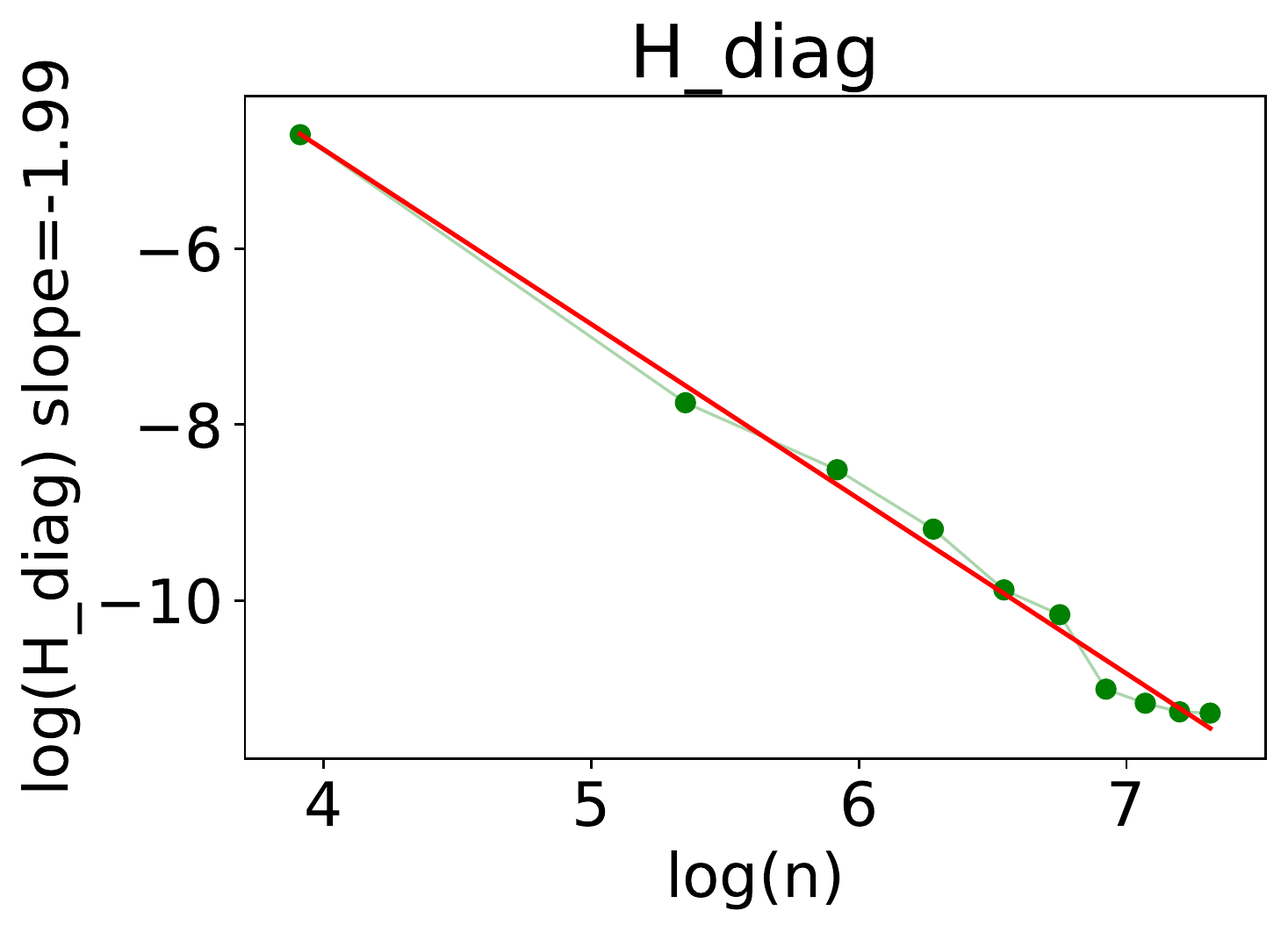}
    \includegraphics[width=0.19\textwidth]{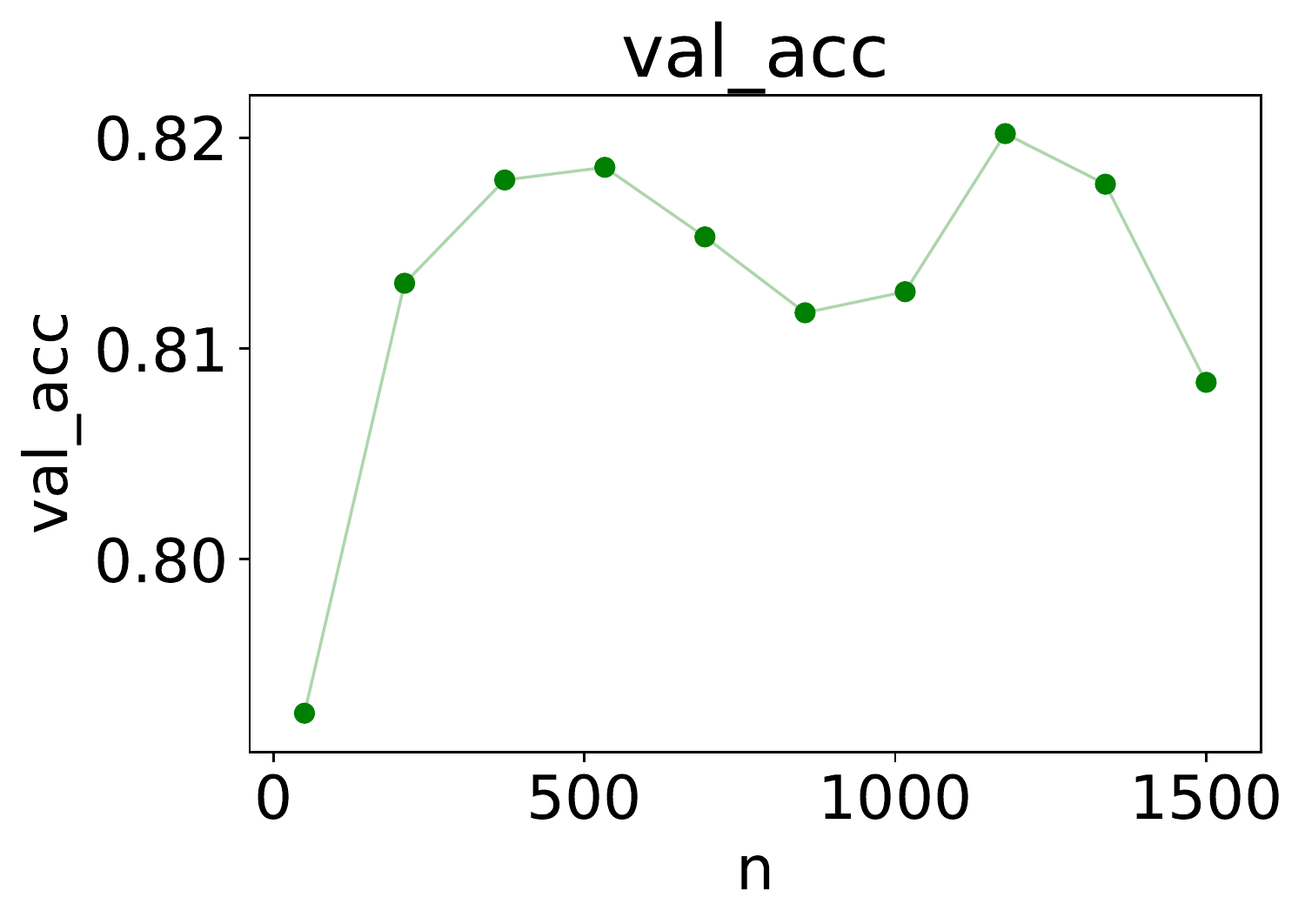}
    \includegraphics[width=0.19\textwidth]{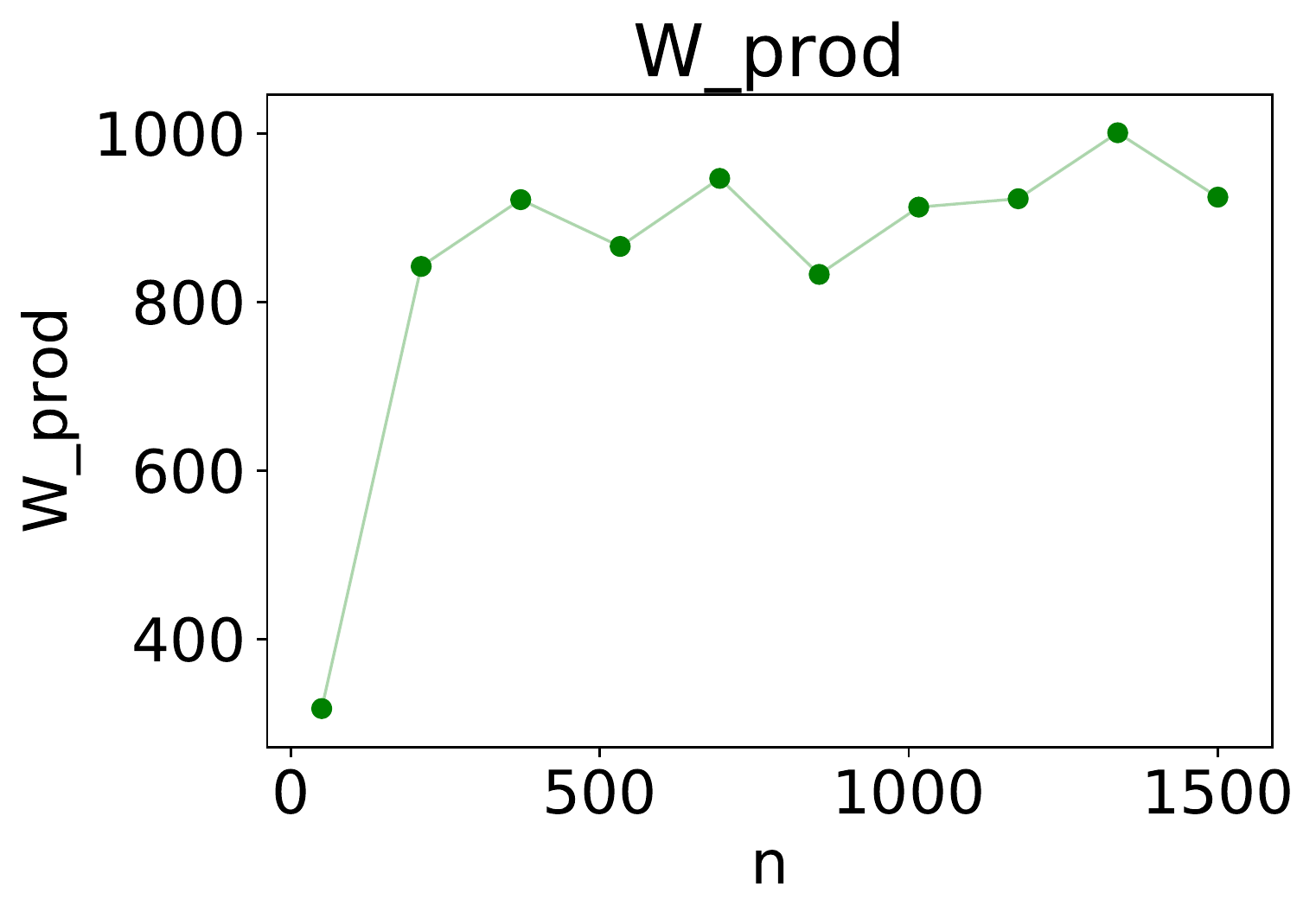}\\
    \includegraphics[width=0.99\textwidth]{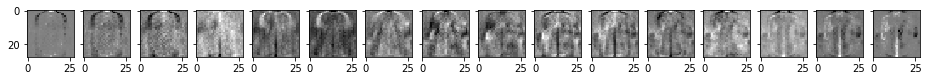}
    
    \caption{Non-regularized Fashion MNIST (first two rows), regularized Fashion MNIST (second two rows). Description is the same as in Fig. \ref{fig:decay_mnist}}
    \label{fig:decay_fashion_mnist}
\end{figure}

\begin{figure}[thb]
    \centering
    \includegraphics[width=0.19\textwidth]{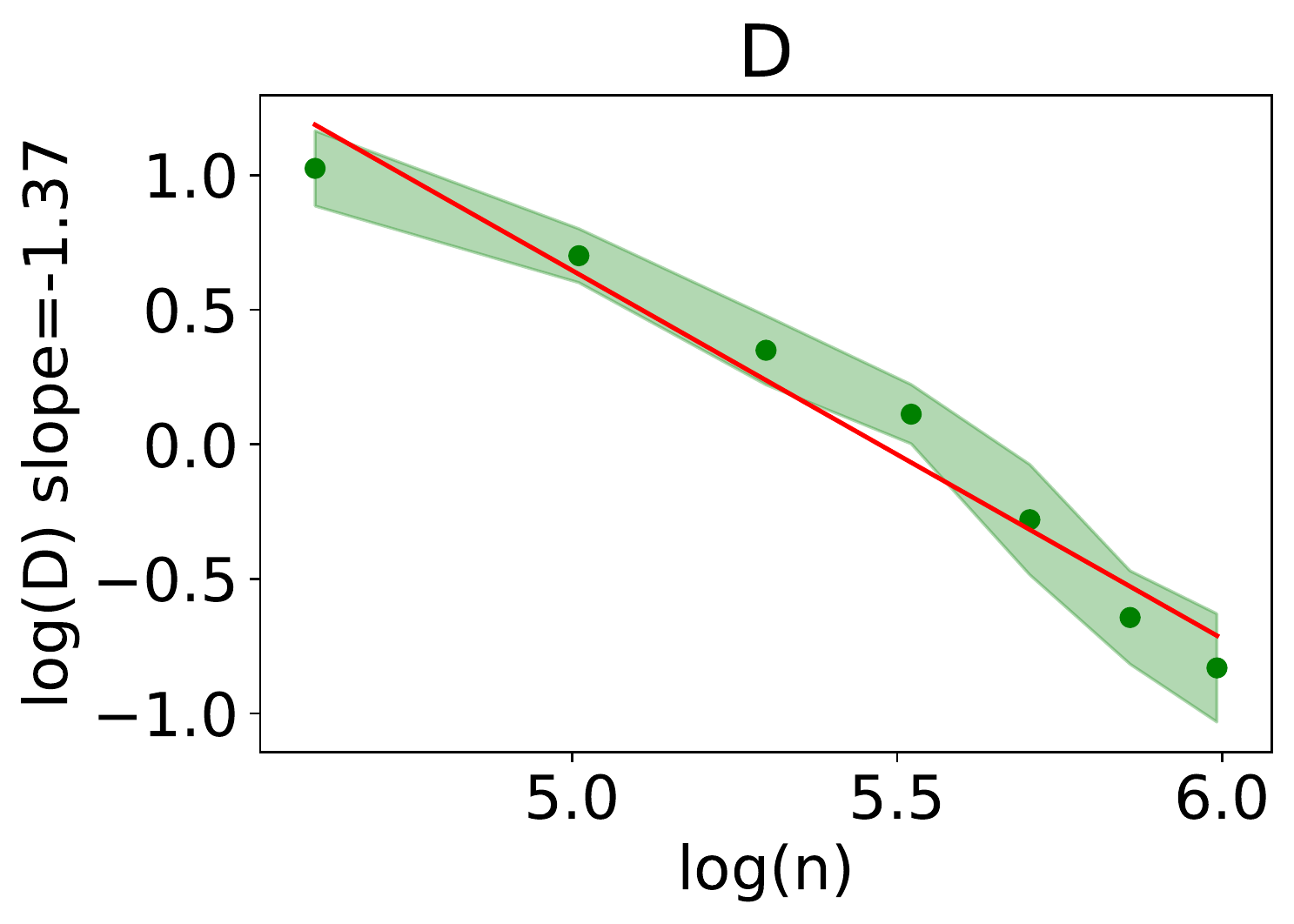}
    \includegraphics[width=0.19\textwidth]{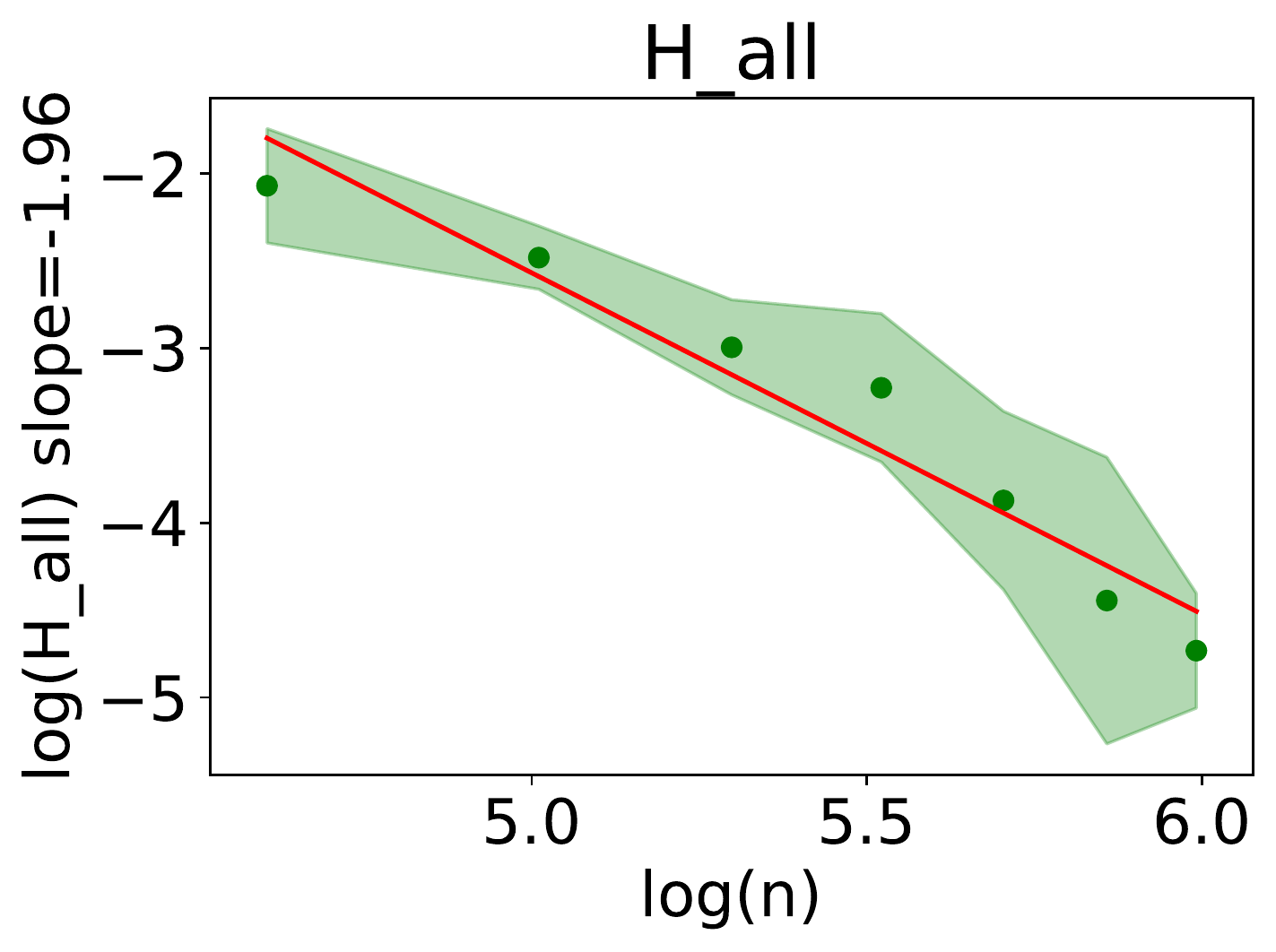}
    \includegraphics[width=0.19\textwidth]{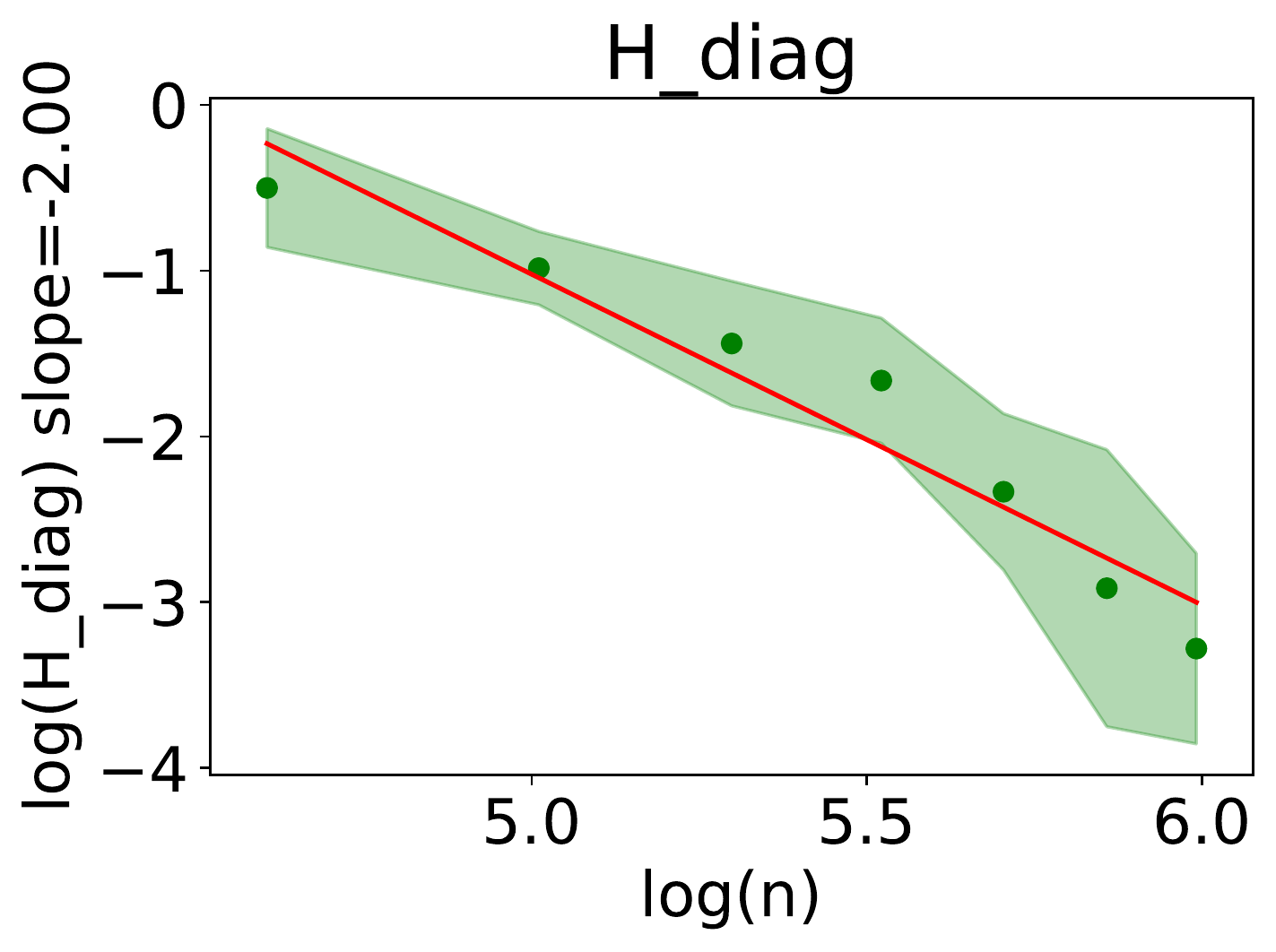}
    \includegraphics[width=0.19\textwidth]{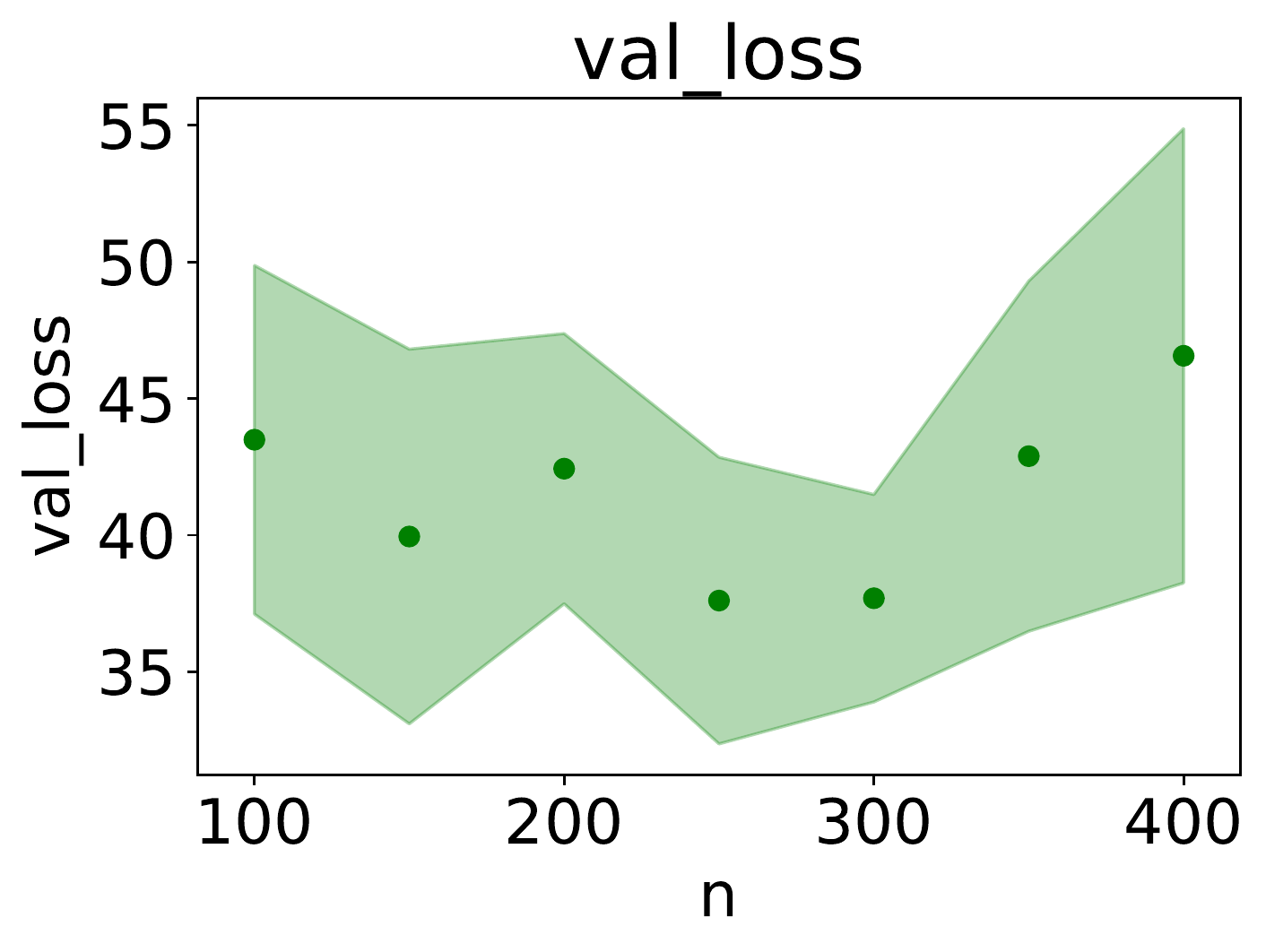}
    \includegraphics[width=0.19\textwidth]{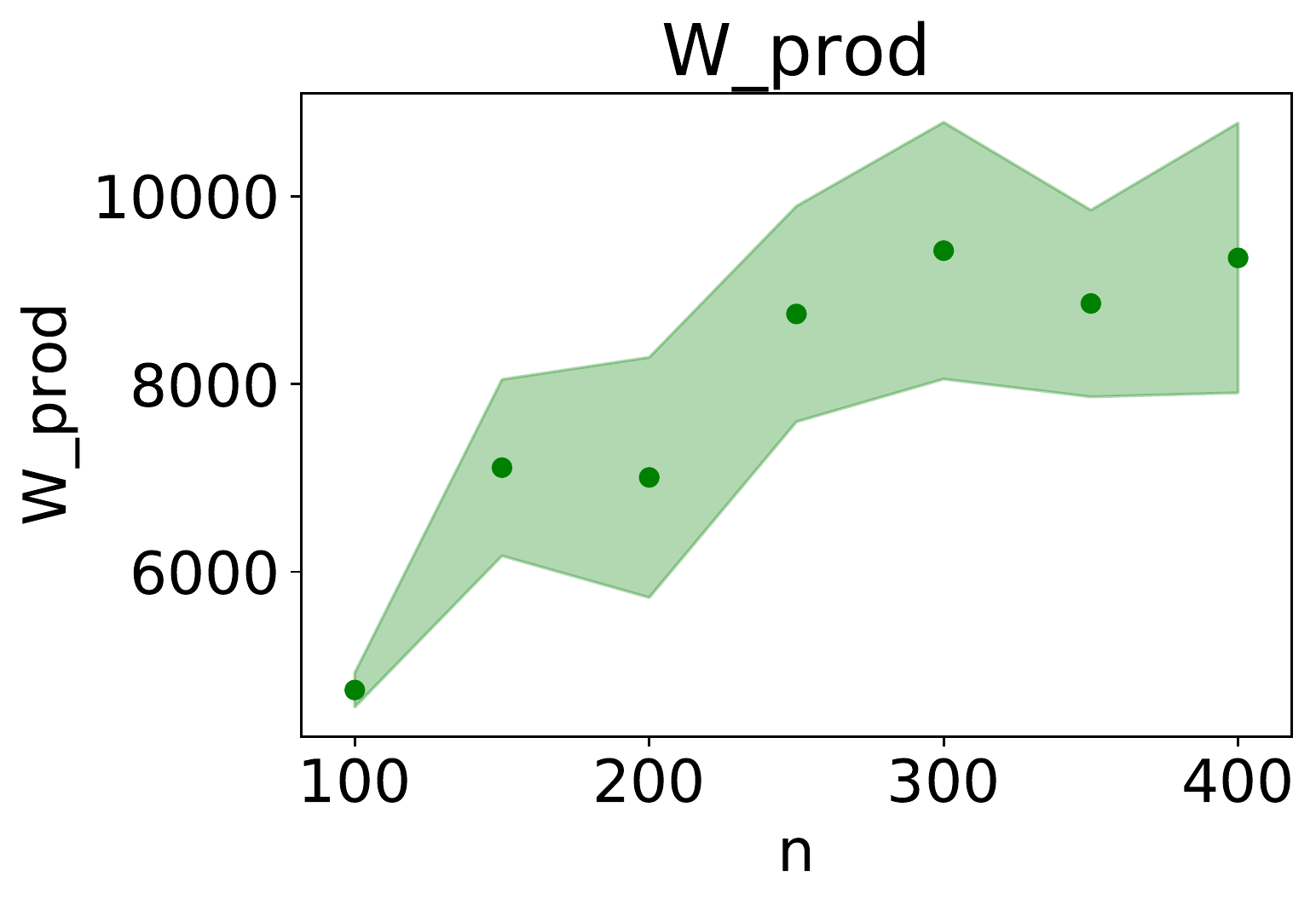}

    \caption{Non-regularized Boston Housing dataset}
    \label{fig:decay_boston}
\end{figure}

\begin{enumerate}

\item {\bf Explicit duplication.} We copy each neuron multiple times and reduce the outgoing weights. If we set $W(t,t')$ to be piecewise-constant, then the approximation error is zero $A=0$, and it does not depend on the degree of duplication. This is the obvious case where the network is becoming more fault-tolerant using duplication, and our framework confirms that. The problem with explicit duplication of non-regularized networks is that their fault-tolerance is suboptimal. Not all neurons are equally important to duplicate. Thus, it's more efficient to utilize all the neurons in the best way for fault tolerance by duplicating only the important ones.

\item {\bf Explicit regularization.} We make adjacent neurons compute similar functions, thus, allowing for redundancy. We first consider the local "number of changes" metric (Table \ref{tab:continuous_discrete_correspondence}). Specifically, for some function $z(t)\in[0,1]$, $\int |z'(t)|dt$ represents how many times the function goes fully from $0$ to $1$ or vice versa. Our idea is to use that for the weights to quantify their discontinuity: 

\begin{align*}
C_1=\sum\limits_{ij}|W_l^{ij}-W_l^{i+1,j}|&=\frac{1}{n_0}\sum\limits_{ij}|W(\hat{i},\hat{j})-W(\hat{i+1},\hat{j})|\\
&=\frac{1}{n_0}\sum\limits_{ij}\left|W\left(\frac{i-1}{n_1-1},\frac{j-1}{n_0-1}\right)-W\left(\frac{i}{n_1-1},\frac{j-1}{n_0-1}\right)\right|\\
&\approx\frac{1}{n_0n_1}\sum\limits_{ij}\left|W'_{t}\left(\frac{i-1}{n_1-1},\frac{j-1}{n_0-1}\right)\right|\\
&\approx\int |W'_{t}(t,t')|dtdt'
\end{align*}

The term above, if small, guarantees that, for each {\em input} neuron, neighboring output neurons will use it in similar ways.
The same is applied for $W^T$ as well: $C_2=n_0/n_1\sum_{ij}|W^{ij}-W^{i,j+1}|\approx \int |W'_{t'}(t,t')|dtdt'$ which guarantees that, for each {\em output} neuron, neighboring input neurons are used similarly by it.

In addition to making adjacent neurons computing similar functions, we add another term $C_3$ by Gaussian smoothing: a Gaussian kernel is convolved with the weights, and the weights are subtracted from the result. The difference shows how much the current value differs from an aggregated local "mean" value.

We explicitly enforce the continuous limit by adding a regularization term of $smooth(W):=C_1+C_2+C_3$. Here $smooth$ consists of three parts, see {\tt FilterPlayground.ipynb }. The first part $C_1, C_2$ computes the numerical derivative with $2, 4, 10, 14$ points in the array. The second part convolves the input with a Gaussian kernel and subtracts the original, resulting in a measure of discontinuity. All metrics are normalized to work with any size, so that scaling the network up does not change the magnitude of $C_i$ significantly. The implementation can be found in {\tt continuity.py}.

We check the derivative decay prediction (P\ref{proposition:decay}) which must follow from the continuity assumption A\ref{assumption:continuous_net}, experimentally. We run experiments for the MNIST dataset and architecture $(784, N, 100, 10)$ -- 2 hidden layers with sigmoid activations, unit Lipshitz coefficient, batch size of 1000, and $N$ from 50 to a few thousand. We measure $\|W\|_{\infty}=\max_i\sum_j|W^{ij}|\approx \max_t\int_{t'}|W(t,t')|dt'$ which should stay constant (we measure the product of these over layers $l$, as only the total norm is expected to stay constant, whereas individual layers can change the magnitude of $W$). We also measure the components of the first derivative $avg(|D_i|)$ and the components of the Hessian $avg_{i}(|H_{ii}|)$, $avg_{(ij)}(|H_{ij}|)$ which we expect to decay with $n$ as by P\ref{proposition:decay}. We show that without regularization, the weights $\|W\|_{\infty}$ increase, see {\tt WeightDecay-FC-MNIST.ipynb}, and the derivatives and Hessians do decay but with a smaller rate, insufficient for T\ref{prop:taylor} to work, see {\tt DerivativeDecay-FC-MNIST.ipynb}. We repeat each experiment $10$ times and report mean and standard deviation, see Figure \ref{fig:decay_mnist}. In contrast, our proposed regularization results in a smooth transition between neurons at each layer. They are grouped by their similarity. This can be visually seen in {\tt WeightDecay-Continuity-FC-MNIST.ipynb} or in Figure \ref{fig:decay_mnist}. Contrary to networks without regularization, first layer weight profiles seem to have a meaningful image, and these images are similar for close-by neurons. There is also an increase in continuity of activations for a fixed input. For this regularization technique, we show that the product of the weights stabilizes, and the derivatives decay with proper slopes of $-1$ and $-2$. The accuracy, however, drops from 98\% to 90\%. We note that this does not demonstrate that our approach necessarily leads to a decrease in accuracy, as continuous networks are general by AP\ref{prop:completeness}.

We test the result on Fashion MNIST as well (Figure \ref{fig:decay_fashion_mnist}, notebook names are the same but with {\tt Fashion}) and report similar behavior. On the Boston Housing dataset, non-regularized networks already have the desired characteristics (Figure \ref{fig:decay_boston}).

\begin{flushright}
\begin{sidenote}
\paragraph{\bf Note:} the condition above of $C_1+C_2+C_3$ being small is, strictly speaking, only a {\em necessary} condition for A\ref{assumption:continuous_net}, but not a sufficient one. Even if networks are smooth enough, they might not have a limit $NN_n\to NN_c$, as they could implement the function in drastically different ways: for example, the networks $NN_n$ can be all smooth, but approximate {\em different} continuous functions. However, this condition {\em is} sufficient for the derivatives $D_k$ to stay constant (second part of A\ref{assumption:continuous_net}), which is the only requirement for T\ref{prop:taylor} to work. \footnote{A discussion of why $D_k$ stay constant is given in the analysis of the correctness for Algorithm 1.} Thus, our approach can give a formal guarantee of fault tolerance. An attempt to give a {\em truly} sufficient condition for A\ref{assumption:continuous_net} would be to train the bigger network given duplicated weights of a smaller network, and penalizing a bigger network from having weights different from their initialization.

{\bf Different smoothing techniques.} Currently, $C_2$ is unused in our implementation, as it was sufficient to use $C_1+C_3$ to achieve the fault tolerance required in our experiments. Another method to make close-by neurons similar could be to use some clustering loss in the space of weights accounting for locality, like the Kohonen self-organizing map. One more idea is to regularize the weights with a likelihood of a Gaussian process, enforcing smoothness and a required range of dependencies. Another idea is to use the second derivative, which is connected to the curvature of a curve $(x, y(x))$ in the 2D space: $\kappa=y''/(1+(y')^2)^{3/2}$. The interpretation here is that $\kappa=1/R$ for $R$ being the radius of curvature, a geometrical property of a curve showing how much it deviates from being a line.
\end{sidenote}
\end{flushright}

\item {\bf Convolutional networks.} For images, this limit naturally corresponds to scaling the images \cite{guss2016deep} with intermediary pixels being just added between the original ones. Images are piecewise-continuous because adjacent pixels are likely to belong to the same object having the same color, and there are only finitely many objects on the image. Convolutional networks, due to their locality, fit our assumption on one condition. We need to have a large enough kernel size, as otherwise the non-smoothness is high. Specifically, for CNNs, $C_1$ is small, as neighboring neurons have similar receptive fields. In contrast, $C_2$ can be large in case if the kernel size is small: for example, the kernel $(-1,1)$ in the 1D case will always result in a high discontinuity: the coefficients are vastly different and require more values in between them to allow for a continuous limit and redundancy.

The notebook {\tt ConvNetTest-VGG16-ManyImages.ipynb} investigates into this. We note that we do not present this result in the main paper and make it a future research direction instead. In this paper, we give qualitative description of applicability of our techniques to convolutional networks (big kernel size for a small $C_2$ with respect to kernel size, smooth activation and pooling to allow for T\ref{prop:taylor} to work), as it would be out of the scope of a single paper to go into details.

\item {\bf Theoretical considerations for the general case.}
\begin{flushright}
\begin{sidenote}
\paragraph{} How does the network behave when the number of neurons increases, and it is trained with gradient descent from scratch? First, there are permutations of neurons, which we ignore. Secondly, there could be many ways to represent the same function. One constraint is that the magnitude of outputs in the output layer is preserved. Intermediate layers need to have a non-vanishing pre-activation values to overcome vanishing/exploding gradients. In addition, input limit might be enforced such that $x_i\approx x(\hat{i})$. Now, gradient descent results in a discrete network $NN_d$ which can be seen as a discretization of some continuous network $NN_c$. Since NN's derivatives are globally bounded, GD converges to a critical point. Each critical point determines the range of initializations which lead to it, partitioning the whole space into regions. Each fixed point with a sufficiently low loss thus corresponds to a set of continuous networks "passing through" a resulting discrete network. Each of the continuous networks can have different implementations in discrete networks of larger size. We choose a path in networks of different sizes $n$ and denote the probability $s_n$ over initializations to choose that particular continuous network. Therefore, on that path, derivatives decay as we want since $NN_n\to NN_c$. The problem might arise if a particular continuous limit $NN_c$ has an extremely small probability (over initializations) of gradient descent giving $NN_n$: if $s_n\to 0$, this particular network $NN_c$ is unlikely to appear. We leave the study of this as a future research direction.
\end{sidenote}
\end{flushright}
\end{enumerate}

Now we use the derivative decay from P\ref{proposition:decay} to show fault tolerance using a Taylor expansion. We write $q=1/n_l$ and $r=p+q$. In the following we will use Assumption \ref{assumption:continuous_net} only by its consequence -- Proposition \ref{proposition:decay}. We note that the conclusion of it can hold in other cases as well. We just give sufficient conditions for which it holds.

\begin{theorem}
	\label{prop:taylor}
	For crashes at layer $l$ and output of layer $L$ under assumption \ref{assumption:continuous_net} the mean and variance of the error can be approximated as
	$$
	\E\Delta_L  =p_l\sum\limits_{i=1}^{n_l}\frac{\partial y_L}{\partial \xi^i_l}+\Theta_{\pm}(1)D_2 r^2,\,
	\Var\Delta_L=p_l\sum\limits_{i=1}^{n_l}\left(\frac{\partial y_L}{\partial \xi^i_l}\right)^2+\Theta_{\pm}(1)D_{12}^2r^3\\
	$$
	By $\Theta_{\pm}(1)$ we denote any function taking values in $[-1,1]$. The derivative $\partial y_L/\partial \xi^i_l(\xi)\equiv-\partial y_L(y_l-\xi\odot y_l)/\partial y_l^i\cdot y_l^i$ is interpreted as if $\xi^i_l$ was a real variable.
	%\vspace{-10pt}
\end{theorem}
\begin{proof}
We consider crashes at layer $l$ as crashes in the input $x$ to the rest of the layers of the network. Thus, without loss of generality, we set $l=0$.

Consider $\Delta(\xi)=y(\underbrace{(1-\xi)\odot x}_{\hat{x}(\xi)})-y(x)$. Then we explicitly compute
$$
\frac{\partial \Delta(\xi)}{\partial \xi^i}=-\frac{\partial y(\hat{x}(\xi))}{\partial x_i}x_i,\,
\frac{\partial^2 \Delta(\xi)}{\partial \xi^i\partial\xi^j}=\frac{\partial^2 y(\hat{x}(\xi))}{\partial x_i\partial x_j}x_ix_j
$$

Now, consider $\Delta(\xi)=\Delta(0)+(\Delta'(0),\xi)+\frac{1}{2}\xi^T\Delta''(t(\xi))\xi$ by the Taylor theorem with a Lagrange remainder.

We assume $\|x\|_{\infty}\leq 1$ (otherwise we rescale $W_1$). We group the terms into distinct cases $i=j$ and $i\neq j$:
$$\E\Delta=-p\Delta'(0)+\underbrace{\sum\limits_{i=j}\E (\xi^i)^2\Delta''(t(\xi))}_{E_1}+\underbrace{\sum\limits_{i\neq j}\E\xi^i\Delta''(t(\xi))\xi^j}_{E_2}$$

The second term is $|E_1|\leq p\cdot 2D_2/n_0^2\cdot n_0=\mathcal O(pD_2/n_0)=\mathcal(pqD_2)$

The third term is $|E_3|\leq n_0^2D_2p^2/n_0^2=\mathcal O(p^2D_2)$.

Therefore, we have an expansion $\boxed{\E\Delta=-\nabla_\xi \Delta(0)\cdot x+\mathcal(q+p)\cdot pD_2=-p\sum\limits_{i}x_i\frac{\partial y(x)}{\partial x_i}+\mathcal O(D_1p+D_2r^2)}$.

The expectation just decays with $p$, but not with $n_0$.

Now, consider the variance $\Var\Delta=\Var(\underbrace{\Delta'(0)\xi}_{V_1}+\underbrace{0.5\xi^T\Delta''(t(\xi))\xi}_{V_2})=\E\Delta^2-(\E\Delta)^2=\E(V_1+V_2)^2-(\E(V_1+V_2))^2=\Var V_1+\Var V_2+2Cov(V1, V2)$

Consider $\Var V_1=p\sum (\frac{\partial y(x)}{\partial x_i})^2x_i^2=\mathcal O(pqD_1)$. This is the leading term, the rest are smaller.

And the second term $\Var V_2\leq \E V_2^2=\sum\limits_{ijkl}D_2^2/n_0^4\E \xi^i\xi^j\xi_k\xi_l$. Here we consider various cases for indices $i,j,k,l$, based on the partition of $4=0+4=2+2=1+3$. If all indices are different, we get $\mathcal O(p^4)$ from $\E\xi$. If all are the same, we get $\mathcal O(p/n^3)$. If we have $2$ groups of $2$, we get $\mathcal O(p^2/n^2)$, if $3+1$ we get $\mathcal O(p^2/n^2)$. Thus, $\Var V_2=\mathcal O(D_2^2r^4)$

Consider the final term $Cov(V_1, V_2)\leq \E |V_1V_2|+\E |V_1|\E |V_2|$. $\E |V_1|\leq \mathcal O(pD_1)$, $\E |V_2|\leq D_2 p/n$. $\E V_1V_2=\sum\limits_{ijk}D_1D_2/n_0^3\E \xi^i\xi^j\xi_k$. All different indices give $p^3$, all same indices give $p/n_0^2$, and $2+1$ give $p^2/n$. Thus, $Cov(V_1, V_2)=\mathcal O(D_1D_2r^3)$

Finally, $\boxed{\Var\Delta=p\nabla^2\Delta(0)x+\mathcal O(r^3D_1D_2)+\mathcal O(r^4 D_2^2)=\sum \left(\frac{\partial y(x)}{\partial x_i}\right)^2x_i^2+\mathcal O(r^3D^2)=\mathcal O(D_1^2p/n+D^2r^3)}$
\end{proof}

\paragraph{A better remainder.} It is possible to obtain a remainder of $\mathcal O(p^2/n)+\mathcal O(p/n^2)$ for the variance instead of a generic $\mathcal O(r^3)$. This makes the remainder decay with $n$ as well. A more fine-grained analysis of the remainder could consider the expression for the variance and explicitly compute a correlation between $\xi^i \Delta'_i(t(\xi))$ and $\xi^j \Delta'_j(t(\xi))$. However, we were unsuccessful in doing so. Another approach is to take one more term in the expansion: $\mathcal O(r^5)$ -- it will make the previous term with $p^3$ go away, leaving only terms $p^2/n$ and $p/n^2$, as the difference expressions in the variance will cancel out. Another idea is to use a Taylor series $\Var\Delta=\sum\limits_{k=0}^\infty \frac{T_k}{k!}r^k$ (existing in principle). Next, we explicitly bound the remainder since we have a global bound on $D_k\leq D$. Finally, if we expand around the mean $\mu=\E x$: $\E f(x)=f(\mu)+\frac{1}{2}\Theta_{\pm}(1)(f''_{ii},\Var x)$\cite{taylor_cross_validated}, we get a remainder $p/n$ right away, but need to compute $f(\mu)$ which is the network at a modified input $(x-px)$.

\section{Probabilistic Guarantees on The Fault Tolerance Using Tail Bounds}
\label{sec:guarantee}
Under our assumptions, $\Var\Delta\sim \sum_l\frac{C_lp_l}{n_l}$. The constant $C_i$ comes from Theorem \ref{prop:taylor} (or its more computationally tractable form in Add. Coroll. \ref{th:error_eq_w}). Since we know that the error superposition is linear (Additional Proposition \ref{th:lin}), we sum the individual layer terms, hence $\sum_l$. Proposition \ref{proposition:decay} motivates the inverse dependency on $n_l$.

\paragraph{Median trick.} Suppose that we have a random variable $X$ satisfying $\Prob\{X\geq \varepsilon\}<1/3$. Then we create $R$ independent copies of $X$ and calculate $X=median\{X_i\}_{i=1}^R$. Then $\Prob\{X\geq\varepsilon\}<(1/3)^{R/2}$ because in order for the median to be larger than value $\varepsilon$, at least half of its arguments must be larger than $\varepsilon$, and all $X_i$ are independent. Thus $R=\mathcal O(\log 1/\delta)$ in order to guarantee $(1/3)^{R/2}<\delta$. This is a standard technique.

\begin{mainprop}
	\label{th:guarantee}
	A neural network under assumptions \ref{assumption:psmall}-\ref{assumption:continuous_net} is $(\varepsilon,\delta)$-fault tolerant for $t=\varepsilon-\E\Delta_L>0$ with $\delta=\delta_0+t^{-2}\Var\Delta_L$ for $\E\Delta$ and $\Var\Delta$ calculated by Theorem \ref{prop:taylor} and $\delta_0$ from Proposition \ref{prop:usmall}.
\end{mainprop}
\begin{proof}
%First we upper-bound the probability that weight perturbation is big using Prop. \ref{prop:usmall}, estimating $\delta_0$

We apply the Taylor expansion from Theorem \ref{prop:taylor}. We directly apply the Chebyshev's inequality for $X=\Delta$:
$$
\delta=\Prob\{X\geq t+\mathbb EX\}\leq\frac{\Var X}{t^2}
$$
\end{proof}

%Then we combine these two results using the Union Bound and conclude by $\delta=\delta_0+\delta_1$.
\subsection{Properties of The Fault Tolerance}

\begin{mainprop}
	\label{th:gd_zero}
	Suppose that a $C^2$ network is at a stationary point after training: $\E_x\nabla_W L(x)=0$. Then in the first order on $p,q$,
	$\E_x\E_\xi\Delta_{L+1}=0$
\end{mainprop}
\begin{proof}
Consider the quantity in question, use Additional Proposition~\ref{th:error_mean_eq} for it and apply $\E_x$ on both sides:
$$
\E_\xi\Delta_{L+1}=\E_x \left(\frac{\partial L}{\partial y}\E_\xi\Delta\right)=-p\left(\E_x\frac{\partial L}{\partial W},W\right)
$$
Now since we know that $\E_x\frac{\partial L}{\partial W}=0$, the linear term is $0$
\end{proof}

\begin{proposition}{(Linearity of error superposition for small $p$ limit)}
\label{th:lin}
Consider a network $(L,W,B,\varphi)$ with crashes at each layer with probability $p_l$, $l\in\overline{0,L}$. Then in the first order on $p$, the total mean or variance of the error at the last layer is equal to a vector sum of errors in case crashes were at a single layer% for $p=\max\{p_0,\ldots,p_L\}$:
$$
\arraycolsep=0.6pt
\begin{array}{rlrllrll}
\E\Delta_L^{p_0,...,p_L}  &=&\E&\Delta_L^{p_0}  &+\ldots+&\E&\Delta_L^{p_L}  &\\
\Var\Delta_L^{p_0,...,p_L}&=&\Var&\Delta_L^{p_0}&+\ldots+&\Var&\Delta_L^{p_L}&
\end{array}
$$
\end{proposition}

\begin{proof}
Consider each layer $i$ having a probability of failure $p_i\in[0,1]$ for $i\in\overline{0,L}$.

In this proof we utilize Assumption \ref{assumption:psmall}.
We write the definition of the expectation with $f(p, n, k)=p^k(1-p)^{n-k}$ being the probability that a binary string of length $n$ has a particular configuration with $k$ ones, if its entries are i.i.d. Bernoulli $Be(p)$. Here $S_l$ is the set of all possible network crash configurations at layer $l$. Each configuration $s_l\in S_l$ describes which neurons are crashed and which are working. We have $|S_l|=2^{n_l}$.
$$
\E\Delta_L^{p_0,...,p_L}=\sum\limits_{s_0\in S_0}\ldots\sum\limits_{s_L\in S_L}f(p_0,|s_0|,N_0)\ldots f(p_L,|s_L|,N_L)(\hat{y}_L-y_L)
$$
where

$$
\begin{array}{rllllll}
\hat{y}_L&=(&W_L\varphi(\ldots(\varphi(W_1(&x\odot\xi_0)&+b_1)\odot \xi_1)&\ldots)\odot \xi_{L-1}&+b_L)\odot\xi_L\\
y_L&=&W_L\varphi(\ldots(\varphi(W_1&x&+b_1)&\ldots)&+b_L
\end{array}
$$
We utilize the fact that the quantity $f(p,n,k)=p^k(1-p)^{n-k}=p^k+\mathcal O(p^{k+1})$. Only those sets of crashing neurons $(s_0,....,s_L)$ matter in the first order, which have one crash in total (in all layers). We denote it as $s^1_k\in S_k$. Therefore we write

$$
\E\Delta_L^{p_0,...,p_L}=\sum\limits_{s^1_0\in S_0}f(p_0,1,N_0)(\hat{y}_L-y_L)+\ldots+\sum\limits_{s^1_L\in S_0}f(p_L,1,N_L)(\hat{y}_L-y_L)
$$

This is equivalent to a sum of individual layer crashes up to the first order on $p$:

$$
\E\Delta_L^{p_0,...,p_L}=\E\Delta_L^{p_0}+\ldots+\E\Delta_L^{p_L}
$$

The proof for the variance is analogous.

\end{proof}

\section{Algorithm for Certifying Fault Tolerance}
In case if we consider the RHS quantities in Theorem \ref{prop:taylor} averaged over all data examples $(x, y^*)$, then the Algorithm 1 from the main paper would give a guarantee {\em for every example}: it will guarantee that $\mathbb P_{x,\xi}[\Delta\geq \E\Delta+t]$ is small. Indeed, if we know that $\E_x \Var\Delta$ is small, we know that the total variance $\Var_{x,\xi}\Delta=\E_x\Var_{\xi}\Delta+\Var_{x}\E_{\xi}\Delta$ (by the law of total variance) is small as well. Indeed, the second term $\Var_{x}\E_{\xi}\Delta$ bounded by $\sup_x\E_{\xi}\Delta$ which we assume to be small. In case if it is not small, it is unsafe to use the network, as even the expectation of the error is too high. Given a small total variance $\Var_{x,\xi}\Delta$, we apply, as in Proposition \ref{th:guarantee} a Chebyshev's inequality to the random variable $\Delta$ over the joint probability distribution $x,D|x$. This will give $\mathbb P_{x,\xi}[\Delta\geq \E\Delta+t]\leq t^{-2}\Var_{x,\xi}\Delta$. This probability indicates how likely it is that a network with crashes and random inputs will encounter a too high error.

Median aggregation of $R$ copies works for the input distribution as well. Denote "$y_i(x)$ is bad" as the event that the loss exceeds $\varepsilon$ for the i'th copy. We denote $[True]=1$ and $[False]=0$ (Iverson brackets). Now, $\delta=P_{x,\xi}(Med(y_i(x)) \mbox{ bad})=\E_x\E_{\xi}[\mbox{at least half of } y_i \mbox{ are bad}]$. Since the inner probability can be bounded as $\leq t^{-2}\Var\Delta(x)\exp(-R)$, taking an expectation over $\E_x$ results in the quantity discussed before, $\E_x\Var\Delta$.

We note that it would not be possible to consider $y_{L+1}$ to be the total loss, as it makes quantities such as $\partial L/\partial W^{ij}(x, W)$ ill-defined, as they depend on $x$ as well.
%\todo[inline]{Is this true if we take $y$ with an average?}

\subsection{Analysis of the Algorithm 1}
The algorithm consists of the main loop which is executed until we can guarantee the desired $(\varepsilon,\delta)$-fault tolerance. It trains networks, and upon obtaining a good enough network with $\delta<1/3$, it repeats the network a logarithmic number of times. We note that the part on $q$ and $\delta_0$ is not strictly required to guarantee fault tolerance. Rather, satisfying these conditions is a natural necessary conditions to satisfy the more strict ones (on $R_3$, $\E\Delta$ and $\delta$). These conditions are necessary because, by AP\ref{prop:cont_to_qfactor}, continuous limit implies that the $q$ is reasonable.

\paragraph{Space requirements.} After each iteration of the main loop of the algorithm, the previous network can be deleted as it is no longer used. Therefore, each new iteration does not require any additional space. We only need to store the network itself and the computational graph for $\Var\Delta$, which depends on network's gradients. The total space complexity is then $\mathcal O\left(\sum_{l=1}^Ln_l\times n_{l-1}\right)$ which is just the space to store the network's weights.

\paragraph{Time and neurons requirements.} Each iteration trains the network and then performs computations which involve only 1 forward pass, so the bottleneck is still the training stage which we assume can be done in time $\mathcal O(T)$. Now let us calculate the number of iterations. New iteration is requested in case if one of the conditions is not satisfied. We assume the loss to be bounded in $[-1,1]$ by D\ref{def:nn}. We analyze each of the possible non-satisfied conditions:
\begin{enumerate}
	\item $q<10^{-2}$. In that case the target value of $\max |W_i|/\min |W_i|>100$ which is many times greater than the loss itself (by assumption $\omega \in[-1,1]$). In this case the algorithm increases the regularization parameter $\mu$ to make $q$ smaller. We assume that we double $\mu$ each time. So by setting $\mu^*=1$ the network will definitely achieve $q>10^{-2}$. Therefore the number of doublings of $\mu$ is at most $\log 1/\mu^0$. Since $\mu^0$ is represented as a float, $\log 1/\mu^0=\mathcal O(1)$
	\item $\delta_0>1/3$. Since $\delta_0=\exp(-n_lqd_{KL}(\alpha|p_l))$, it will decay exponentially with the decay of $n_l$. In order to make $\delta_0<1/3$, we need to set $n_lqd_{KL}(\alpha|p_l)>2$ which leads to $n\geq \frac{200}{d_{KL}(\alpha|p_l)}$. Since both $\alpha$ and $p$ are small, we take only the first-order term in $d_{KL}(\alpha|p)=\alpha\log\alpha/p+(1-\alpha)\log(1-\alpha)/(1-p)$. The last term here is $\sim p-\alpha\geq -\alpha$. In case if $\alpha\geq e^2p$\footnote{For $e\colon \log e=1$}, $\log \alpha/p\geq 2$ and $d_{KL}(\alpha|p)\gtrsim \alpha$. Now, $\alpha$ needs to be sufficiently small in order to guarantee the second-order term in the Taylor expansion $\alpha^2D_{12}\ll 1$\footnote{Technically, here we silently implied that $e^2p\ll 1/\sqrt{D_{12}}$ in order to make $\alpha\geq e^2p$, which means that we {\em cannot} implement a function with too high second derivatives in neuromorphic hardware with a constant $p$, not matter how many neurons we take. Intuitively, this happens because for such a function, even a failure of $e^2p$ fraction of neurons, which is a reasonable expectation, is too large to begin with. We assume that we are fitting a function which is not like that. If we encounter this, we will see that by having a too high $\E\Delta$ and the algorithm will output {\tt infeasible}}. Therefore, $\alpha^2\sim 1/D_{12}$ and $n\sim 200 D_{12}$. If the algorithm increases $n$ by a constant amount, it will need to take $\mathcal O(D_{12})$ operations
	\item $R_3>C$. In this case, we increase $\psi$. Since this directly influences $R_3$ via regularization, doubling $\psi$ every time results in a similar $\mathcal O(1)$ performance as in the analysis for $q$ and $\mu$. Since $R_3\approx C$ is the number of changes that the weights make, which is $>1$, and the loss is bounded by $1$.
 	\item $\delta=\varepsilon^{-2}\Var\Delta\approx \varepsilon^{-2}\frac{C_lp_l}{n_l}> 1/3$ (see Section \ref{sec:guarantee}) for $C_l$ dependent on the function approximated by the NN and the continuous limit. Therefore, $n\sim \mathcal O(C_lp_l/\varepsilon^2)$.
\end{enumerate}
The total number of iterations is therefore $\mathcal O(D_{12}+C_lp_l/\varepsilon^2)$ for $C_l=n_l\Var\Delta/p_l$ for some $n_l$ (this is now a property of the function being approximated and the continuous limit). We note that the constants $1/3$ and $10^{-2}$ are chosen for the simplicity of the proof. The asymptotic behavior of the algorithm does not depend on them, as long as they are constant.

\paragraph{Correctness: guarantee of robustness.} Now we analyze correctness. We note that the algorithm is not guaranteed to find a good trade-off between accuracy and fault-tolerance. Up to this point, there is no complete theory explaining the generalization behavior of neural networks or their capacity. Therefore, we cannot give a proof for a sufficient trade-off without discovering first the complete properties of NNs capacity. We only show that the algorithm can achieve fault tolerance.

First, the condition on $\E\Delta$ and $\Var\Delta$ implies that the first-order terms in the expansion from T\ref{prop:taylor} are small enough. Now we argue that the remainder is small as well.

The condition $R_3<C$ implies that discrete function is smooth enough to apply $\int |W'_t(t,t')dtdt'|\approx C_1< R_3<C$ as well as $\int |W'_{t'}(t,t')|dtdt'\approx C_2<R_3<C$. This means that the integral is small, which allows to bound the Riemann remainder $R_3/n_l$ from the proof of AP\ref{prop:cont_to_discr}. This implies that there exist a continuous network $NN_c$ such that the approximation error $A$ from AD\ref{def:cont_discr_nn} is small. Here, the right metric is $R_3/n_l\approx C/n_l\ll 1$. The number of changes $C$ must be less than a number of neurons at a layer $n_l$.

Now, the remainder depends on the operator derivative bound $D_{12}$. By the Assumption \ref{assumption:continuous_net} (part on $D_k$), they are small. Below we describe why this part holds.

\paragraph{Experimental evidence for $D_k$ being small.}
First, since we explicitly have a term $R_1$ for $\Var\Delta\geq \left(\partial y/\partial y^{i}_l\right)^2y^{i}_l$ (AC\ref{th:error_eq_w}) in the regularizer, the first-order derivatives $D_1$ would be small for each layer by design of our algorithm. Next, second-order derivatives $\partial^2 y_L/\partial W_l^{ij}\partial W_l^{i'j'}$ are found to be small in experimental studies of the Hessian \cite{ghorbani2019investigation}. Since derivatives w.r.t. $y_l$ can be expressed via the derivatives w.r.t. weights by AP\ref{th:nn_eq}, and the continuous limit holds, $D_2$ is small as well.

\begin{flushright}
\begin{sidenote}
\paragraph{\bf General considerations for $D_k$ being small.}
Note, we can always renormalize $|x(t)|\leq 1$ and $|y(t)|\leq 1$ by rescaling the input and output weights. For the input layer, $D_{12}$ can be bounded via the properties of the ground truth function $y^*$, if $y$ approximates $y^*$ well. However, it is known that it can be false: the network can have a sufficient accuracy on the training dataset, but have much larger output-input derivatives ($d[out]/d[in]$) than the ground truth function. Specifically, in the task of image recognition, we (humans) assume the problem to be quite smooth: a picture of a cat with a missing ear or whiskers is still a picture of a cat. However, it was shown \cite{ilyas2019adversarial} that modern CNNs use {\em non-robust} features. This implies that CNNs are much more sensitive to small changes in input $x$, in contrast to the smooth ground truth function $y^*$ that we want it to learn, making the bound for the derivative $D_1$ large.

For the hidden layers, it is possible that the continuous limit has large derivatives. For example, we can first apply an injective transformation $\mathcal F$ with high derivatives, and then apply the inverse transform implemented as another neural network $\mathcal F^{-1}$ existing by AP\ref{prop:completeness}\footnote{The idea to construct $y=\mathcal F\circ \mathcal F^{-1}$ with $\mathcal F^{-1}$ implementable by a neural network is taken from \cite{doerig2019unfolding}}. Then, we have an overall smooth (identity) operator $y_L$ which, however, consists of two very non-smooth parts.

We list the following approaches that potentially could resolve this issue theoretically rather than experimentally. First, we can consider the infinite depth limit \cite{sonoda2017double}. This would allow to have regularities throughout the network' layers. Another approach is to study mathematically the ways that an operator can be decomposed into a hierarchical composition of operators. For example, for images, a natural decomposition of the image classification operator could first detect edges, then simple shapes, then groups them into elements pertinent to specific classes and then detects the most probable class \cite{zeiler2014visualizing}. At each stage, only robust features are used. Thus, for such a decomposition, output-hidden derivatives would be reasonable as well as the output-input ones: indeed, the decision to recognize a cat would not change significantly if some internal hidden layer features (ears or whiskers) are not present. Interestingly, just enforcing the continuous limit seems to make features more robust in our experiments, see hidden layer weights on Figures \ref{fig:decay_mnist} and \ref{fig:decay_fashion_mnist}. Without regularization, the weights seem noisy, there are a lot of unused neurons, and even the used ones contain noisy patterns. In contrast, continuity-regularized networks seem to have first-layer weights similar to the input images. This could be an interesting research direction to continue.

We note, however, that such a study is not connected to the fault tolerance anymore, as it is a fundamental investigation into the properties of the hierarchical functions that we want to approximate, and into the properties of the neural networks which we can find by gradient descent.
\end{sidenote}
\end{flushright}

\section{Experimental Evaluation}
\label{sec:experiments}
Experiments were performed on a single machine with 12 cores, 60GB of RAM and 2 NVIDIA GTX 1080 cards running Ubuntu 16.04 LTS and Python/Tensorflow/Keras. We test the proposed bounds on two datasets as a proof of concept: the Boston Housing dataset ({\tt https://www.cs.toronto.edu/\~delve/data/boston/bostonDetail.html}) (regression) and the MNIST dataset (classification) ({\tt http://yann.lecun.com/exdb/mnist/}). In addition, we use the Fashion MNIST dataset (classification, {\tt https://www.kaggle.com/zalando-research/fashionmnist}). We use standard pre-trained networks from Keras (VGG, MobileNet, ..., {\tt https://keras.io/applications/}).% and one image of a cat found via Google images (Figure \ref{fig:thecat}).

For the Dropout experiment $p<0.03$ is used as a threshold after which no visible change is happening. We use the unscaled version of dropout\footnote{We note that it is the same as the scaled version up to the first order}.

In the experiments, we use computationally tractable evaluations of the result given by T\ref{prop:taylor}. The "b1 bound" is the Spectral bound from P\ref{th:spectral}. The "b2 bound" corresponds to AP\ref{th:bound_v2}. The "b3 bound" corresponds to first-order terms from T\ref{prop:taylor}, or the Additional Corollary \ref{th:error_eq_w}, and the "b4 bound" corresponds to an exact evaluation of single-neuron crashes (taking $\mathcal O(n_l)$ forward passes).

\subsection{Boston-trained networks with different initialization: rank loss with experimental error (additional)}
\begin{figure}
	\centering
	\includegraphics[width=0.2\textwidth]{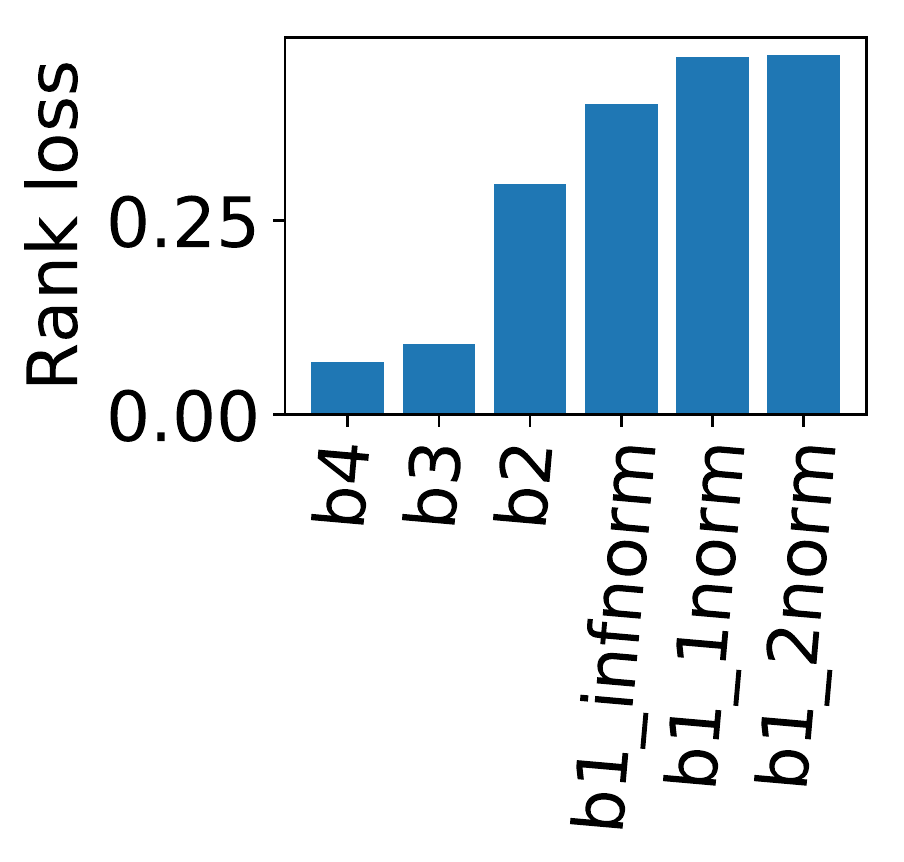}
	\includegraphics[width=0.2\textwidth]{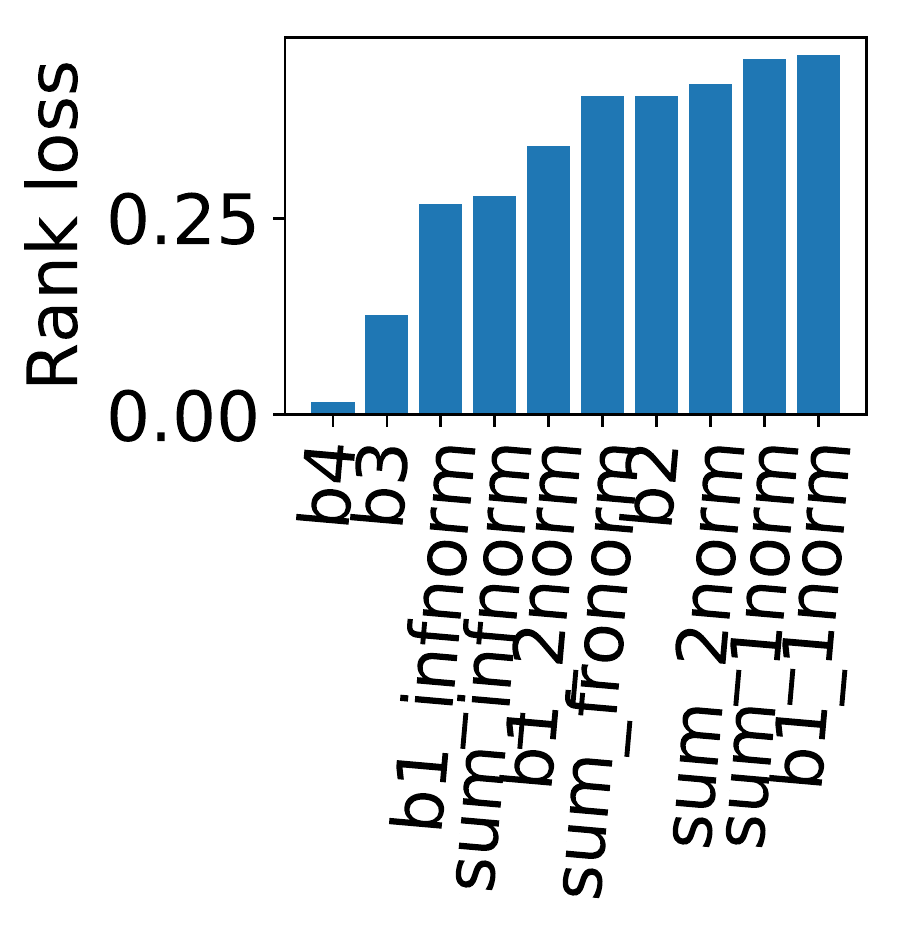}
	\includegraphics[width=0.2\textwidth]{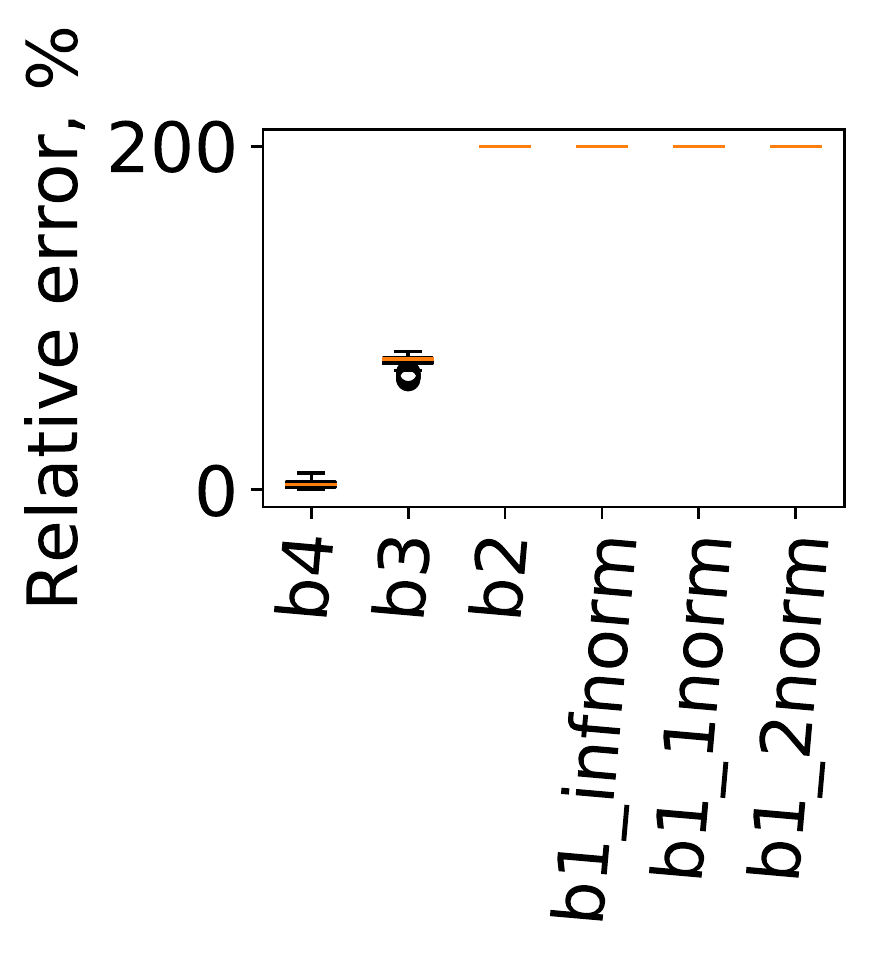}
	\includegraphics[width=0.2\textwidth]{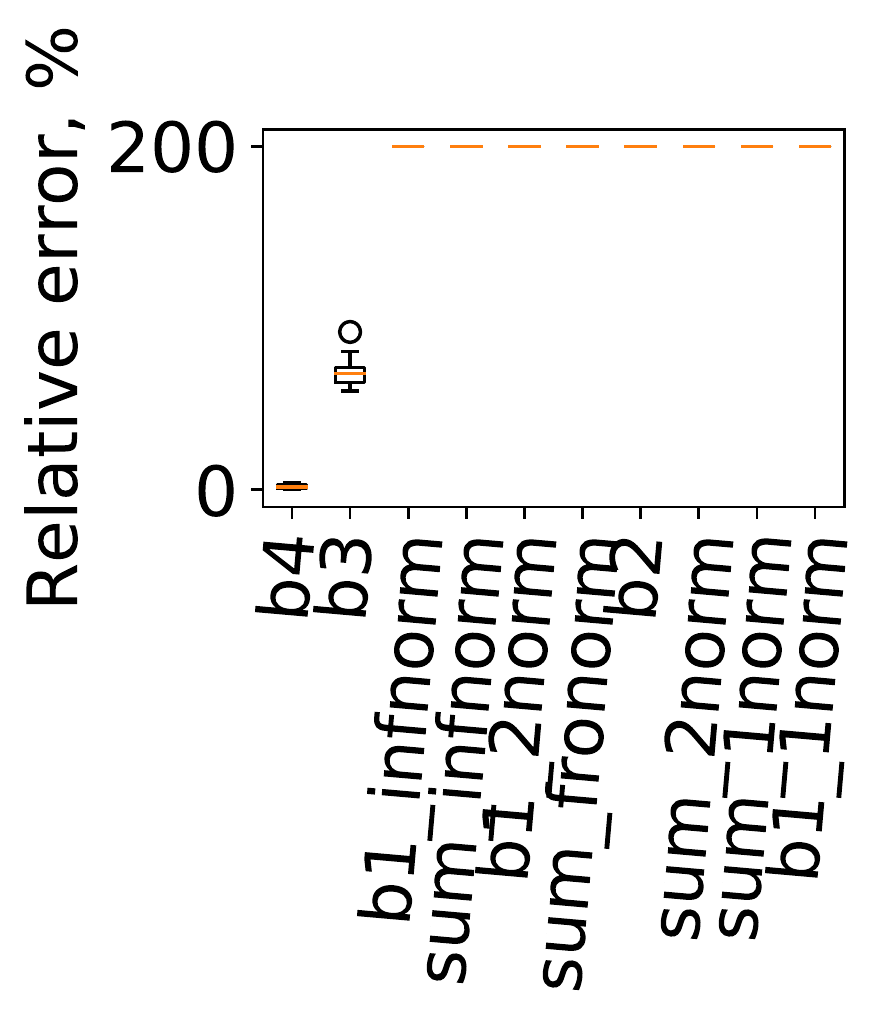}\\
	\includegraphics[width=0.2\textwidth]{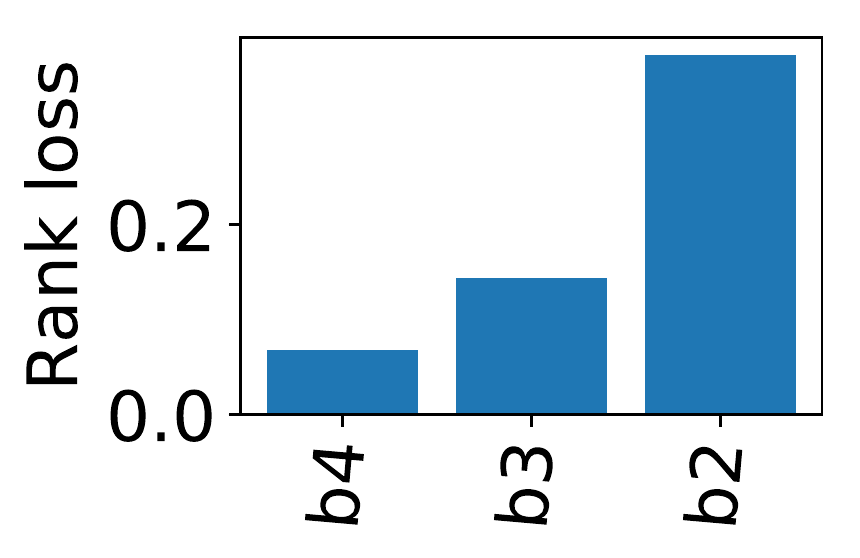}
	\includegraphics[width=0.2\textwidth]{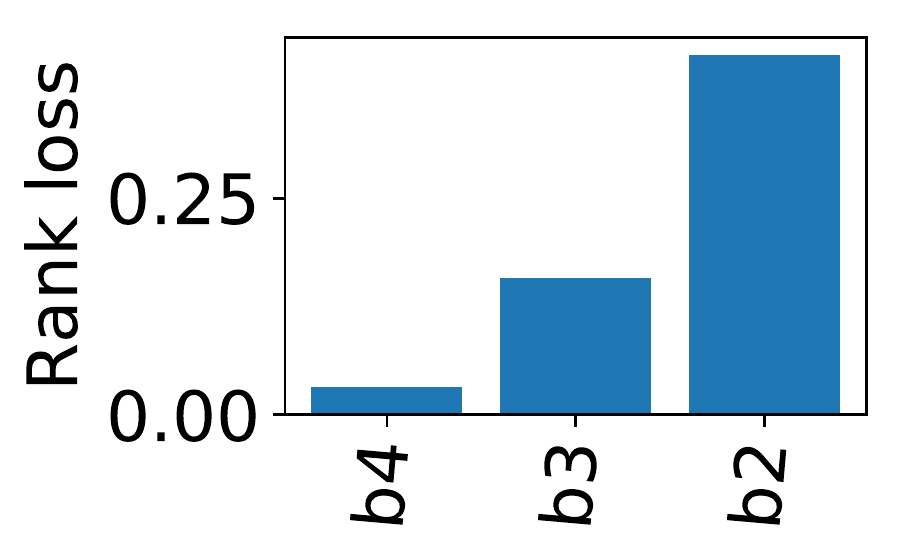}
	\includegraphics[width=0.2\textwidth]{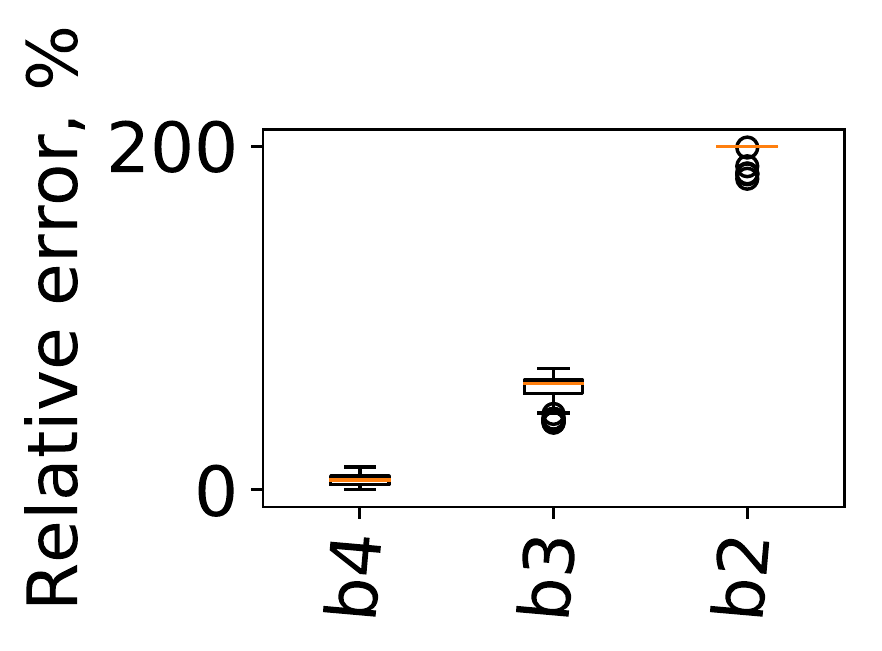}
	\includegraphics[width=0.2\textwidth]{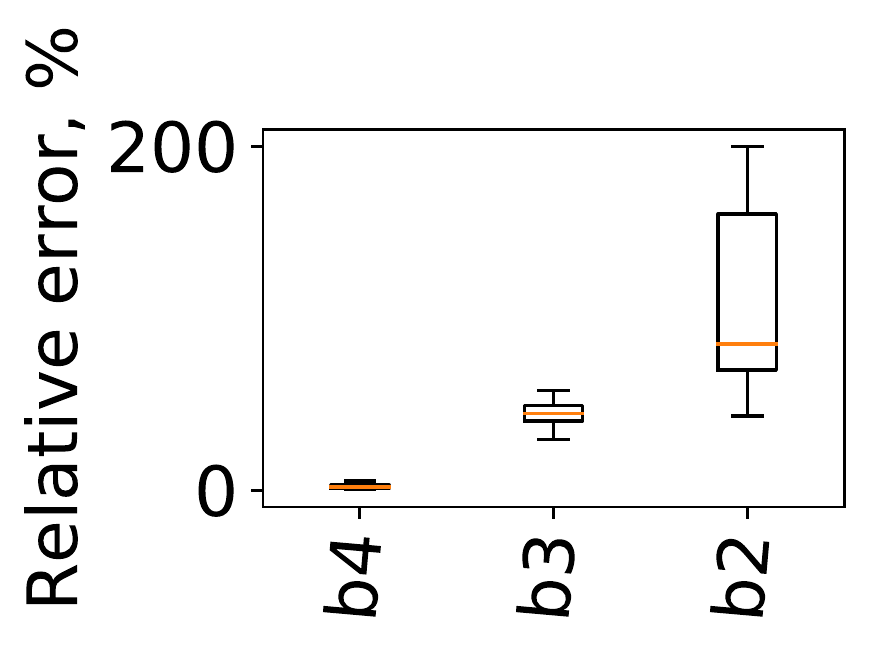}
	\caption{Boston-trained networks. Rank loss (worst value $0.5$) of ordering networks using bounds (left 4) and relative error in predicting the value of $\Delta$ (right 4). Mean of the error $\E\Delta$ (top) and standard deviation (bottom). Variable input on a fixed network (1st and 3rd columns) and variable network with fixed input (2nd and 4th columns).}
	\label{fig:comp_boston}
\end{figure}

We compare sigmoid networks with $N\sim50$ trained on Boston Housing dataset (see {\tt ErrorComparisonBoston.ipynb}). We use different inputs on a single network and single input on different networks. We compare the bounds using rank loss which is the average number of incorrectly ordered pairs. The motivation is that even if the bound does not predict the error exactly, it could be still useful if it is able to tell which network is more resilient to crashes. The second quantity is the relative error of $\Delta$ prediction, which is harder to obtain. Experimental error computed on a random subset $S'$ of all possible crashed configurations $S$ with $|S|=2^{20}\sim 10^6$ is used as ground truth, with $|S'|$ big enough to make the expectation and variance results not change from one launch to another.
The results are shown in Figure~\ref{fig:comp_boston}.
Even on this simple dataset we have only \boundthird{} and \boundfourth{} giving meaningful results both for rank loss and relative error of error. The results show that \boundfourth{} is always better than \boundthird{}, as expected. The same is done for random networks. %We note that in these experiments bounds \boundfourth{} and \boundthird{} work, therefore the small $p$ limit $np=10^{-3}\cdot 50<1$ holds as does the infinite width limit. 

\subsection{Comparison on random networks (additional)}
\begin{figure}[thb]
	\centering
	\includegraphics[width=0.2\textwidth]{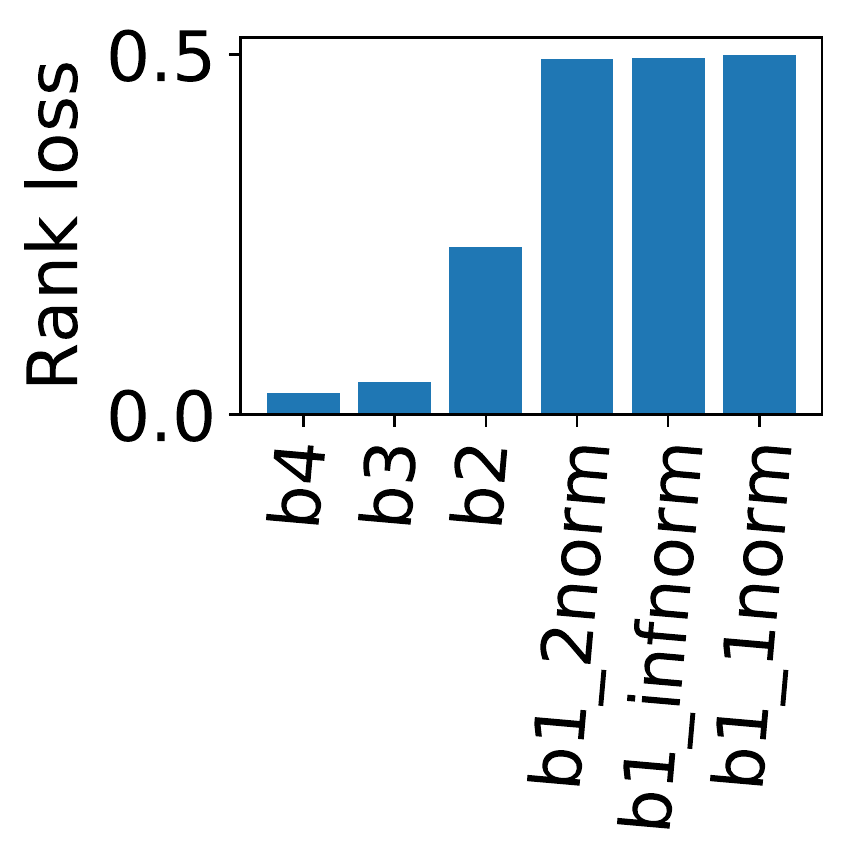}
	\includegraphics[width=0.2\textwidth]{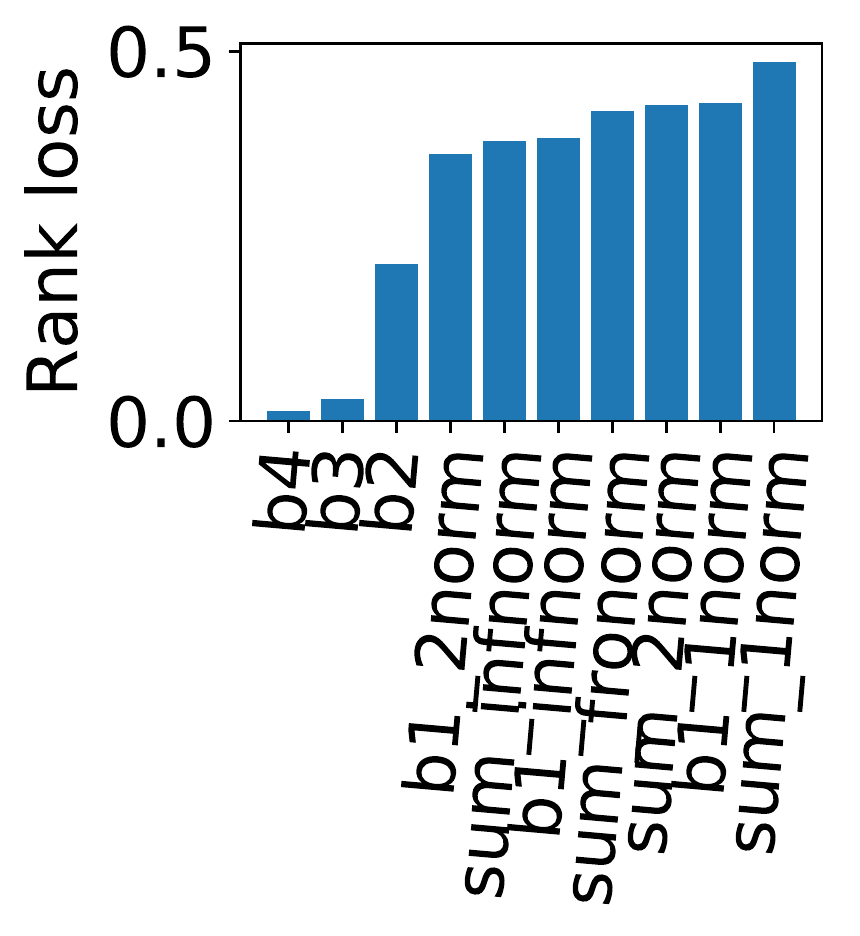}
	\includegraphics[width=0.2\textwidth]{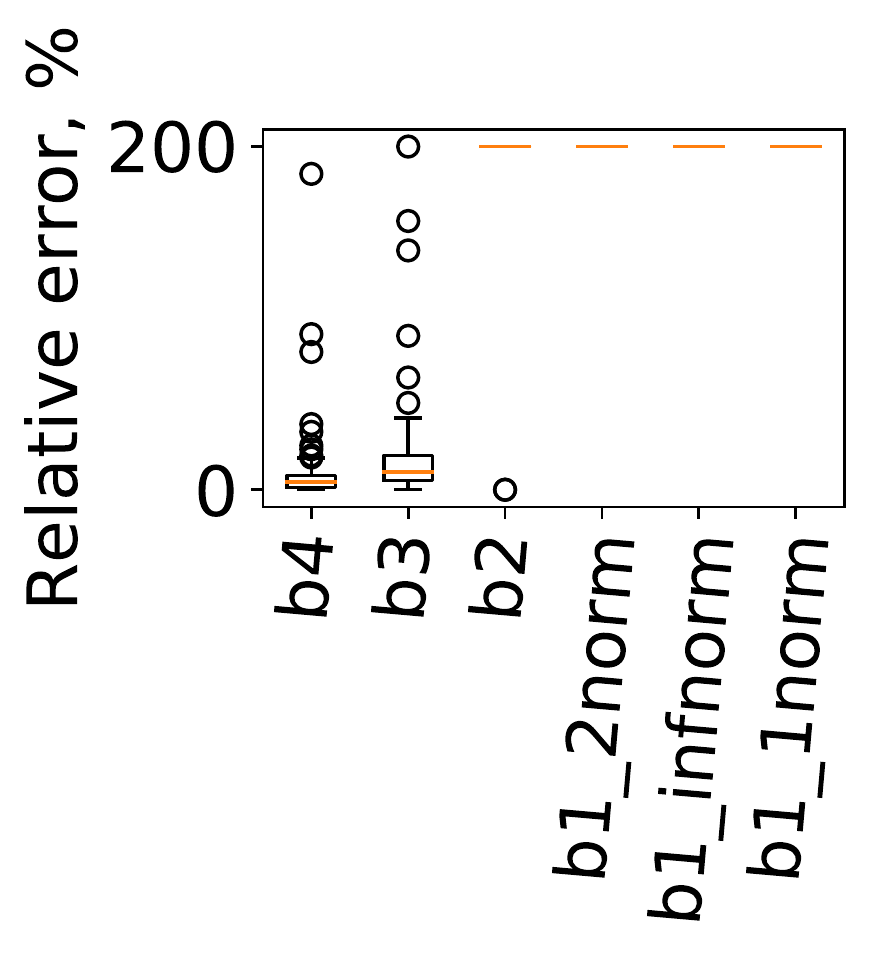}
	\includegraphics[width=0.2\textwidth]{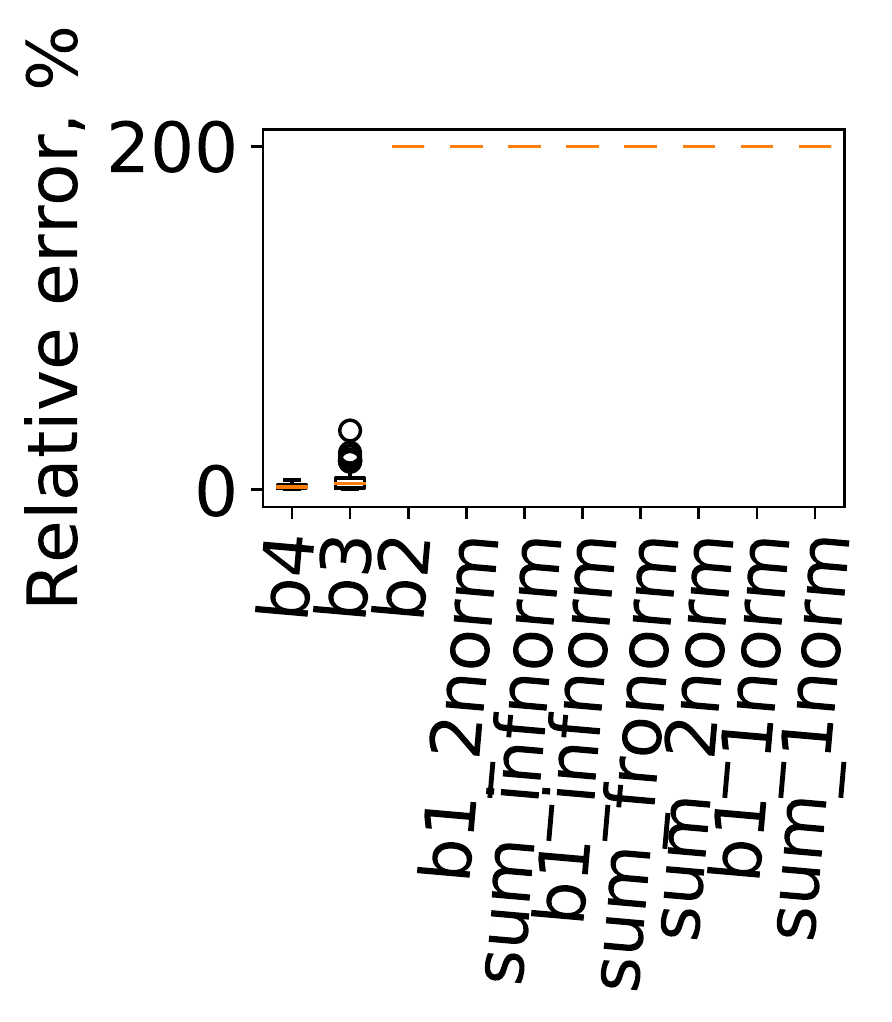}\\
	\includegraphics[width=0.2\textwidth]{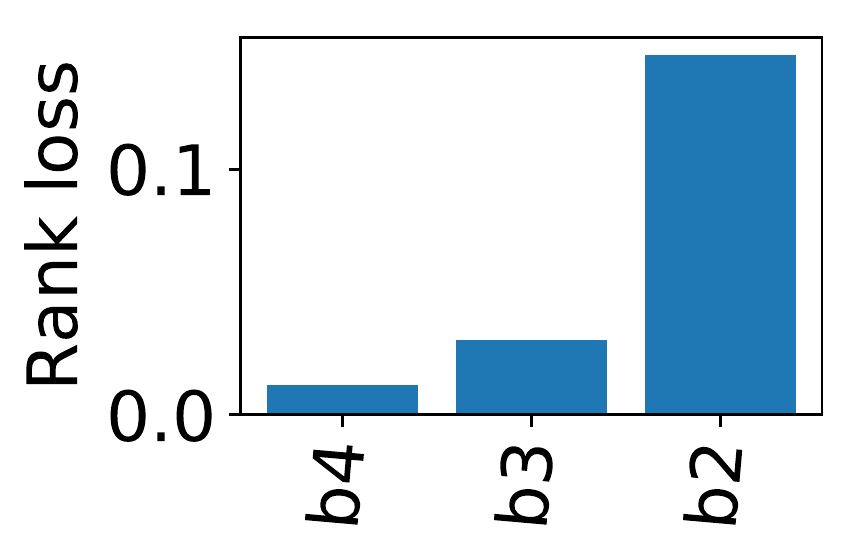}
	\includegraphics[width=0.2\textwidth]{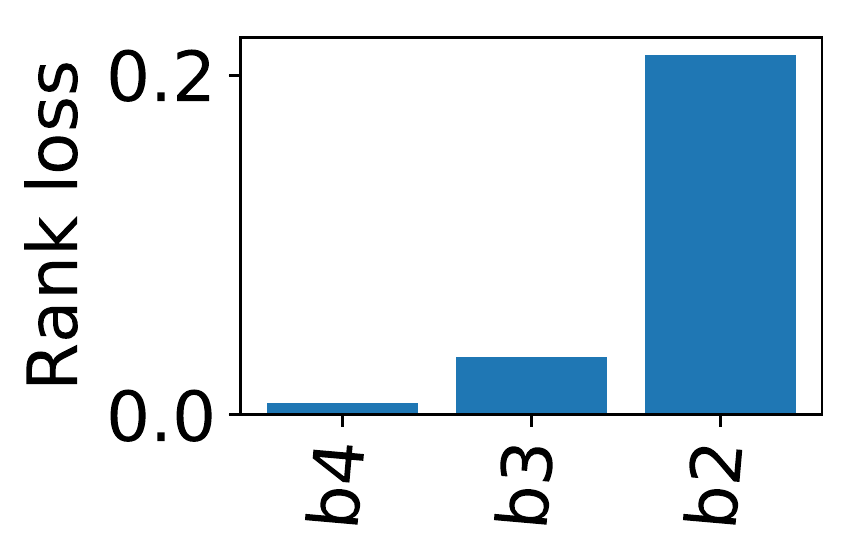}
	\includegraphics[width=0.2\textwidth]{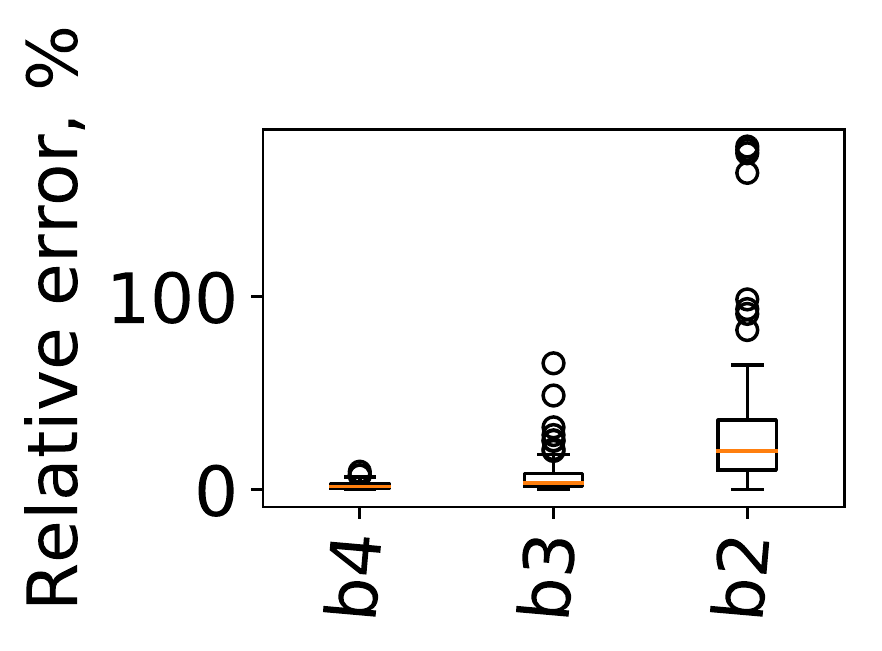}
	\includegraphics[width=0.2\textwidth]{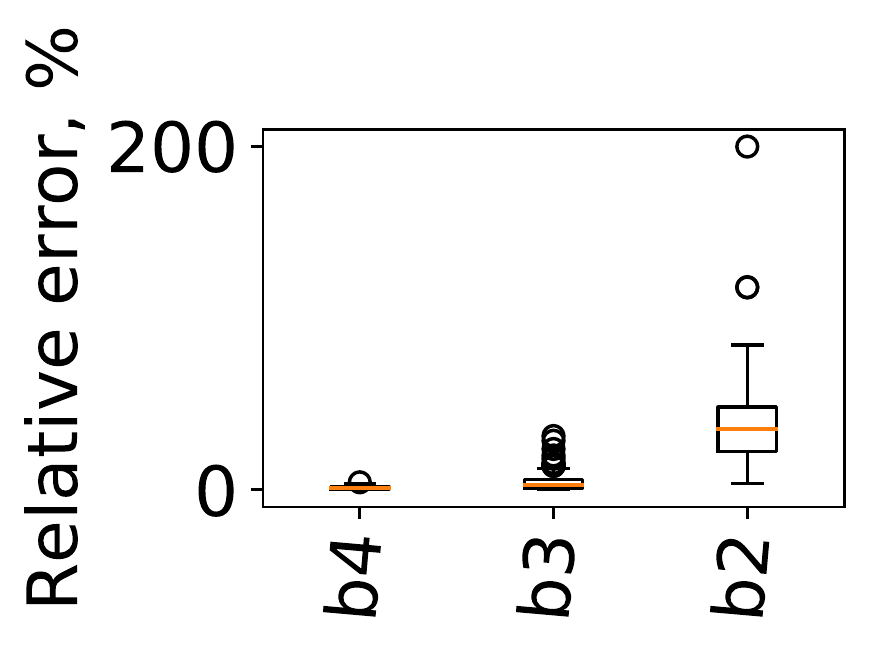}
	\caption{Random networks. Rank loss (worst value $0.5$) of ordering networks using bounds (left 4) and relative error in predicting the value of $\Delta$ (right 4). Mean of the error $\E\Delta$ (top) and standard deviation (bottom). Variable input on a fixed network (1st and 3rd columns) and variable network with fixed input (2nd and 4th columns).}
	\label{fig:comp_random}
\end{figure}

The same setup as for the Different Dropout Comparison experiment in the main paper is done for random networks with $N\sim 20$, ReLU activation function, see {\tt ErrorComparisonRandom.ipynb}. The results are shown in Figure~\ref{fig:comp_random} and they are qualitatively the same as for Boston dataset.

\subsection{Additional charts}
Table \ref{tab:comparison_big_cnns} presents the results of {\tt ConvNetTest-ft.ipynb}.

\begin{figure}[tbh]
	\centering
	\includegraphics[width=0.25\textwidth,valign=t]{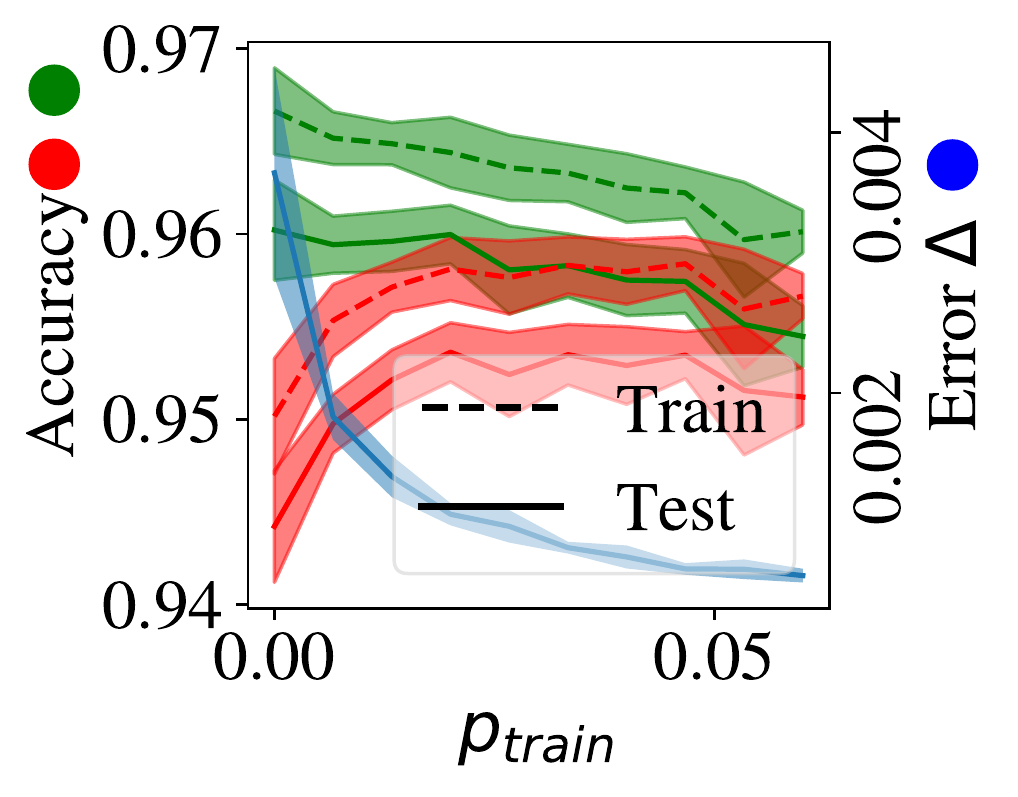}
	\caption{Comparison of networks trained with different dropout. Shown: accuracy and $\Var\Delta$ plots for different train dropout probability. Green curve shows accuracy of correct network, red shows accuracy for the crashing network. Dashed line show train dataset and solid represent test dataset. Variance of $\Delta$ estimated by \boundfourth{} shown in orange, by \boundthird{} in blue. Error bars are standard deviations in $10$ repetitions of the experiment.}
	\label{fig:comp_do}
\end{figure}

Figure \ref{fig:comp_do} presents the results from {\tt ComparisonIncreasingDropoutMNIST.ipynb}.

\begin{figure}[tbh]
	\centering
	\includegraphics[width=0.45\textwidth]{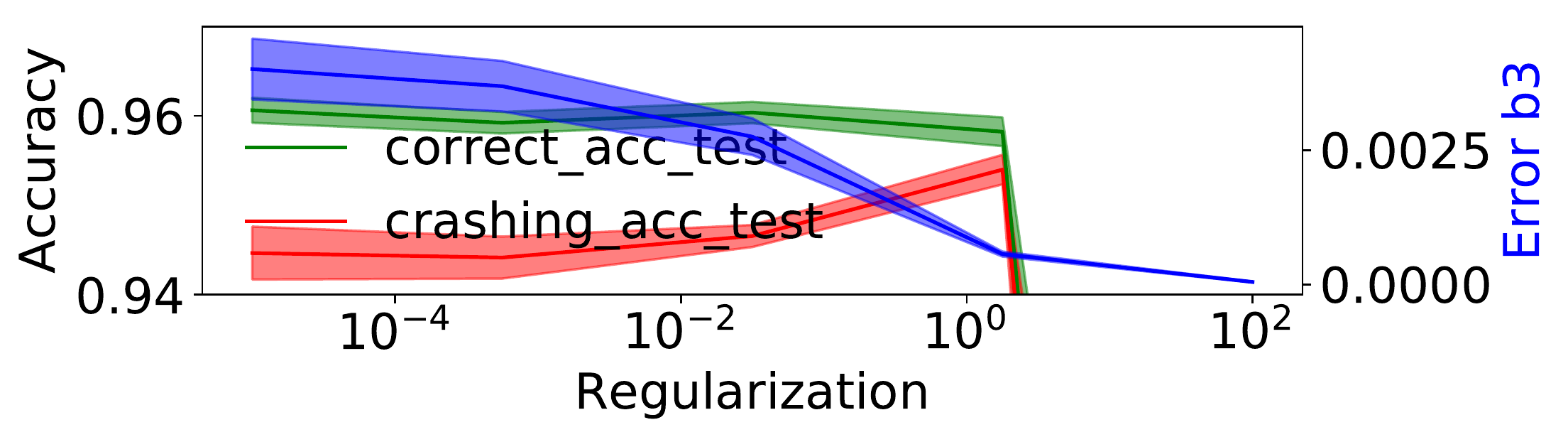}
	\caption{Horizontal axis represents increasing regularization parameter. Left vertical axis represents accuracy and has test crashing network accuracy (red) and correct network accuracy (green). Right axis shows variance of the error $\Var\Delta$ estimated by bound \boundthird{} used as a regularizer. Error bars are standard deviations in $5$ repetitions of the experiment.}
	\label{fig:comp_do_reg}
\end{figure}

Figure \ref{fig:comp_do_reg} shows results from {\tt Regularization.ipynb}

\begin{figure}
    \centering
    \includegraphics[width=0.5\textwidth]{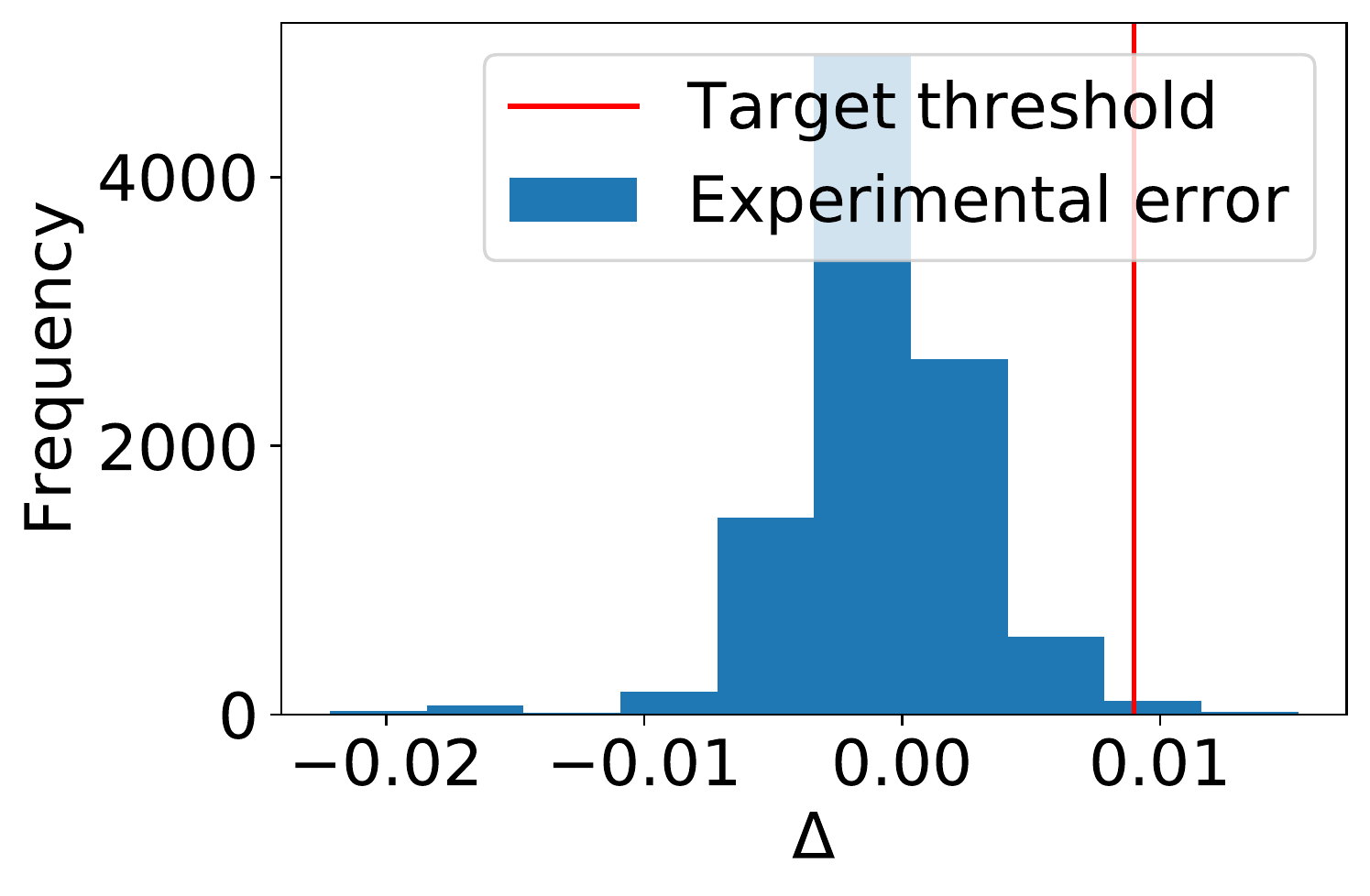}
    \caption{The distribution $\Delta$ of the error in the output for the network obtained using Algorithm 1 from the main paper}
    \label{fig:thealgo_result}
\end{figure}

\begin{figure}
    \centering
    \includegraphics[width=0.9\textwidth]{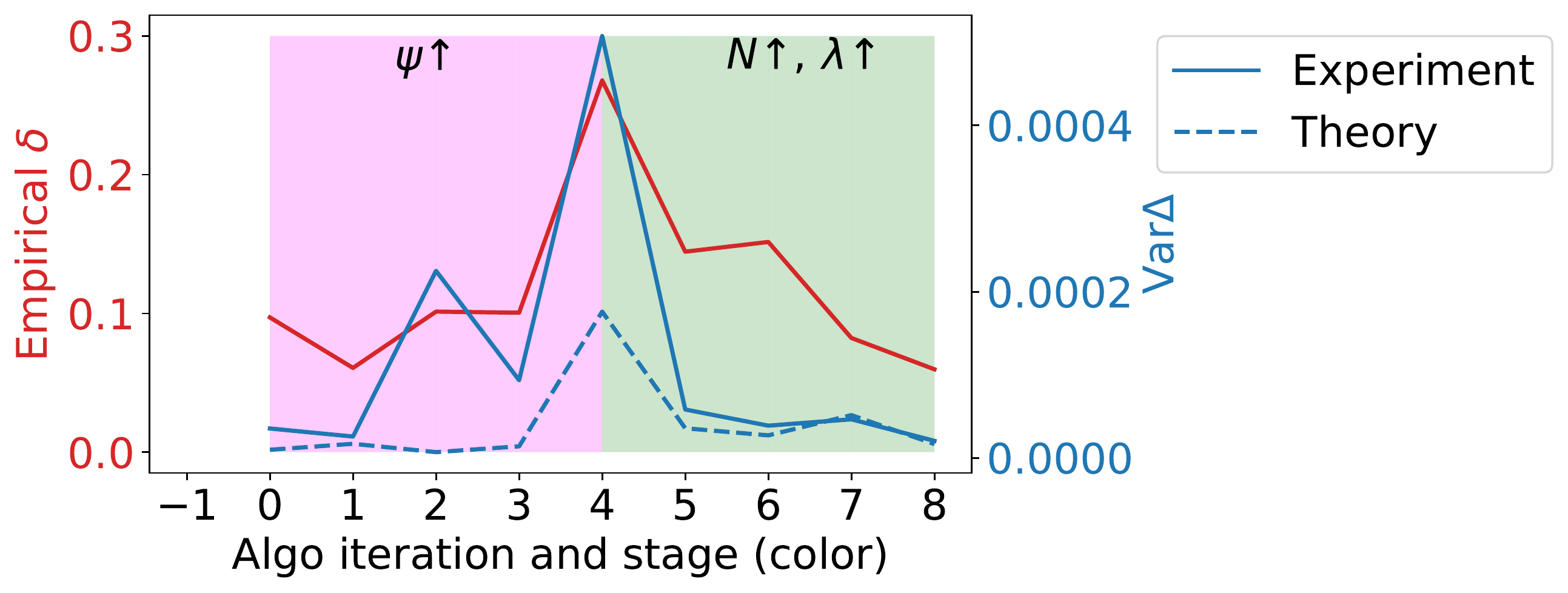}
    \caption{Evolution of the experimental error probability and the variance of the error over algorithm's iterations. There are two stages. In the first stage (red, iterations $\leq 4$) the algorithm increases continuity of the network via increasing $\psi$. In the second stage (green, iterations $\geq 5)$ the algorithm increases the number of neurons $n_1$ and the regularization parameter $\lambda$. First, it can be seen that at first stages the network is not continuous enough, as the algorithm makes it more continuous. This leads to an increase in the empirical probability $\delta$ of the network outputting a loss $>\varepsilon$ and in the increase in the gap between Theoretical $\Var\Delta$ and Experimental $\Var\Delta$. This happens because there are not enough neurons in the network. Later, as the number of neurons increases, the gap becomes smaller and the empirical probability decreases. Note that the first network (at iteration $0$) empirically satisfies our fault tolerance guarantee. Nevertheless, we do not have a proof for such a network because it is not continuous enough. Therefore, in order to guarantee robustness, we need to proceed with the iterations.}
    \label{fig:thealgo}
\end{figure}

Figure \ref{fig:thealgo} shows how the algorithm's (Main paper, Algorithm 1, {\tt TheAlgorithm.ipynb}) state evolves over iterations. Figure \ref{fig:thealgo_result} shows the resulting distribution of the error $\Delta$ at the output layer for the final network given by the algorithm.

\begin{table}[tbh]
    \centering
    \begin{tabular}{lrrrr}
\toprule
     Model &  $\log\Var\Delta$ &  $\log[$Parameters$/$Layers$]$ &  $\log[$Parameters$]$ &  Layers \\
\midrule
     VGG16 &        -18.7 &                    15.6 &           18.7 &      23 \\
     VGG19 &        -18.4 &                    15.5 &           18.8 &      26 \\
 MobileNet &         -9.1 &                    10.7 &           15.3 &      93 \\
\bottomrule
\end{tabular}
    \caption{Comparison of bigger convolutional networks when there are faults at every layer with $p=10^{-2}$}
    \label{tab:comparison_big_cnns}
\end{table}

\subsection{Error superposition (testing AP\ref{th:lin}, additional)}
\begin{figure}
	\centering
	\includegraphics[width=0.45\textwidth]{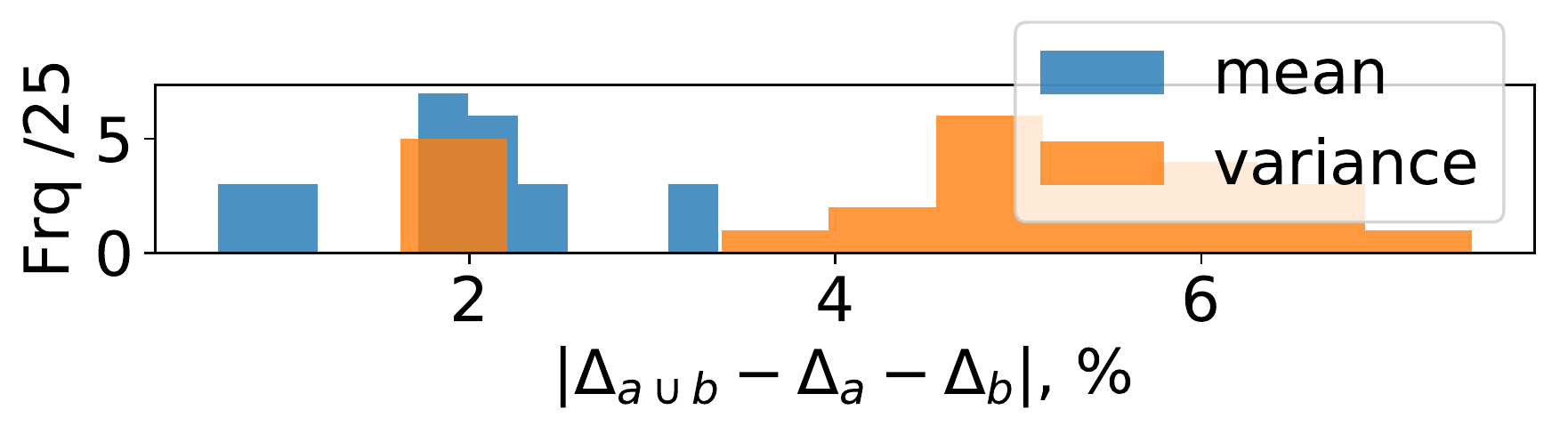}
	\caption{Distribution of relative error in the predicted $\Delta$ in percent for $25$ subset pairs. Distribution for the mean is shown in blue, for the variance in orange.}
	\label{fig:linearity}
\end{figure}

We test a random network with $L=4$ on random input (see {\tt ErrorAdditivityRandom.ipynb}). The error is computed on subsets of failing layers $a,b$. Then all pairs of disjoint subsets are considered, and the relative error of $\Delta$ estimation using linearity is computed as $\|\Delta_{a\cup b}-\Delta_a-\Delta_b\|/\|\Delta_{a\cup b}\|$.
The results (see Figure~\ref{fig:linearity}) show that this relative error is only few percent both for mean and variance, with better results for the mean.

\subsection{Testing AP\ref{th:error_mean_eq} (additional)}
\begin{figure}
	\centering
	\includegraphics[width=0.45\textwidth]{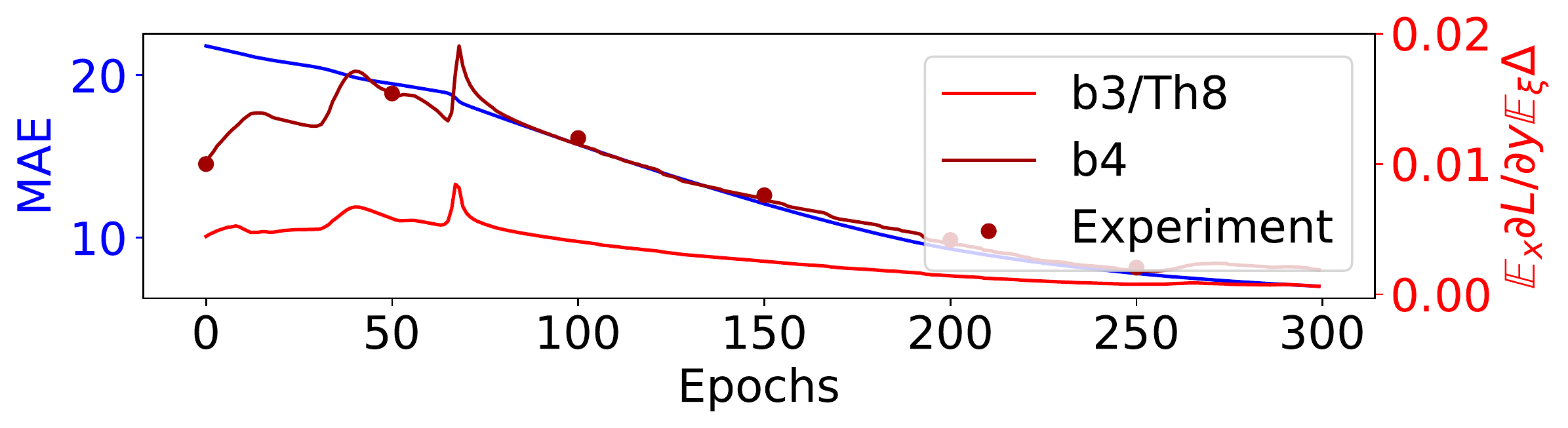}
	\caption{Decay with training of the loss (blue line, left vertical axis) and the error $\E_x\frac{\partial L}{\partial y}\E_\xi\Delta$ (red, right axis) together with bounds \boundthird{} and \boundfourth{} predictions (red and brown curves) and experimental values (dots). "Th8" means AC\ref{th:error_eq_w}}
	\label{fig:train_error}
\end{figure}

The result predicts the expected error $\E_x\E_\xi\frac{\partial L}{\partial y} \Delta$ to decay with the decay of the gradient of the loss. We tested that experimentally on the Boston dataset using sigmoid networks (see {\tt ErrorOnTraining.ipynb}) and note that a similar result holds for ReLU\footnote{ReLU$\notin C^1$. However only a small percentage of neurons are near $0$ thus approximation works for most of them.}. The results are shown in Figure~\ref{fig:train_error}. The chart shows first that the experiment corresponds to \boundfourth{}, and \boundthird{} is close to \boundfourth{}. \boundthird{} is also equal to the result of AP\ref{th:error_mean_eq}, both of which decay, as predicted.

\subsection{Practical bounds}

\begin{proposition}\label{th:nn_eq}
For neural network we have for any weight matrix $W_l$ and input to the $l$'th layer $y_{l-1}$:
$$
	\sum\limits_{i,j}\frac{\partial y}{\partial W^{ij}_l}W^{ij}_l=\sum\limits_{j}\frac{\partial y}{\partial y^j_{l-1}}y^j_{l-1}
$$
    
    The equality also holds for one particular $y_{l-1}^j$:
    $$
    \sum\limits_i \frac{\partial y}{\partial W_l^{ij}}W_l^{ij}=\frac{\partial y}{\partial y_{l-1}^j}y_{l-1}^j
    $$
\end{proposition}

\begin{proof}
Fix some layer $l$. The output of the network $y$ depends on the weight matrix $W_l$ and on the input to the $l$'th layer $y_{l-1}$. However we note that it only depends on their product and not on these quantities separately.

Therefore we write $y=y_L(W_l,y_{l-1},...)=\eta(W_ly_{l-1})$ and denote $x=y_{l-1}$, $W=W_l$ and $z=Wx$:
$$
y=\eta(Wx)
$$

Now we take one
$$\frac{\partial y}{\partial x_j}=\sum\limits_i\frac{\partial \eta}{\partial z_i}\frac{\partial z_i}{\partial x_j}$$

Since $z_i=\sum\limits_k W_{ik}x_k$, $\frac{\partial z_i}{\partial x_j}=W_{ij}$. Therefore we plug that in:
$$
\frac{\partial y}{\partial x_j}=\sum\limits_i\frac{\partial \eta}{\partial z_i}W_{ij}
$$

And multiply with $x_j$:
$$
\frac{\partial y}{\partial x_j}x_j=\sum\limits_i\frac{\partial \eta}{\partial z_i}W_{ij}x_j
$$

Now we compute $\frac{\partial y}{\partial W_{ij}}$ using that in vector $z$ only $z_i$ depends on $W_{ij}$ and also that $z=Wx$:
$$\frac{\partial y}{\partial W_{ij}}=\frac{\partial \eta}{\partial z_i}\frac{\partial z_i}{\partial W_{ij}}=\frac{\partial \eta}{\partial z_i}x_j$$

Then we multiply it by $W_{ij}$ and sum over $i$:
$$
\sum\limits_i\frac{\partial y}{\partial W_{ij}} W_{ij}=\sum\limits_i\frac{\partial \eta}{\partial z_i}x_j W_{ij}
$$

And now we note that the expressions for $\frac{\partial y}{\partial x_j}x_j$ and $\sum\limits_i\frac{\partial y}{\partial W_{ij}} W_{ij}$ are exactly the same.

Differentiating the above expression gives a connection between second derivatives w.r.t weights and activations.
\end{proof}

\begin{corollary}
\label{th:error_eq_w}
The error expressed in weights in the first order is as in T\ref{prop:taylor}:
$$
\E_\xi\Delta_L^l=-p\sum\limits_{i,j}\frac{\partial y}{\partial W^{ij}_l}W^{ij}_l
$$
$$\Var\Delta_L^l=p\sum\limits_i\sum\limits_{j,k}\frac{\partial y}{\partial W_l^{ji}}\frac{\partial y}{\partial W_l^{ki}}W_l^{ji}W_l^{ki}$$

Where $\Delta^l_L$ is the error in case of neurons failing at layer $l$. Layer $0$ means failing input.
	
In other words, both mean and variance are defined by a tensor $X_l=W_l\odot \frac{\partial y}{\partial W_l}$, $z^j_l=-\sum\limits_{i}X_l^{ij}$: 
$$
E\Delta_L^l=p\sum\limits_{i} z_l^{i},\,\Var\Delta_L^l=p\sum\limits_i (z_l^i)^2
$$
\end{corollary}

\begin{proof}
According to Theorem \ref{prop:taylor} the expression for mean and variance are:
\begin{align*}
    \E\Delta&\approx-p(\nabla y(x),x)=-p\sum\limits_{i}\frac{\partial y}{\partial x_i}x_i\\
    \Var\Delta&\approx ((\nabla y(x))^2,x^2)=p\sum\limits_{i}\left(\frac{\partial y}{\partial x_i}x_i\right)^2
\end{align*}

Then by Additional Proposition~\ref{th:nn_eq} the expression used can be rewritten as (note the index swap $i\leftrightarrow j$):
$$\frac{\partial y}{\partial x_i}x_i=\sum\limits_{j}\frac{\partial y}{\partial W_{ji}}W_{ji}$$

Then for the mean it is:
$$
\E\Delta=-p\sum\limits_i\sum\limits_j\frac{\partial y}{\partial W_{ji}}W_{ji}
$$
and the indices can be swapped again for the sake of notation. For the variance we write the square as the inner sum repeated twice with different indices $k$ and $j$:
$$
\Var\Delta=p\sum\limits_i\sum\limits_j\sum\limits_k\frac{\partial y}{\partial W_{ji}}W_{ji}\frac{\partial y}{\partial W_{ki}}W_{ki}
$$

This gives the first statements. Then we notice that both expressions depend on the values of tensor $\frac{\partial y}{\partial W}\odot W$ but not on $W$ or $\frac{\partial y}{\partial W}$ alone. We therefore define $X=W\odot\frac{\partial y}{\partial W}$ and $z^j_l=-\sum\limits_i X_{ij}$ (sum over the first index) and rewrite
$$
\E\Delta=-p\sum\limits_i\sum\limits_j X_{ij}=p\sum\limits_{j}z^j
$$

$$
\Var\Delta=p\sum\limits_{ijk}X_{ji}X_{ki}=p\sum\limits_i\left(\sum\limits_j X_{ji}\right)^2=p\sum\limits_i(-z_i)^2
$$
\end{proof}

\begin{proposition}\label{th:error_mean_eq}
For a neural network with $C^1$ activation function we have for a particular input $x$ in the first order as in T\ref{prop:taylor}:
$$
\E_\xi\Delta_{L+1}=-p(\nabla_W L, W)
$$
\end{proposition}

\begin{proof}
Take the expression $A=\E_\xi\Delta_{L+1}=\left(\frac{\partial L}{\partial y},\E_\xi\Delta\right)$ and consider $\E_\xi\Delta$. By Additional Corollary~\ref{th:error_eq_w} we have $\E_\xi\Delta=-p\sum\limits_{ij}\frac{\partial y}{\partial W_{ij}}W_{ij}$

Therefore,
$$A=-p\left(\frac{\partial L}{\partial y},\sum\limits_{ij}\frac{\partial y}{\partial W_{ij}}W_{ij}\right)=-p\sum\limits_{ij}W_{ij}\left(\frac{\partial L}{\partial y},\frac{\partial y}{\partial W_{ij}}\right)$$

By the chain rule for a single input $x$ we have
$$
\frac{\partial L}{\partial W_{ij}}=\left(\frac{\partial L}{\partial y},\frac{\partial y}{\partial W_{ij}}\right)
$$

We plug that back in:
$$
A=-p\sum\limits_{ij}\frac{\partial L}{\partial W_{ij}}W_{ij}
$$

This is the exact statement from the proposition.
\end{proof}

\section{Unused extra results}

\paragraph{Skip-connections.} If we consider a model with skip-connections (which do not fit Definition \ref{def:nn}) with faults at every node, we expect that an assumption similar to A\ref{assumption:continuous_net} would lead to a result similar to T\ref{prop:taylor}. However, we did not test our theory on models with skip-connections.

\paragraph{Another idea for fault tolerance.} In case if we know $p$ exactly, we can compensate for $\E\Delta$ by multiplying each neuron's output by $(1-p)^{-1}$. In this way, the mean input would be preserved, and the network output will be unbiased. However, this only works in case if we know exactly $p$ of the hardware.

\begin{proposition}[Variance for bound \boundfirst{}, not useful since the bound is not tight. Done in a similar manner as the mean bound in~\cite{ontherobustness17}]
The variance of the error $\Var\Delta$ is upper-bounded as
$$\Var\Delta\leqslant \sum\limits_{i=1}^na_i\prod\limits_{j=i+1}^n b_j$$
for $$
\begin{array}{rl}
a_L&=C_L^2p_L(\alpha_L+p_L\beta_L)+2C_Lp_L(1-p_L)\beta_L\E\Delta_{L-1},\\
b_L&=(1-p_L)(\alpha_L+(1-p_L)\beta_L)\\
C_l&=\max\{|y_{li}|\}\\
\alpha_L&=\max\{(w^{(L+1)},w^{'(L+1)})\}\\ \beta_L&=\max\{\|w^{(L+1)}\|_1\|w^{'(L+1)}\|_1-(w^{(L+1)},w^{'(L+1)})\}
\end{array}
$$
\end{proposition}

\begin{proof}In the notation of \cite{ontherobustness17}, consider
$|\Delta_L|=|\Fneu-\Ffail|=|\sum\limits_{i=1}^{N_L}w_i^{(L+1)}(y_j^{(L)}-\hat{y}_j^{(L)}\xi_j^{(L)})|$, $\E\Delta_L\leqslant p_l\|w^{(L+1)}\|_1C_L+K\|w^{(L+1)}\|_1(1-p_L)\E\Delta_{L-1}$, where $\hat{y}_j^{(L)}$ is the corrupted output from layer $L$ and $K$ is the activation function Lipschitz constant.
    Thus, $\E\Delta_L\leqslant \sum\limits_{l=1}^Lp_l\|w^{(L+1)}\|_1C_lK^{L-l}\prod\limits_{s=l+1}^L(1-p_s)\|w^{(s)}\|_1=\sum\limits_{l=1}^Lp_l\|w^{(l+1)}\|_1C_l\prod\limits_{s=l+1}^LK(1-p_s)\|w^{(s+1)}\|_1$
    \begin{enumerate}
    \item Variance: $\Var \Delta=\E(\Delta^2)-(\E\Delta)^2\leqslant\E(\Delta^2)$, here we will calculate $\E\Delta^2$:
    \item $\E\Delta^2_1=\E\sum\limits_{j_1=1}^{N_1}\sum\limits_{j_2=1}^{N_1}w_{j_1}^{(2)}y_{j_1}^{(1)}(1-\xi_{j_1}^{(1)})w_{j_2}^{(2)}y_{j_2}^{(1)}(1-\xi_{j_2}^{(1)})\leqslant f_1C_1^2\overline{w_j^2}+C_1^2\frac{f_1^2-f_1}{N_1^2-N_1}\sum\limits_{j_1\neq j_2}w_{j_1}w_{j_2}$
    
    $\frac{f_1^2-f_1}{N_1^2-N_1}\leqslant \frac{f_1^2}{N_1^2}=p^2$. $\sum\limits_{j_1\neq j_2}w_{j_1}w_{j_2}=\|w\|_1^2-\|w\|_2^2$.
    
    Therefore, $\leqslant \boxed{C_1^2p\|w\|_2^2+C_1^2p^2(\|w\|_1^2-\|w\|_2^2)}$
    \item $\E\Delta_1\Delta_1'=\E\sum\limits_{j_1=1}^{N_1}\sum\limits_{j_2=1}^{N_1}w_{j_1}^{(2)}y_{j_1}^{(1)}(1-\xi_{j_1}^{(1)})w_{j_2}^{'(2)}y_{j_2}^{(1)}(1-\xi_{j_2}^{(1)})\leqslant f_1C_1^2\overline{w_iw_i'}+C_1^2\frac{f_1^2-f_1}{N_1^2-N_1}\sum\limits_{j_1\neq j_2}w_{j_1}w_{j_2}'$
    
    $\leqslant \boxed{pC_1^2(w,w')+p^2C_1^2(\|w\|_1\|w'\|_1-(w,w'))}$
    \item $\E\Delta_L^2\leqslant \sum\limits_{i=1}^{N_L}(w_i^{(L+1)})^2\E(y_i^{(L)}-\hat{y}^{(L)}_i\xi_i^{(L)})^2+\sum\limits_{i\neq j}^{N_L}w_i^{(L+1)}w_j^{(L+1)}\E \left[(y_i^{(L)}-\hat{y}^{(L)}_i\xi_i^{(L)})(y_j^{(L)}-\hat{y}^{(L)}_j\xi_j^{(L)})\right]$
    
    $\leqslant \boxed{P_L\|w\|_2^2+Q_L(\|w\|_1^2-\|w\|_2^2)}$
    \item $\E\Delta_L\Delta_L'\leqslant \sum\limits_{i=1}^{N_L}w_i^{(L+1)}w_i^{'(L+1)}\E(y_i^{(L)}-\hat{y}^{(L)}_i\xi_i^{(L)})^2+\sum\limits_{i\neq j}^{N_L}w_i^{(L+1)}w_j^{'(L+1)}\E \left[(y_i^{(L)}-\hat{y}^{(L)}_i\xi_i^{(L)})(y_j^{(L)}-\hat{y}^{(L)}_j\xi_j^{(L)})\right]$
    
    $\leqslant\boxed{P_L(w,w')+Q_L(\|w\|_1\|w'\|_1-(w,w'))}$
    \item $P_1=pC_1^2$, $Q_1=p^2C_1^2$
    \item $P_L=\E(y-\hat{y}\xi)^2=\E y_i^2\Prob\{\xi=0\}+K^2\E\Delta_{L-1}^2\Prob\{\xi=1\}\leqslant C_L^2p_L+K^2\E\Delta_{L-1}^2(1-p_L)$.
    \item $Q_L=\E(y-\hat{y}\xi)(y'-\hat{y}'\xi')\leqslant C_L^2\underbrace{\E\eta\eta'}_{\leqslant p_L^2}+(1-p_L)p_LKC_L\E(|\Delta_{L-1}|+|\Delta_{L-1}'|)+(1-p_L)^2K^2\E\Delta_{L-1}\Delta_{L-1}'$.
    \item Consider a recurrence $x_1=a_1$, $x_n=a_n+b_nx_{n-1}$. Then $x_n=\sum\limits_{i=1}^na_i\prod\limits_{j=i+1}^{n}b_j$
    \item Define $\alpha_L=\max\{(w^{(L+1)},w^{'(L+1)})\}$ and $\beta_L=\max\{\|w^{(L+1)}\|_1\|w^{'(L+1)}\|_1-(w^{(L+1)},w^{'(L+1)})\}$. Then $$\boxed{\E\Delta_L'\Delta_L,\,\E\Delta_L^2\leqslant P_L\alpha_L+Q_L\beta_L}$$
    \item Thus, $\E\Delta_{L}^2\leqslant a_L+b_L\E\Delta_{L-1}^2$, where
    \begin{eqnarray}
    a_L&=&C_L^2p_L(\alpha_L+p_L\beta_L)+2KC_Lp_L(1-p_L)\beta_L\E\Delta_{L-1},\\
	b_L&=&K^2(1-p_L)(\alpha_L+(1-p_L)\beta_L).
    \end{eqnarray}
	 Therefore,
    
    $$\boxed{\E\Delta_L^2\leqslant \sum\limits_{l=1}^L\left(C_l^2p_l(\alpha_l+p_l\beta_l)+2KC_lp_l(1-p_l)\beta_l\E\Delta_{l-1}\right)\prod\limits_{l'=l+1}^{L}K^2(1-p_{l'})(\alpha_{l'}+(1-p_{l'})\beta_{l'})}$$
    \end{enumerate}
\end{proof}

The goal of this proposition was to give an expression for the variance in a similar manner as it is done for the mean in~\cite{ontherobustness17}. However this proposition did not make it to the article because bound b1 was not showing any good experimental results.

\section{Introduction into fault tolerance for neuromorphic hardware}
{\bf Neuromorphic hardware (NH).}
Given the amount of processing power required for modern Machine Learning applications,   emerging hardware technologies are nowadays reviving the {\em neuromorphic} project and its promise to cut off energy consumption of machine learning by several orders of magnitude~\cite{neuromorphicIBM}. Neuromorphic implementation of an NN is a physical device where each neuron corresponds to a piece of hardware, and neurons are physically linked forming weights. Thus, the computation is done (theoretically) at the speed of light, compared to many CPU/GPU cycles. The surge in performance could arguably even exceed the one that followed the switch from training on CPUs to training on GPUs and  TPUs~\cite{amodei}. Recent results on neuromorphic computing report on concrete successes such as milliwatt image recognition~\cite{neuromorphicIBM} or basic vowel recognition using only four coupled nano-oscillators~\cite{julie}. Since the components of a neuromorphic network are small \cite{tran2011design} and unreliable \cite{liu2019fault}, there are crashes in individual {\em neurons} or {\em weights} inside the network \cite{tran2011design,liu2019fault}. They lead to a performance degradation. A failure within a neuromorphic architecture involved in a mission-critical application could be disastrous. Hence, fault tolerance in neural networks is an important concrete Artificial Intelligence (AI) safety problem~\cite{amodei2016concrete}. In terms of fault tolerance, the unit of failure in these architectures is fine-grained, i.e., an individual {\em neuron} or a single {\em synapse}, with failure mode frequently being a complete crash. This is in contrast with the now classical case of a neural network as a software deployed on a single machine where the unit of failure is coarse-grained, i.e., the whole {\em machine}  holding the entire neural network. For instance, in the popular distributed setting of ML, the so-called parameter-server scheme~\cite{li2014scaling}, the unit of failure is a {\em worker} or a {\em server}, but never a {\em neuron} or a {\em synapse}. Whilst very important,  fine-grained fault tolerance in neural networks has been overlooked as a concrete AI safety problem.

% \begin{figure}[ht]
% 	\centering
% 	\setlength{\fboxsep}{2pt}%
% 	\setlength{\fboxrule}{1pt}%
% 	\fbox{
% 	\includegraphics[width=0.3\textwidth]{figures/cat.jpg}
% 	}
% 	\label{fig:thecat}
% 	\caption{The image of a cat from Google images}
% \end{figure}